\documentclass[10pt,twocolumn,letterpaper]{article}

\usepackage{iccv}
\usepackage{times}
\usepackage{epsfig}
\usepackage{graphicx}
\usepackage{overpic}
\usepackage{cite}
\usepackage{amsmath}
\usepackage{amssymb}
\usepackage{color}
\usepackage{multirow} 
\usepackage{bm}
\usepackage{subcaption}

\iccvfinalcopy 

\newcommand{\tabincell}[2]{
	\begin{tabular}{@{}#1@{}}#2\end{tabular}
}

\def\iccvPaperID{****} 

\newcommand{\x}		{\mathbf{x}}	
\newcommand{\X}		{\mathbf{X}}	
\newcommand{\y}		{\mathbf{y}}
\newcommand{\A}		{\mathbf{A}}
\newcommand{\z}	 {\mathbf{z}}
\newcommand{\W}	 {\mathbf{W}}
\renewcommand{\P}	 {\mathbf{P}}
\newcommand{\omg}	 {\mathbf{\Omega}}

\newcommand{\Z}		{\mathbf{Z}}

\newcommand{\R}	     {\mathsf{R}}
\newcommand{\Q}	     {\mathbf{Q}}
\newcommand{\DIFadd}{\textcolor{black}}

\ificcvfinal\fi

\begin{document}
\bibliographystyle{unsrt}


\title{SUPER Learning: A Supervised-Unsupervised Framework for Low-Dose CT Image Reconstruction~\thanks{Copyright  (c)  2019  IEEE.  Personal  use  of  this  material  is  permitted.Permission from IEEE must be obtained for all other uses, in any current or future media, including reprinting/republishing this material for advertising or promotional purposes, creating new collective works, for resale or redistribution  to  servers  or  lists,  or  reuse  of  any  copyrighted  component  of  this  work in other works.} \vspace{-0.0in}}

\author{Zhipeng Li$^1$ \quad Siqi Ye$^1$ \quad Yong Long$^1$\thanks{Yong Long is the corresponding author.} \quad Saiprasad Ravishankar$^2$\\
	$^1$University of Michigan - Shanghai Jiao Tong University Joint Institute, \\Shanghai Jiao Tong University\thanks{This work was supported by NSFC (61501292).} , Shanghai, China\\
	$^2$Departments of Computational Mathematics, Science and Engineering, \\and Biomedical Engineering, Michigan State University, East Lansing, MI, USA\\
	{\tt\small \{zhipengli,~yesiqi,~yong.long\}@sjtu.edu.cn, ravisha3@msu.edu}  \vspace{-0.1in}	
}

\maketitle

\ificcvfinal\thispagestyle{empty}\fi

\begin{abstract}
	Recent years have witnessed growing interest in machine learning-based models and techniques for low-dose X-ray CT (LDCT) imaging tasks. The methods can typically be categorized into supervised learning methods and unsupervised or model-based learning methods. Supervised learning methods have recently shown success in image restoration tasks. However, they often rely on large training sets.
	Model-based learning methods such as dictionary or transform learning do not require large or paired training sets and often have good generalization properties, since they learn general properties of CT image sets.
	Recent works have shown the promising reconstruction performance of methods such as PWLS-ULTRA that rely on clustering the underlying (reconstructed) image patches into a learned union of transforms.
	In this paper, we propose a new Supervised-UnsuPERvised (SUPER) reconstruction framework for LDCT image reconstruction that combines the benefits of supervised learning methods and (unsupervised) transform learning-based methods such as PWLS-ULTRA that involve highly image-adaptive clustering. 
	The SUPER model consists of several layers, each of which includes a deep network learned in a supervised manner and an unsupervised iterative method that involves image-adaptive components. The SUPER reconstruction algorithms are learned in a greedy manner from training data.
	The proposed SUPER learning methods dramatically outperform both the constituent supervised learning-based networks and iterative algorithms for LDCT, and use much fewer iterations in the iterative reconstruction modules.
\end{abstract}


\section{Introduction}
\vspace{-0.03in}
X-ray computed tomography (CT) is a popular imaging modality in many clinical and industrial applications.
There has been particular interest in CT imaging with low X-ray dose levels that would reduce the potential risks to patients from radiation.
However, image reconstruction at low X-ray dose levels is challenging.
Conventional X-ray CT image reconstruction methods include analytical methods, and model-based iterative reconstruction (MBIR) methods. 
A classical analytical method is the filtered back-projection (FBP) method \cite{feldkamp1984practical}. 
That can be degraded excessively by noise and streak artifacts in low-dose situations~\cite{streak-artifact-mA09,nonlocalmean2014}. 

MBIR methods incorporate the system physics, statistical model of measurements, and typically certain simple prior information of the unknown object~\cite{survey2013-ct}. 
A typical method of this kind is the penalized weighted-least squares (PWLS) method, for which the cost function includes a weighted quadratic data-fidelity term that models the measurement statistics, and a penalty term called a regularizer that models the prior information \cite{pwls1994jf,thibault:07:atd,beister2012iterative}. 
For PWLS, various optimization approaches and regularization designs have been exploited with efficiency and convergence guarantees. 

Adopting appropriate prior knowledge of images for MBIR approaches is also important to improve CT reconstruction. 
More recently, with the availability of data sets of \mbox{CT} images, methods based on big-data priors have gained interest, such as dictionary learning-based techniques~\cite{xu:12:ldx}. 
The dictionary can be either pre-learned from training data, or adaptively learned with the reconstruction. 
In particular, the synthesis dictionary learning approaches represent a signal or image patch as a sparse linear combination of the atoms or columns of a learned dictionary, and have obtained promising results in many applications \cite{ravishankar2010mr,mairal2007sparse,elad2006image}. 
However, the dictionary learning based MBIR approaches are often computationally expensive due to expensive sparse coding (where typically NP-hard problems are optimized for estimating sparse coefficients). Different from synthesis dictionary learning, sparsifying transform (a generalized analysis dictionary model) learning techniques efficiently adapt an operator to approximately sparsify signals in transform domains, and the corresponding transform sparse coding problem can be solved exactly and cheaply by thresholding~\cite{STlearning13}.
Sparsifying transform learning techniques including using doubly-sparse transforms and unions of transforms have been applied to image reconstruction and obtained promising results~\cite{doublyST13,pwls-ultra2018,ye2018spultra}.

\vspace{-0.03in}
Very recently, there has been growing interest in \DIFadd{deep learning approaches for medical imaging problems \cite{yang2017dagan,yu2017deep,schlemper2018stochastic,jin:17:dcn,kang2017deep,WavResNet18}}. \DIFadd{In the LDCT image reconstruction field, typical deep learning methods} learn the reconstruction mapping from large datasets of pairs of (low-dose and regular-dose) scans.
\DIFadd{These methods} include image-domain learning, sensor-domain learning, and hybrid-domain learning. 
For example, a particular image-domain learning approach is the FBPConvNet scheme~\cite{jin:17:dcn} that solves the normal-convolutional inverse problems by applying a (learned) CNN after the direct inversion that encapsulates the system physics. 
The image-domain learning approaches can have many variations. For example, instead of directly working in the image domain, one can transform the images to a specific domain and learn in such domain the relationship between training pairs. Kang et al.~\cite{kang2017deep} designed a neural network that learns a mapping between contourlet transform coefficients of the low-dose input and its high-dose counterpart. This work was later extended to learn a wavelet domain residual network (WavResNet)~\cite{WavResNet18}. 

\vspace{-0.03in}
In the sensor-domain deep learning category, W{\"u}rfl et al.~\cite{wurfl2016deep} proposed an end-to-end neural network for low-dose CT that maps the sinogram to the reconstructed image by mapping the filtered back-projection algorithm to a basic neural network. This allows one to take into account the artifacts in the sensor-domain, e.g., the scatter and beam-hardening artifacts, and compensate them in the learning process. 
Another framework named the Automated Transform by Manifold Approximation (AUTOMAP) \cite{AUTOMAP2018} \mbox{learns} a direct mapping from the measurement domain to image domain. However, due to the high memory requirements for storing fully connected layers, it is challenging for AUTOMAP to handle large scale reconstruction tasks such as CT image reconstruction.
Hybrid-domain learning approaches exploit data-fidelity terms in the neural network architecture. The Learned ISTA (LISTA) \cite{LISTA2010} was one of the earliest work of this kind. LISTA unfolds the iterative soft-thresholding (ISTA) algorithm~\cite{fista2009}, and learns the weight matrices and the sparsifying soft-thresholding operator. Later, Yang et al. \cite{admm-net2016} proposed an ADMM-Net which unfolds the alternating direction method of multipliers algorithm for image reconstruction. Each step of the \mbox{algorithm} is mapped to a neural network module. 
This idea was then extended to a learnable primal-dual approach based CNN~\cite{adler2018primaldual}. 
These methods fall in the class of physics-driven deep learning methods~\cite{ravchfess17,ravacfess18,chfess18}.
Hybrid-domain approaches also include a type that applies a plug-and-play model. He et al.~\cite{3pADMM} applied the plug-and-play model to the ADMM algorithm and unfolded it into a deep reconstruction network, so that each network module is learnable and replaceable.

Most deep learning algorithms are learned in a supervised manner (using task-specific cost functions) and require large training sets. However, in CT imaging, it is often difficult to acquire large datasets of training image pairs. Even though in the AAPM X-ray CT Low-Dose Grand Challenge, both regular-dose and the matched quarter-dose images were provided, only the regular-dose images were reconstructed from real scans, while the matched quarter-dose images were synthesized by adding noise to the regular-dose sinogram data. Therefore, training with smaller number of paired data (and yet generalizing) or without reference data is highly conducive for CT image reconstruction.
Moreover, different machine learning (as well as conventional) approaches such as dictionary or transform learning and deep learning use different types of big-data priors and are advantageous in different ways. For example, transform learning approaches learn general image properties and features in an unsupervised or model-based manner, and can easily and effectively adapt to specific image instances.

\begin{figure*}[!t]
	\centering
	\includegraphics[scale=0.67, trim = 10 170 6 130, clip]{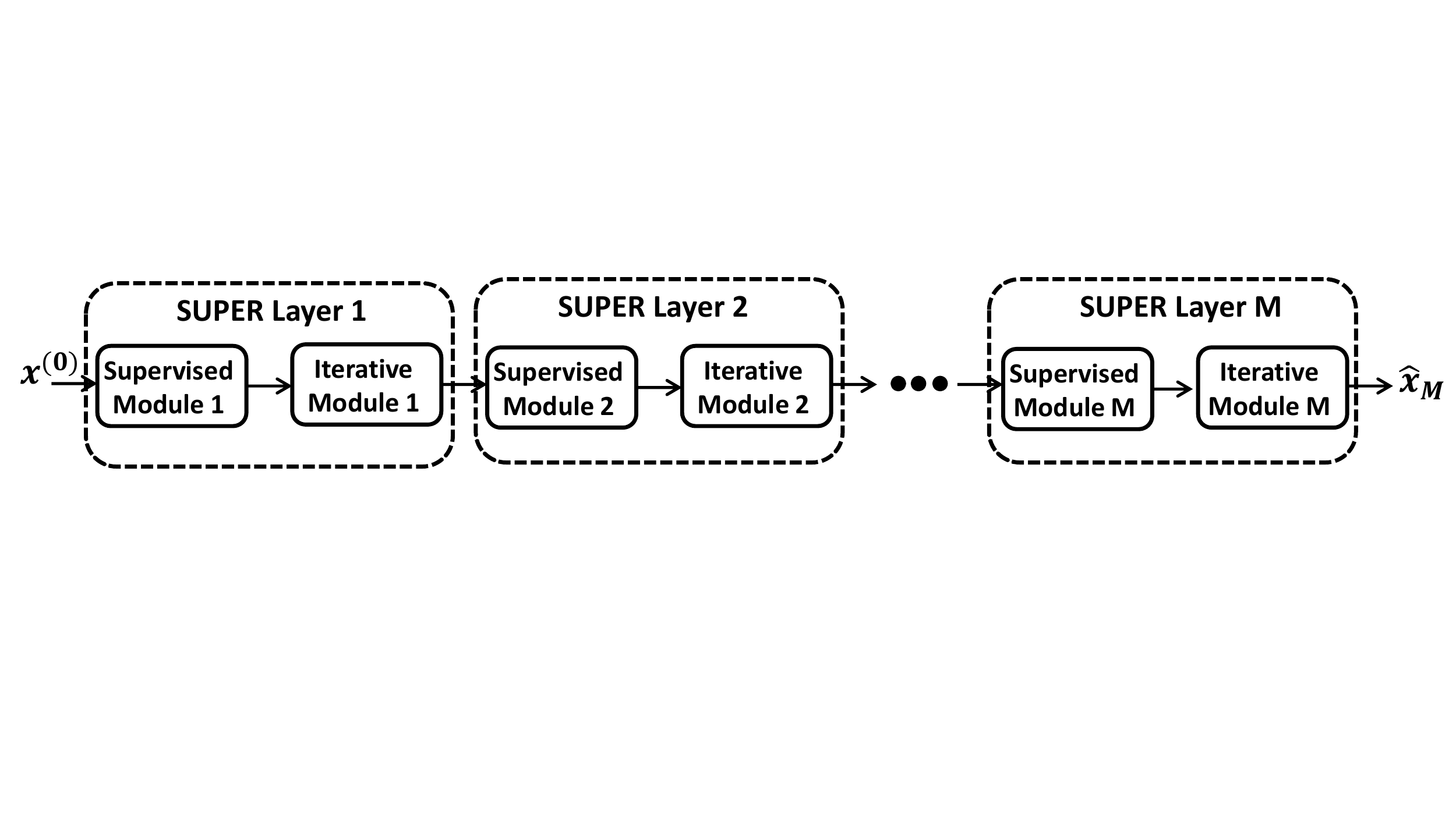}
	\vspace{-0.09in}
	\caption{Overall structure of the proposed reconstruction framework.}
	\label{fig:flowchart}
	\vspace{0.02in}
\end{figure*}

In this work, we propose a new image reconstruction framework for LDCT dubbed Supervised-UnsuPERvised (SUPER) learning.
The algorithm architecture involves interconnected supervised (deep network) and unsupervised (iterative reconstruction) modules over many layers. The architecture enables effectively leveraging different kinds of big data learned priors 
for CT reconstruction.
For example, we used FBPConvNet~\cite{jin:17:dcn} as the supervised (deep) learned module and PWLS-ULTRA~\cite{pwls-ultra2018} as the unsupervised module with a pre-learned union of transforms, which provided both high quality image reconstruction and image-adaptive clustering.
The proposed SUPER learning used relatively small training sets and dramatically outperformed both deep learning and transform/dictionary learning by effectively combining task-specific and image or instance-specific adaptivity.
The proposed framework is generalizable to include various constituent modules including non-learning based algorithms, as shown in our experiments.


\section{Proposed Model and Algorithms}
\vspace{-0.03in}
Here, we present the proposed reconstruction model, its motivations and interpretations, example architectures, and training method.

\vspace{-0.0in}
\subsection{Overview of the SUPER Model}
\vspace{-0.0in}
We propose a novel efficient physics-driven learning framework for CT reconstruction that effectively combines the benefits of supervised (deep) learning and unsupervised iterative reconstruction methods.
The proposed reconstruction architecture takes an initial image as input and processes it through multiple ``super" layers (Fig.~\ref{fig:flowchart}). 
Each super layer consists of a network learned in a supervised manner (\emph{supervised module}) and an iterative reconstruction method (\emph{iterative module}) in sequence.
The supervised module is different in each super layer\DIFadd{, i.e., the weights in the supervised module are not shared among the super layers}. 
Importantly, this module is learned in a supervised manner (e.g., to minimize reconstruction error) to remove artifacts and noise.
The iterative module on the other hand iteratively optimizes a regularized image reconstruction problem using image-adaptive priors or regularizers (e.g., the patches of the underlying image can be clustered and sparsified in a learned union of transforms or dictionaries~\cite{pwls-ultra2018}). The iterative algorithm is run for a fixed number of iterations in each super layer.


While the supervised module removes image noise and artifacts using a single learned network, the iterative module could adapt various image-specific features in an MBIR setup to further improve image quality and remove \mbox{artifacts}. Importantly, the iterative module is not learned in a supervised manner. 
The SUPER model in Fig.~\ref{fig:flowchart} is flexible and could use various architectures for the supervised module (e.g., FBPConvNet~\cite{jin:17:dcn}, WavResNet~\cite{WavResNet18}, etc.) and a variety of iterative data-adaptive methods (e.g., PWLS-ULTRA~\cite{pwls-ultra2018}). The model could be potentially used in a variety of imaging as well as other applications.

\vspace{-0.0in}
\subsection{Interpretations and Generalization}
\vspace{-0.0in}
The SUPER model enables combining different kinds of machine learned models and priors in a common reconstruction framework. While the supervised module could be a deep convolutional network learned from a big dataset to optimize task-specific performance metrics, the iterative module could exploit models learned from images using criteria such as sparsity, manifold properties, etc.
For example, an operator could be pre-learned from CT images or patches to approximately sparsify them (a.k.a. transform learning~\cite{STlearning13}) and used to construct the regularizer for the iterative module.
Such image-based learned operators are not typically task-sensitive and often generalize readily to different settings and can help delineate or reconstruct various image features.

The proposed SUPER model can also help combine global adaptivity and image-specific adaptivity to obtain the best of both worlds. While each supervised module is learned from a dataset and fixed during reconstruction, the iterative module could optimize novel and specific features for each image being reconstructed (during training and testing) and thus capture the diversity of images and enable highly adaptive reconstructions.
For example, during training and testing, the iterative module could cluster image patches differently for each image~\cite{pwls-ultra2018}, or even learn novel models such as dictionaries for each image~\cite{ravishankar:16:tci}.

Another interpretation of the SUPER model arises from the perspective of iterative reconstruction. Many \mbox{recent} state-of-the-art MBIR schemes involve complex nonconvex optimization and priors, wherein the initialization of the algorithm is typically quite important, and better initializations can lead to better reconstructions.
In the SUPER model, the iterative module is ``initialized" with a different image (i.e., output of the corresponding supervised module) in each super layer. If the output of the supervised module improves in quality over layers, the iterative module will see increasingly better initializations and could thus provide better quality outputs over layers. Moreover, the \mbox{parameters} of the iterative module could also be varied over layers to provide optimal bias-noise trade-offs.
Thus, the SUPER model could be viewed as minimizing nonconvex costs in sequence with better initializations and parameters.


The proposed SUPER model for LDCT reconstruction can be generalized to incorporate a variety of iterative and MBIR techniques in the iterative modules.
For example, conventional techniques such as PWLS-EP (edge-preserving hyperbola regularizer)~\cite{cho:15:rdf} could be used in the iterative module.
PWLS-EP is a non-adaptive method that penalizes the differences between neighboring pixels in the reconstruction. We show later that combining PWLS-EP with supervised learning in the SUPER model boosts the performance of both methods. Note that we do not run the PWLS-EP modules to near convergence (as it involves a strictly convex problem and a unique minimum) but only for multiple iterations.



\vspace{-0.05in}
\subsection{Examples of SUPER Architectures}
\vspace{-0.03in}
We now discuss some example SUPER models and their properties.
To illustrate the proposed approach, in this work, we focused on the recent FBPConvNet~\cite{jin:17:dcn} for the supervised module. For the iterative module, we chose the conventional PWLS-EP approach that uses a hand-crafted prior (edge-preserving hyperbola regularizer) as well as the learning and clustering-based PWLS-ULTRA \cite{pwls-ultra2018}. Our experiments later show that combining such supervised and iterative methods improves image quality over the constituent methods. In the following, we further discuss the chosen architectures.


\vspace{-0.13in}
\subsubsection{Supervised Module}
We work with FBPConvNet, which is a CNN-based image-domain denoising architecture, originally designed for sparse-view CT.
The backbone of FBPConvNet is a U-net \cite{ronneberger:15:unc} like CNN that takes noisy images reconstructed by the FBP method (from low-dose scans) as input.
The neural network is trained so that its outputs closely match the reference high-quality (true) images, e.g., in an $\ell_2$ norm or root mean squared error (RMSE) sense.


The traditional U-net uses a multilevel decomposition, and employs a dyadic scale decomposition based on max pooling. Thus, the effective filter size in the middle layers is larger than that in the early and late layers. This scheme is important because the filters corresponding to the Hessian of the data fidelity term in \eqref{eq:recon} may have noncompact support. Similar to U-net, FBPConvNet employs multichannel filters, which is the standard approach in CNNs, to increase the capacity of the network. Compared with the traditional U-net, FBPConvNet adopts a residual learning strategy to learn the difference between the input and output. 

\vspace{-0.15in}
\subsubsection{Iterative Module}
\vspace{-0.03in}

The iterative module optimizes an MBIR problem that estimates the linear attenuation coefficients $\x\in \mathbb{R}^{N_p}$ from the measurements $\y\in \mathbb{R}^{N_d}$. The typical PWLS approach involves a cost function of the following form:
\vspace{-0.06in}
\begin{equation}
	\x=\mathop{\arg\min}_{\x\ge \mathbf{0}} \|\y-\A\x\|^2_\W + \beta \R(\x),
	\label{eq:recon}
	\vspace{-0.06in}
\end{equation}
where $\A\in\mathbb{R}^{N_d\times N_p}$ is the CT system matrix, $\W$ is the weighting matrix related to the measurements (capturing measurement statistics), $\R(\x)$ is the regularizer, and $\beta$ is a positive scalar controlling the balance between the data-fidelity term and the regularizer. 
In this paper, we used the unsupervised learning-based PWLS-ULTRA as well as the conventional PWLS-EP for \eqref{eq:recon}.

For PWLS-EP, the regularizer $\R(\x)$ can be written as ${\R(\x) = \sum_{j  =1}^{N_p} \sum_{k\in N_{j}}\kappa_{j} \kappa_{k} \varphi(x_j - x_k)}$, where $x_j$ is the $j$th pixel of $\x$, $N_j$ is the neighborhood of the $j$th pixel, and $\kappa_j$ and $\kappa_k$ are analytically determined weights that encourage resolution uniformity~\cite{cho:15:rdf}. 
The potential function $\varphi (t)$ is often chosen as ${\varphi (t) = \delta^2(|t/\delta| - \log(1+|t/\delta|))}$, for $\delta>0$. PWLS-EP enforces approximate sparsity of gradients of the image, a non-adaptive prior.

PWLS-ULTRA pre-learns a union of sparsifying transforms from a dataset of image patches, and uses the learned model during reconstruction~\cite{pwls-ultra2018}. The formulation for learning the union of sparsifying transforms is as follows: 
\vspace{-0.03in}
\begin{equation}
	\begin{aligned}
		\min_{\{\omg_k, \Z_i, C_k\}} &\sum_{k=1}^{K}\sum_{i\in C_k} \{ \|\omg_k\X_i - \Z_i\|_2^2 + \eta \|\Z_i\|_0 \} \\
		&+ \sum_{k=1}^{K}\lambda_k \Q(\omg_k), \text{ s.t. } C_k \in \bm{\mathcal{G}}.
	\end{aligned} \label{ultralearn}
	\vspace{-0.05in}
\end{equation}
Here, $\omg_k$, $C_k$, and $\Z_i$ represent the learned transform for the $k$th class, the set of indices of patches belonging to the $k$th class, and the sparse coefficient of the $i$th training signal or patch $\X_i$ ($N$ training signals assumed), respectively.
Each signal is grouped with a corresponding best matched sparsifying transform in \eqref{ultralearn}.
The set $\bm{\mathcal{G}}$ is the set of all possible partitions of $\left [ 1:N \right ]$ into $K$ disjoint subsets.
To avoid trivial solutions for $\omg_k$, the penalty terms $\Q(\omg_k)\triangleq\|\omg_k\|_F^2 - \log|\det \omg_k|$ for $1\leq k\leq K$ are used that also control the condition number of the transforms. 
Parameters $\eta$ and $\lambda_k = \lambda_0 \sum_{i\in C_k} \left \| \X_i \right \|_{2}^{2}$ are positive weighting factors with $\lambda_0>0$~\cite{pwls-ultra2018}. Problem~\eqref{ultralearn} is solved efficiently using alternating optimization~\cite{pwls-ultra2018}.


With the pre-learned transforms $\{\omg_k\}$, the regularizer $\R(\x)$ for image reconstruction is as follows:
\vspace{-0.04in}
\begin{equation}
	\begin{split}
		\nonumber \R(\x)\triangleq \min_{\{\z_j, C_k\}} \sum_{k=1}^{K} \sum_{j\in{C_k}} \tau_j \left\{\|\omg_k\P_j\x-\z_j\|^2_2+\gamma^2\|\z_j\|_0\right\},
	\end{split}
	\vspace{-0.04in}
\end{equation}
where $\{\tau_j\}$ are patch-based weights to encourage uniform spatial resolution or uniform noise (see~\cite{pwls-ultra2018}), $\P_j\in\mathbb{R}^{l\times N_p}$ is a patch extraction operator that extracts the $j$th patch from $\x$. $\z_j$ is the sparse coefficient for the $j$th patch, and $\gamma>0$ is a parameter controlling sparsity. 

The PWLS-ULTRA problem is efficiently solved by alternating between updating $\x$ (\textit{image update step}), and solving for $\{\z_j, C_k\}$ (\textit{sparse coding and clustering step}). 
In the image update step, where $\{\z_j, C_k\}$ are fixed, the subproblem is quadratic with non-negativity constraints, and can be solved using fast iterative algorithms such as the relaxed linearized augmented Lagrangian  method with ordered-subsets (relaxed OS-LALM)~\cite{nien:16:rla,OGM2016}.
The sparse coding and clustering step with fixed $\x$ is solved exactly~\cite{ravishankar:16:tci}, with the optimal class assignment $\hat{k}_j$ for each patch given as
\begin{equation}
	\nonumber \mathop{\arg\min}_{1\le k \le K} \|\omg_k\P_j\x-H_\gamma(\omg_k\P_j\x)\|^2_2+\gamma^2\|H_\gamma(\omg_k\P_j\x)\|_0.
	\label{eq:cluster}
\end{equation}
The corresponding optimal ${\hat{\z}_j = \mathit{H}_{\gamma}(\omg_{\hat{k}_j}\P_j\x)}$, where $\mathit{H}_{\gamma}(\cdot)$ is the hard-thresholding operator that sets elements smaller than $\gamma$ to zero.
The hard-thresholding can be viewed as the non-smooth nonlinearity in PWLS-ULTRA. The clustering could vary from image to image and iteration to iteration in the PWLS-ULTRA algorithm.


\subsection{Training and Implementation}

We propose to train the SUPER model layer-by-layer (sequentially) from a dataset of pairs of low-dose and regular-dose CT measurements.
For example, the FBP method can be used to obtain reconstructions from the measurements. The initial low-dose reconstructed images are then used as inputs to the first supervised module, which is trained to minimize the reconstruction error (RMSE) at its output, with respect to the regular-dose reconstructions. The initial images are then passed through the trained network, following which the iterative algorithm in the first iterative module is run for each training image (in parallel) to produce iterative reconstructions.
The iterative reconstructions serve as inputs to the subsequent supervised model, which is trained to minimize reconstruction error. \DIFadd{The subsequent supervised modules are thus learned sequentially.}

Once trained, the SUPER model is readily implemented for test data by passing initial reconstructions sequentially through the supervised learned networks and iterative algorithms.

\vspace{-0.1in}
\section{Experiments}
\vspace{-0.04in}
Here, we first describe our experimental setup, training procedures, and evaluation metrics. Then we present results for the learned SUPER-ULTRA model, and compare these with those obtained by each individual module of SUPER, i.e., FBPConvNet and PWLS-ULTRA. We also tested the generalized SUPER model that replaces the unsupervised learning-based PWLS-ULTRA with the non-adaptive PWLS-EP. This scheme is dubbed FBPConvNet+EP.

\vspace{-0.03in}
\subsection{Experimental Setup}
\vspace{-0.04in}

We used regular-dose CT images of two patients from the Mayo Clinics dataset established for ``\textsl{the 2016 NIH-AAPM-Mayo Clinic Low Dose CT Grand Challenge}"~\cite{McC}, to evaluate the performance of the proposed SUPER learning.
The two patient datasets (L067 and L096) contain 224 and 330 real in-vivo slice images, respectively.  
We simulated low-dose CT measurements $\y_{l}$ from the provided regular-dose images with GE 2D LightSpeed fan-beam CT geometry corresponding to a monoenergetic source. We projected the regular-dose images $\x^*$ to sinograms and added Poisson and additive Gaussian noise to them as follows: 
\begin{equation}
\y_{l_i} = \text{Poisson} \{I_0 e^{-[\A\x^*]_i}\} + \mathcal{N}\{0, \sigma^2\}.
\end{equation}
We chose $I_0 = 1\times 10^5$ photons per ray \DIFadd{ and $\sigma=5$} in our experiments. \DIFadd{We approximated elements of the diagonal weighting matrix $\W$ of the data-fidelity in \eqref{eq:recon} by $\frac{\y_{l_i}^2}{\y_{l_i}+\sigma^2}$ \cite{pre-post-log}}.
The images are of size $512\times 512$ at a resolution of $0.9766 \text{ mm} \times 0.9766 \text{ mm}$, when reconstructed using the FBP method. These reconstructed low-dose FBP images were paired with their corresponding regular-dose CT images for training the SUPER model.

\vspace{-0.08in}
\subsection{Training Procedures}
\vspace{-0.04in}

\begin{figure}[htb]
	\centering
	\includegraphics[scale=0.19]{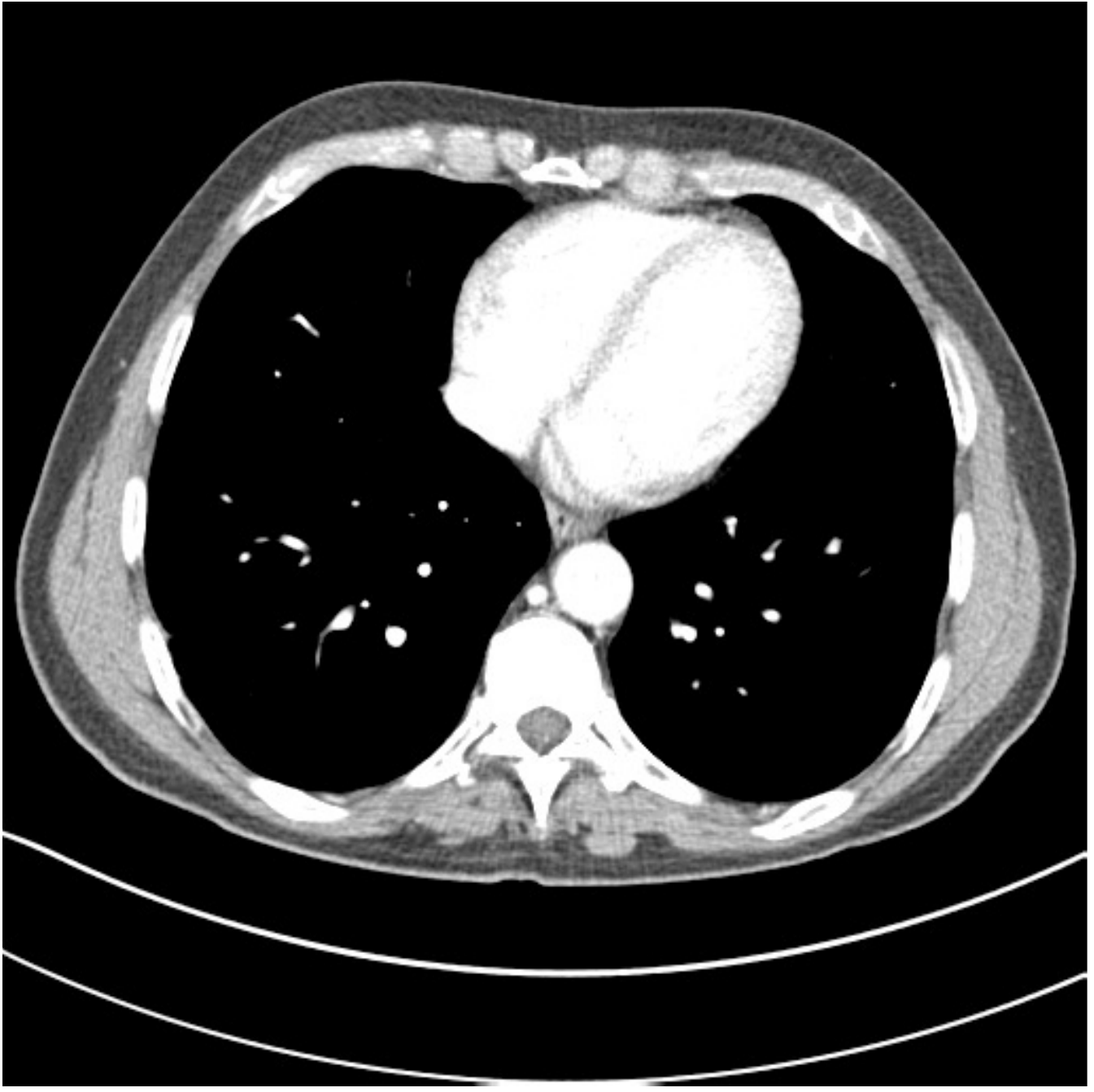}
	\includegraphics[scale=0.19]{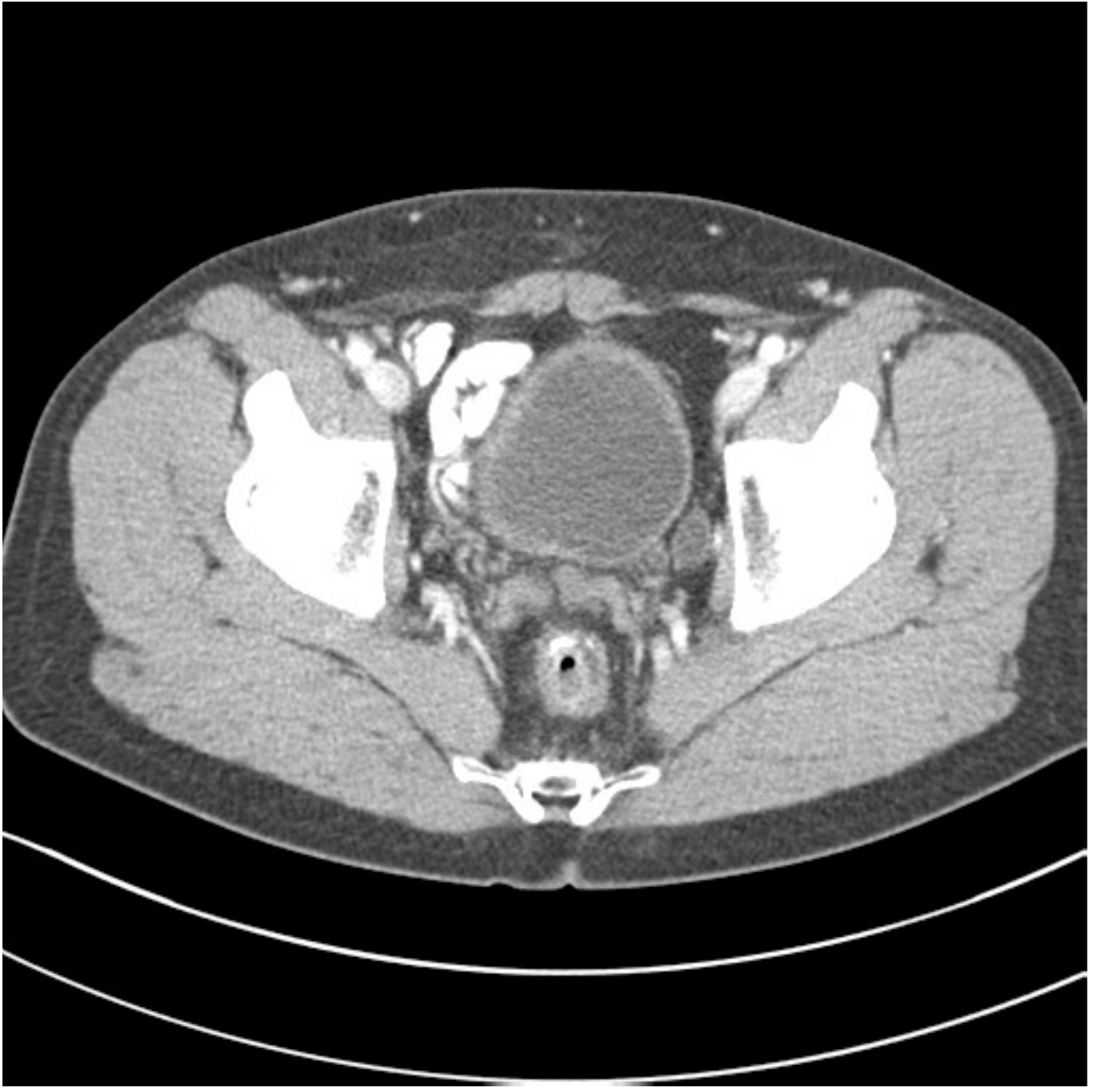}
	\includegraphics[scale=0.19]{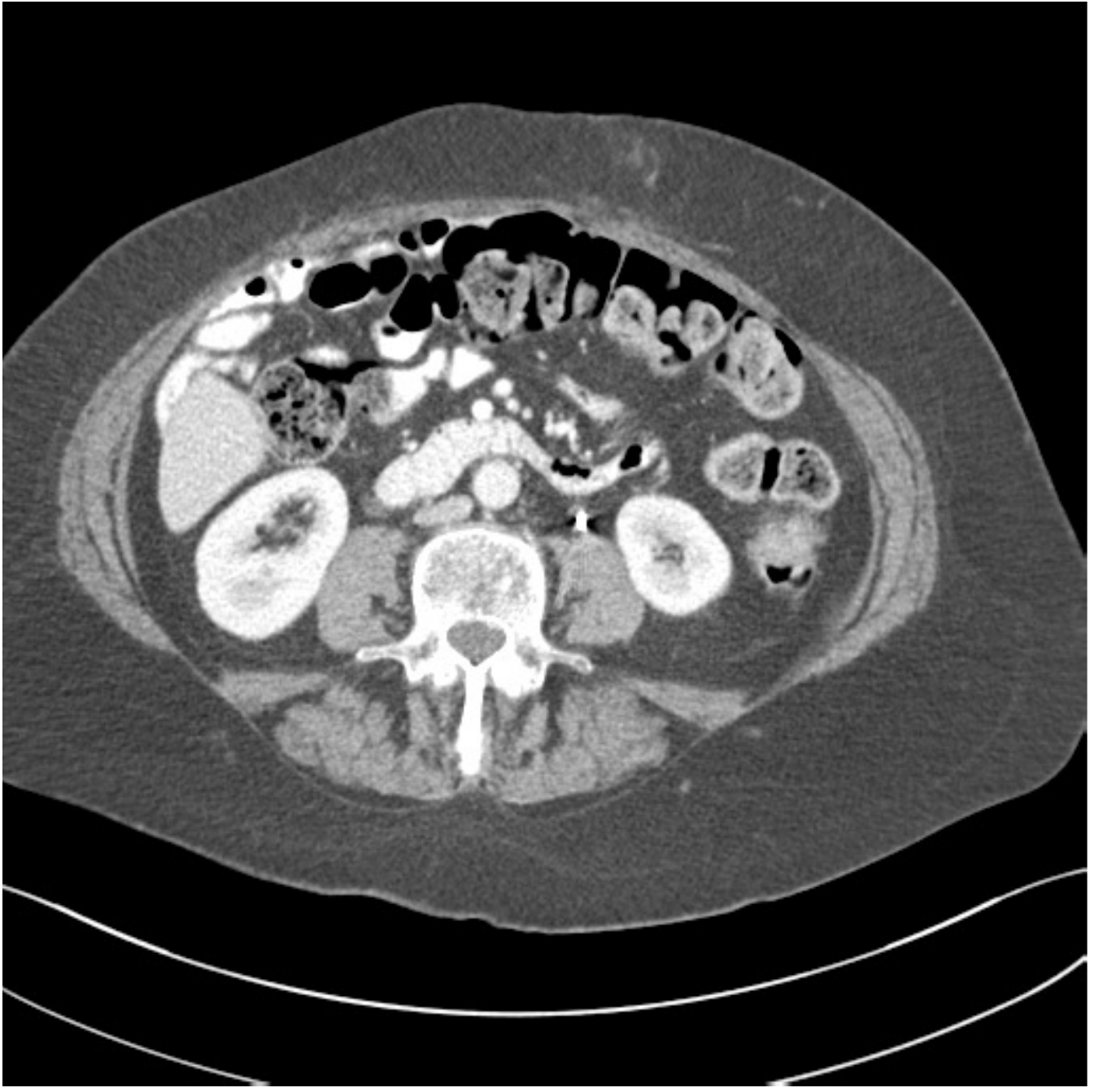} \\
	\includegraphics[scale=0.15]{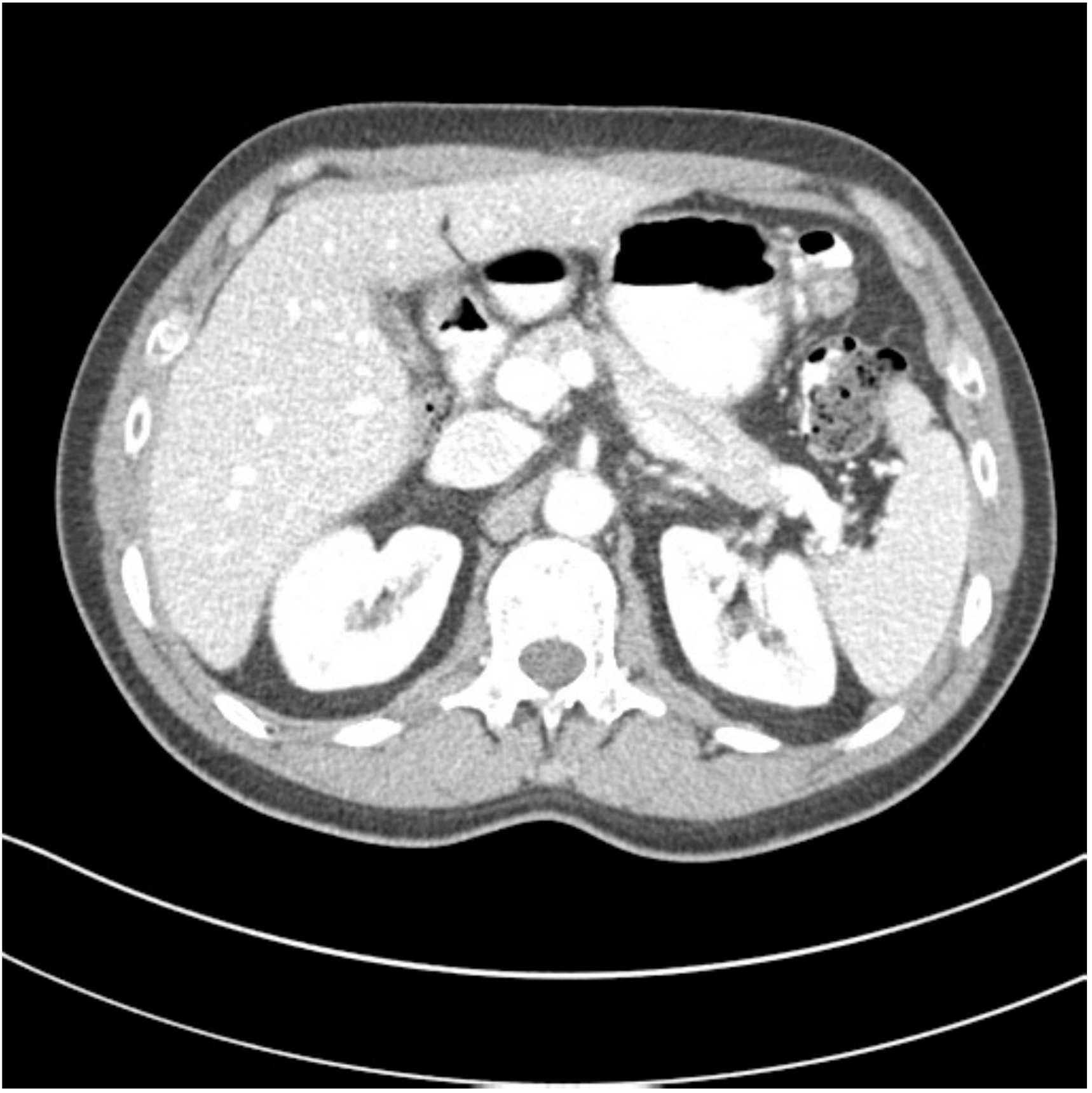}
	\includegraphics[scale=0.15]{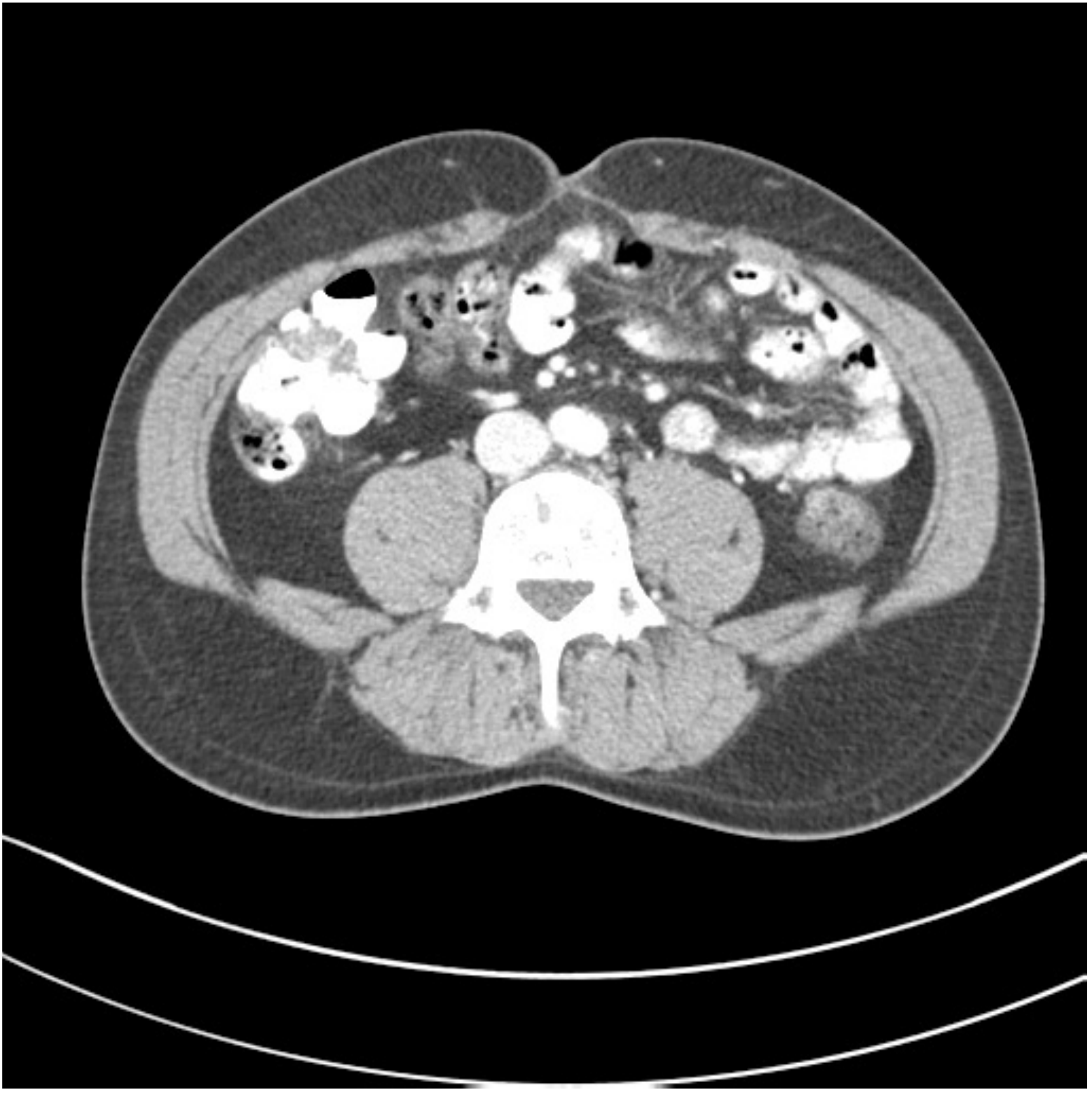}
	\includegraphics[scale=0.19]{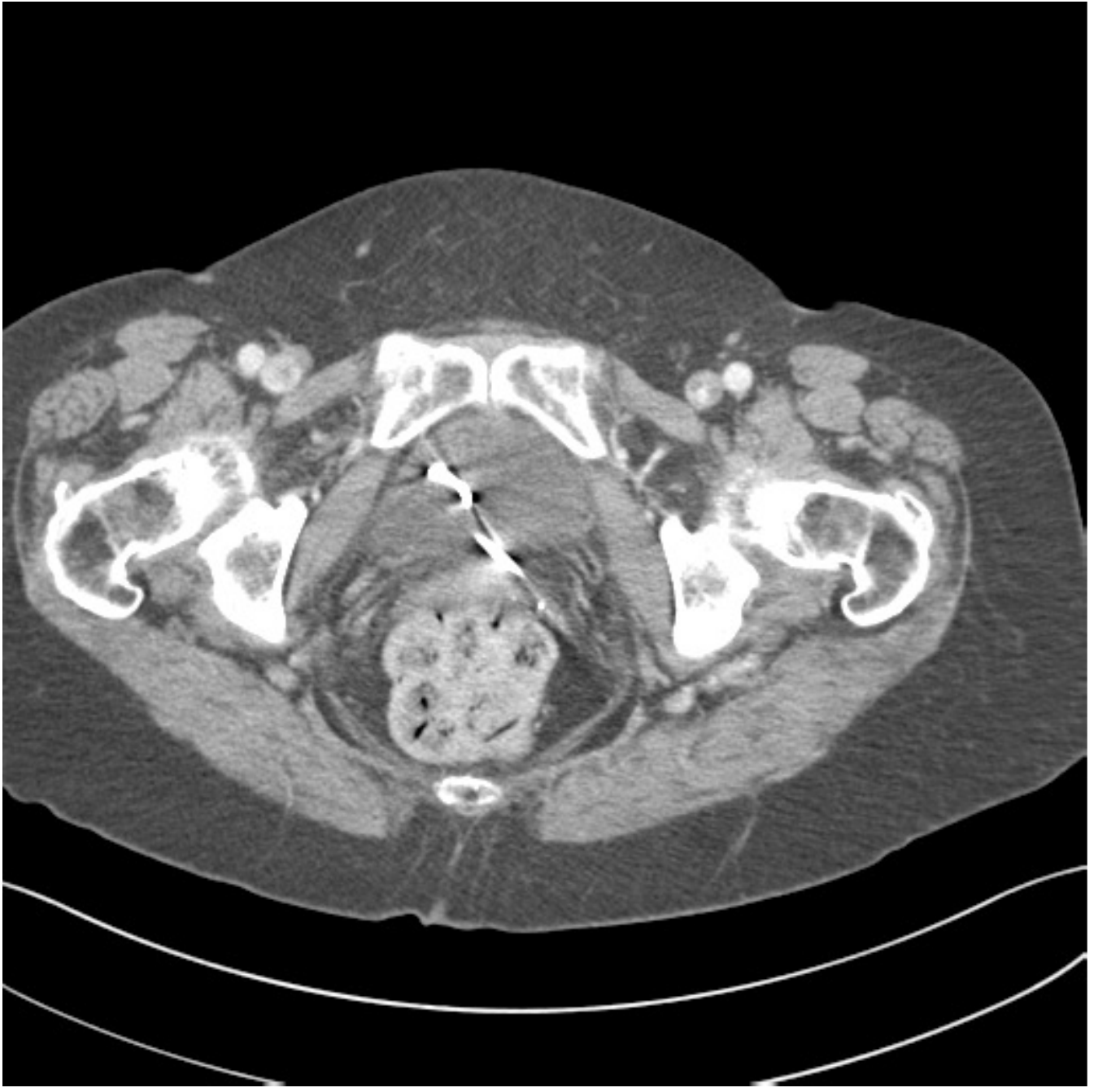}
	\vspace{-0.05in}	
	\caption{Example CT images in the training (top row) and testing (bottom row) datasets. The display window is [800 1200]~HU.}
	\label{fig:training_data}
	\vspace{-0.1in}
\end{figure}

In our experiments, the total number of training image pairs was 100,
of which 50 were chosen from patient L067 and 50 from patient L096. The image set was used to train the networks of FBPConvNet, FBPConvNet+EP, and SUPER-ULTRA. 
Fig.~\ref{fig:training_data} shows some example (regular-dose or reference) images from the training and testing \mbox{datasets}.
Different body parts are included in the datasets. 
In our experiments, we used the default network architecture in the FBPConvNet public implementation.
For PWLS-ULTRA, we pre-learned a union of five sparsifying transforms (corresponding to five classes) from twelve (regular-dose) \mbox{slices} that include three slices each from four patients (L109, L143, L192, L506).


We used a GTX Titan GPU  graphic processor for training and testing.
The union of transforms was learned efficiently using similar parameters as in~\cite{pwls-ultra2018}.
For SUPER-ULTRA and FBPConvNet+EP, the \DIFadd{training} time was about 46 hours and 13 hours, respectively, for 15 super layers.
Since the iterative reconstruction modules are relatively computationally expensive, we chose 15 super layers to balance reconstruction quality and computational time.
The training hyper-parameters for the CNN part of FBPConvNet were set as follows for the various models: the learning rate decreased logarithmically from 0.001 to 0.0001; the batchsize was 1; and the momentum was 0.99. 
We initialized the filters in the various networks during training with i.i.d. random Gaussian entries with zero mean and variance 0.005, and initialized the bias with zero vectors.

\begin{figure*}[htb]
	\centering
	\begin{overpic}[scale=0.305]{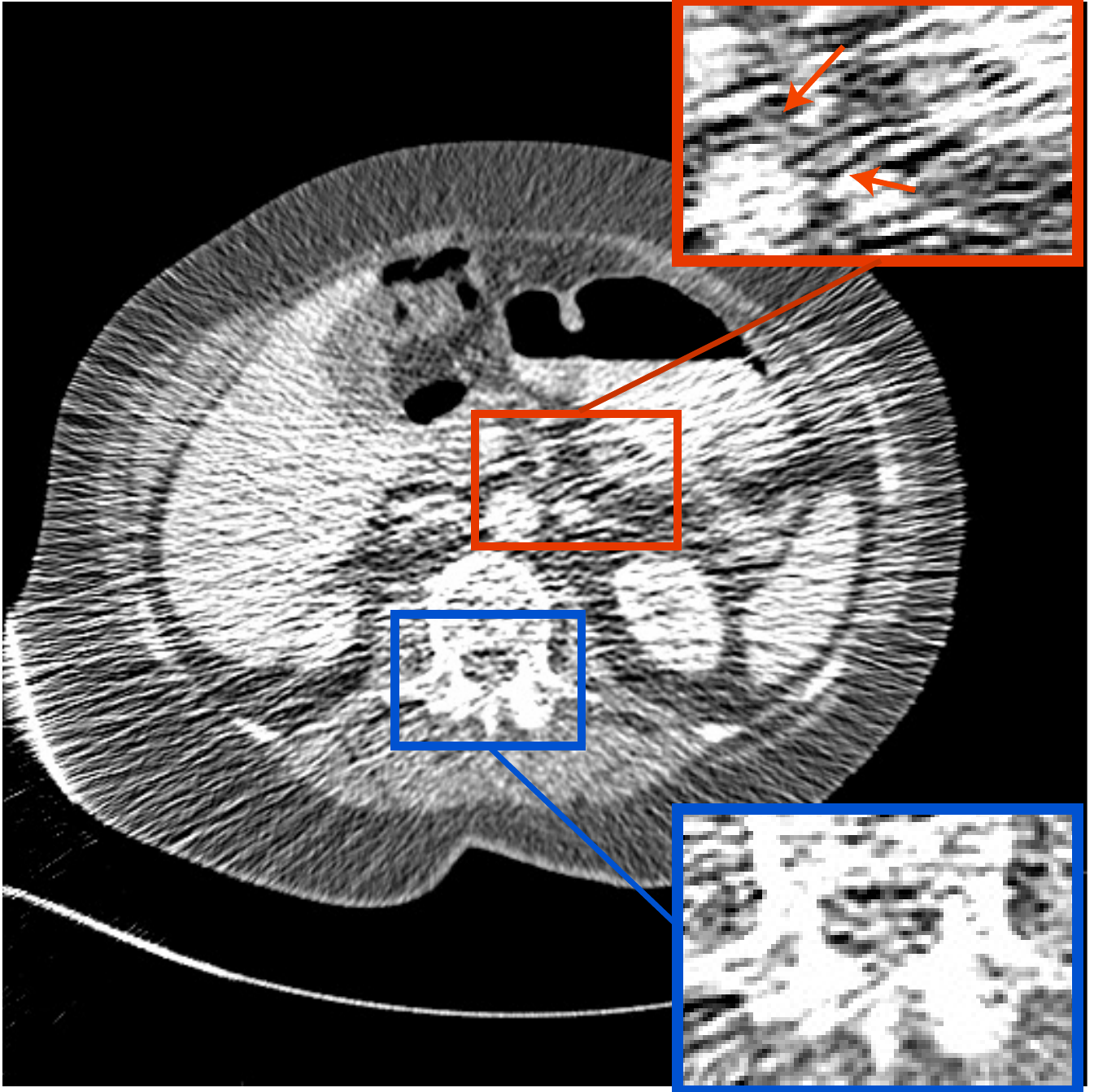} 
	\put(5,90){ \color{white}{\bf \large{PSNR:13.3 }}} 	\end{overpic}
	\begin{overpic}[scale=0.305]{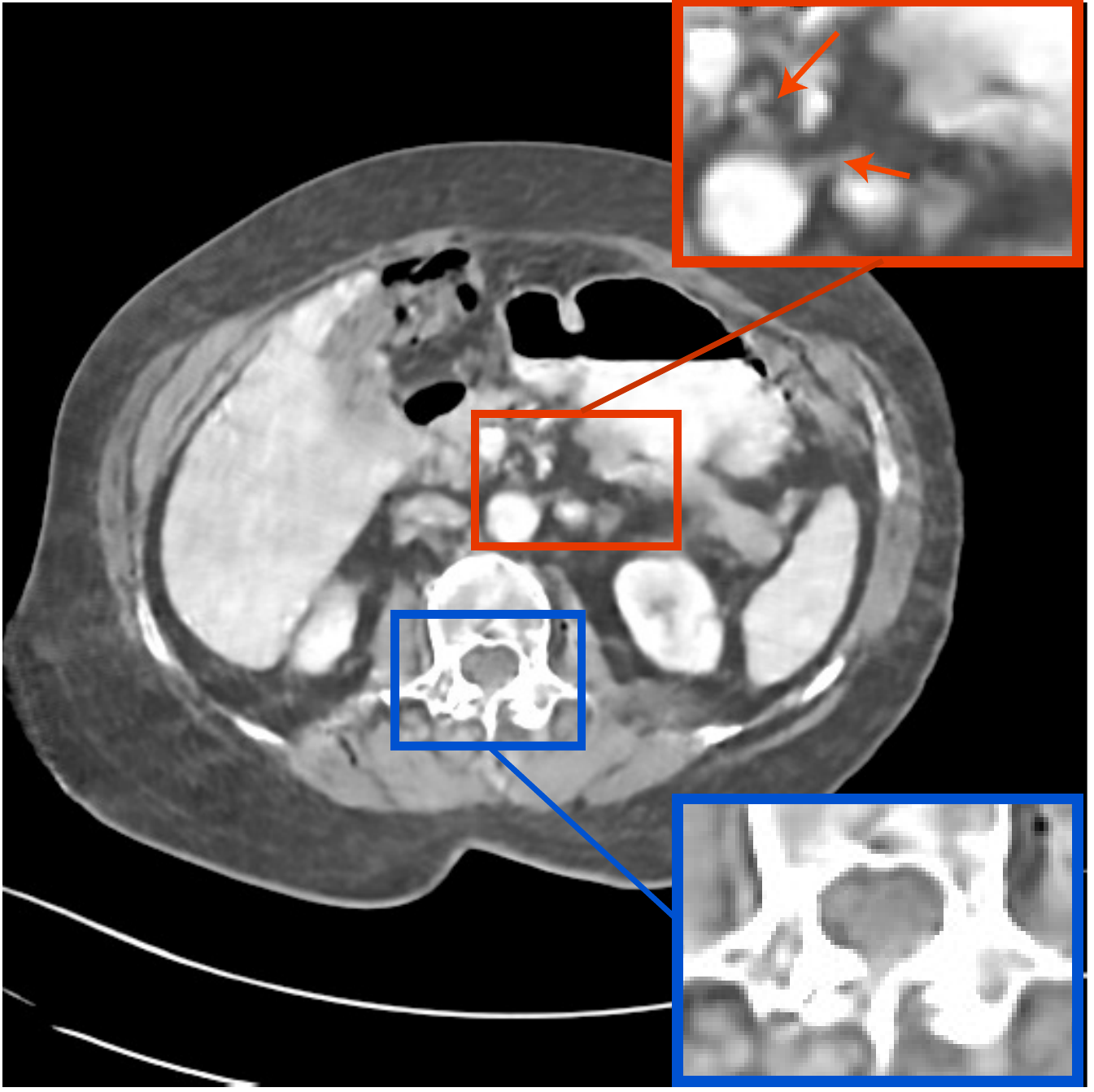} 
		\put(5,90){ \color{white}{\bf \large{PSNR:20.3 }}} 	\end{overpic}
	\begin{overpic}[scale=0.305]{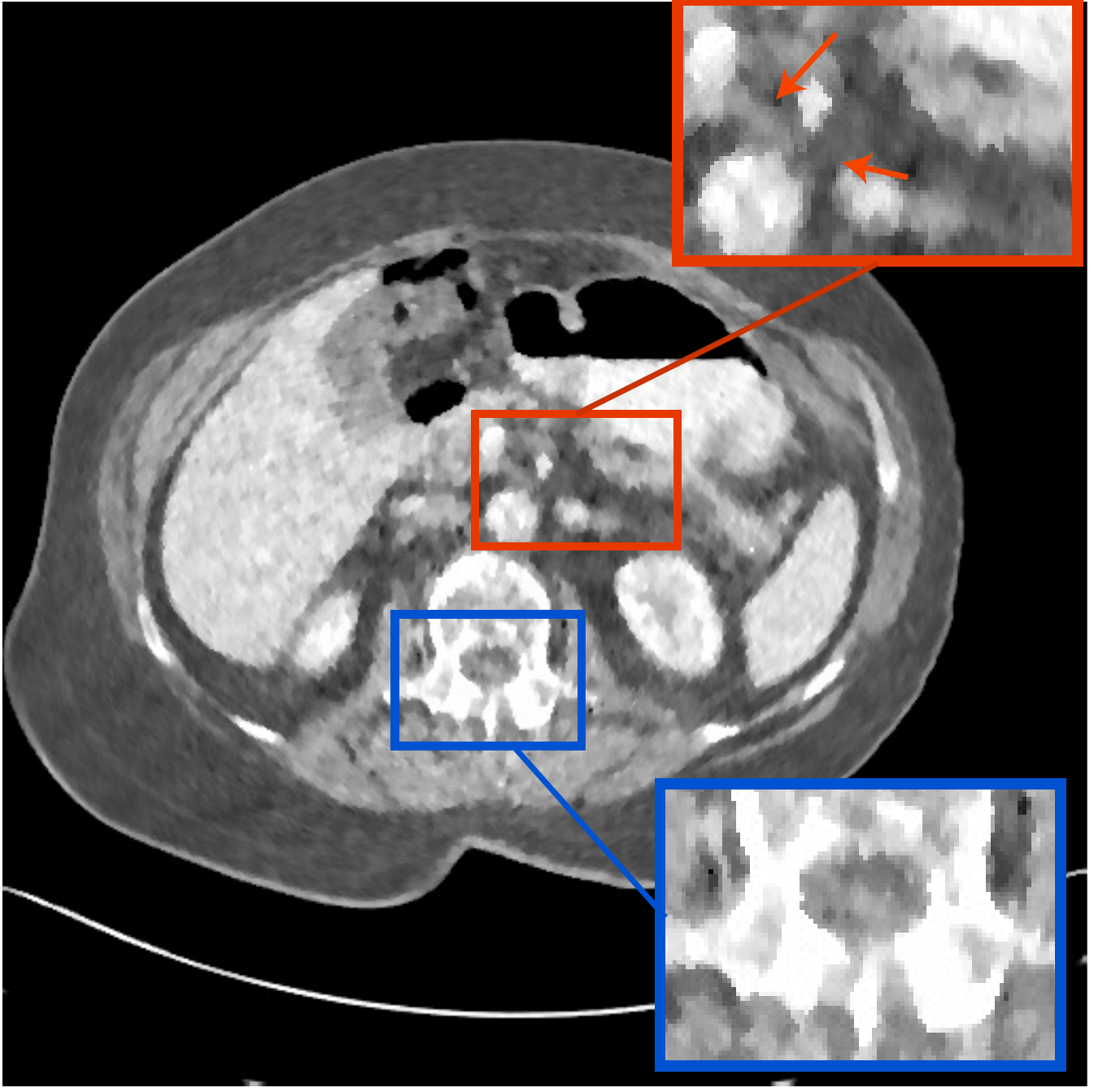} 
		\put(5,90){ \color{white}{\bf \large{PSNR:23.4 }}} 	\end{overpic}
	\begin{overpic}[scale=0.305]{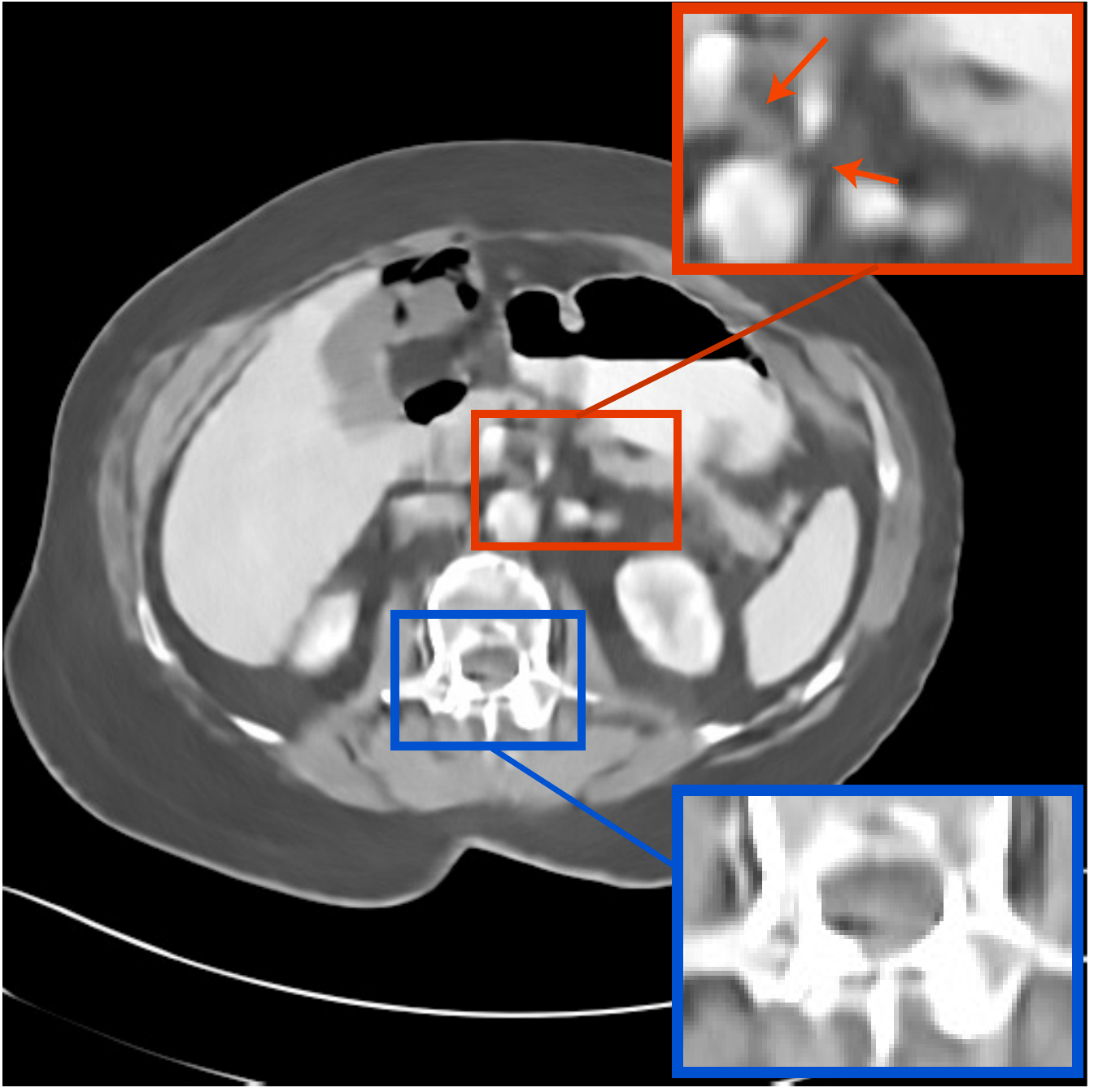} 
		\put(5,90){ \color{white}{\bf \large{PSNR:27.1 }}} 	\end{overpic} \\
	\begin{overpic}[scale=0.305]{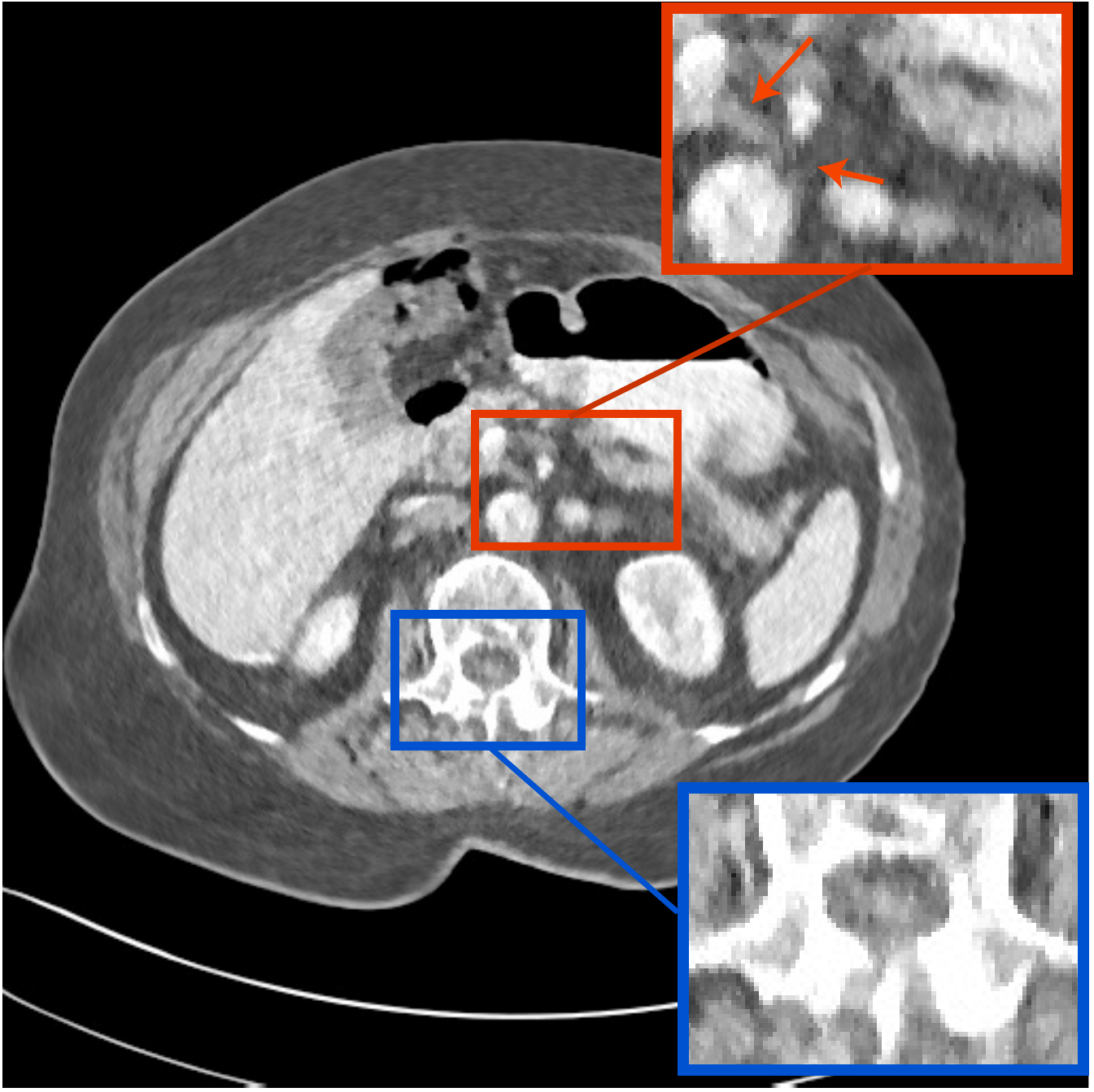} 
		\put(5,90){ \color{white}{\bf \large{PSNR: \color{white}{29.0}}}} 	\end{overpic}
	\begin{overpic}[scale=0.305]{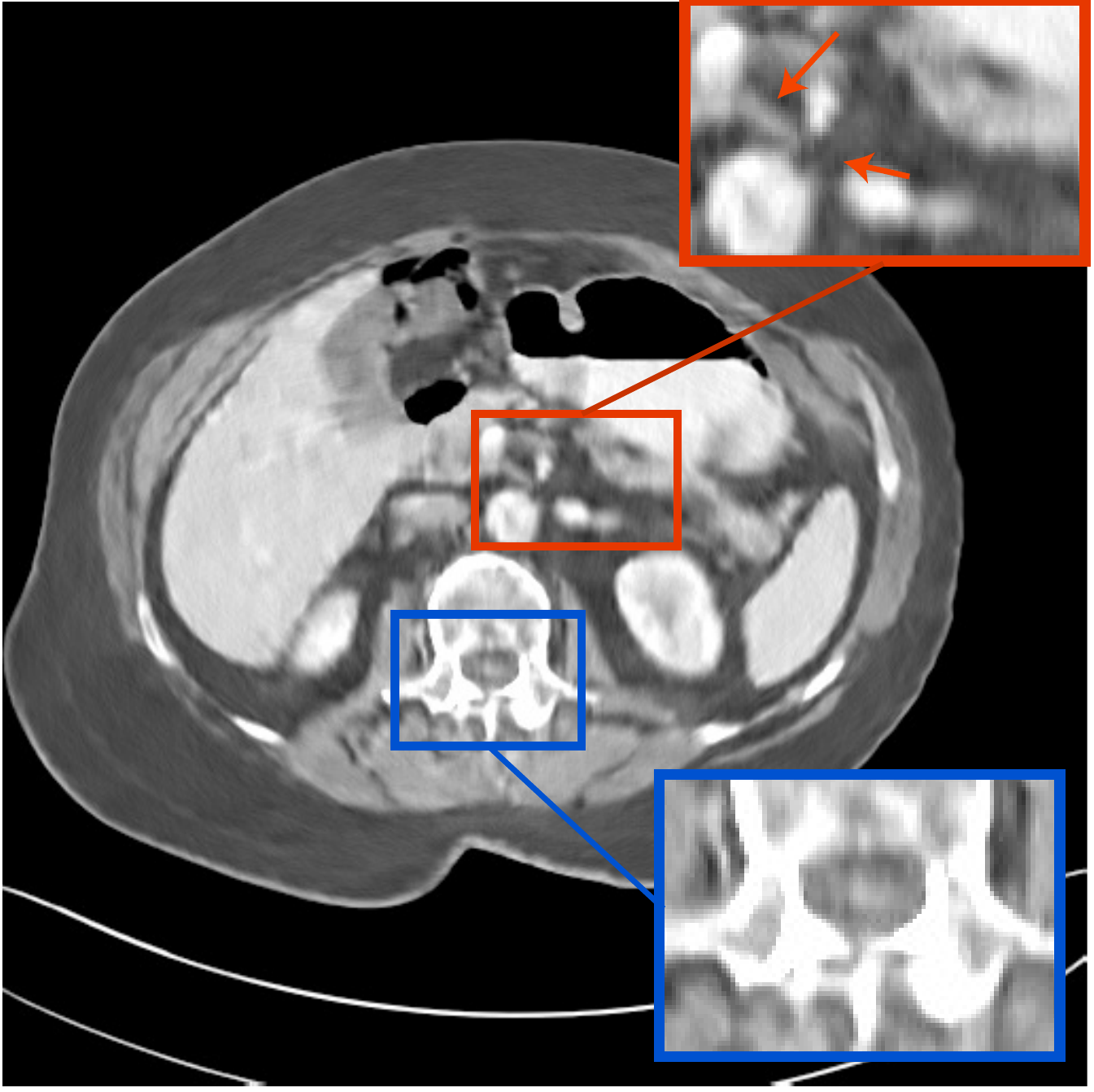} 
		\put(5,90){ \color{white}{\bf \large{PSNR: 31.7}}} 	\end{overpic}
	\begin{overpic}[scale=0.305]{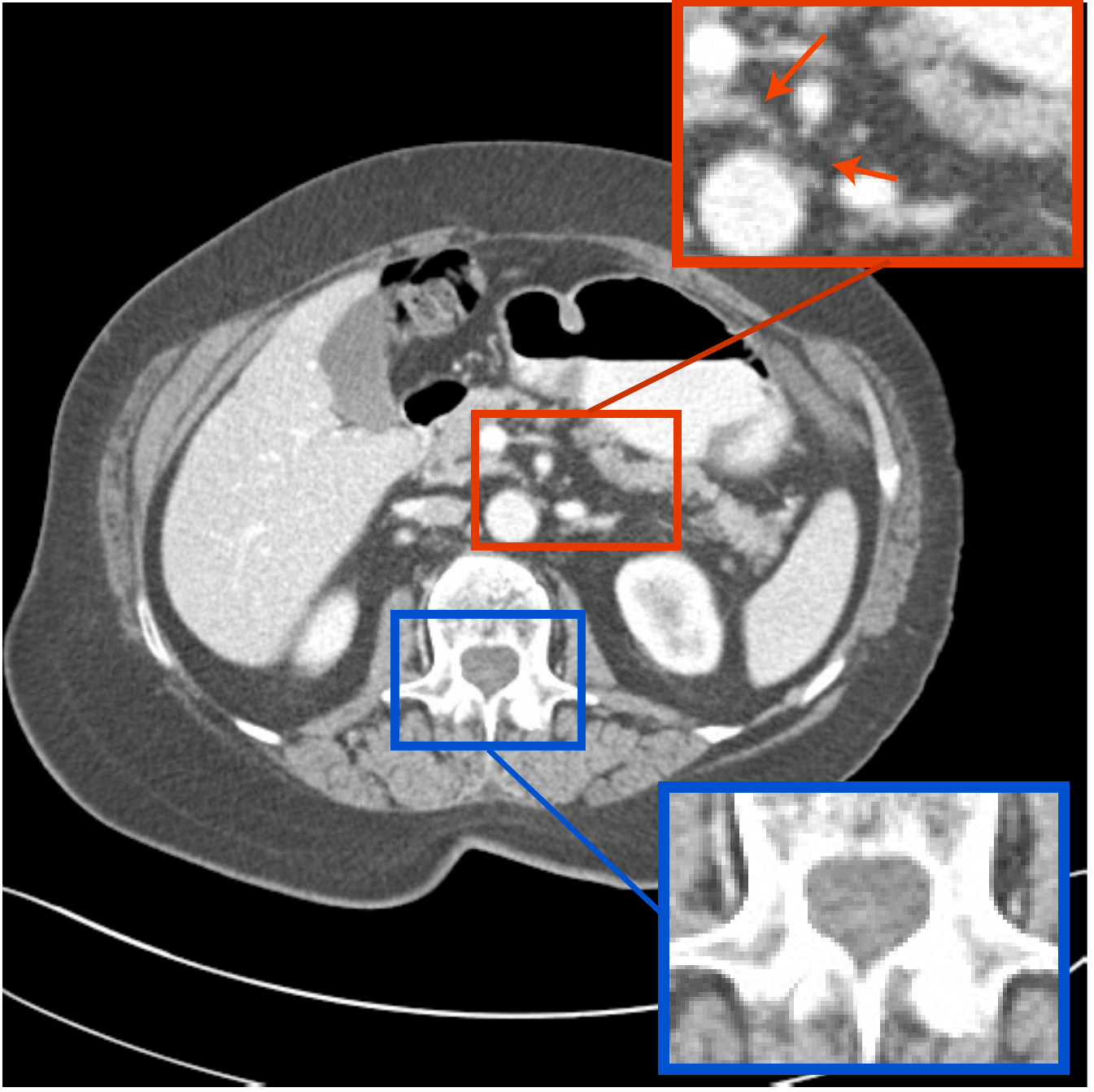} 
		\put(5,90){ \color{white}{\bf \large{}}} 	\end{overpic}
		\vspace{-0.0in}
	\caption{Reconstructed testing image (Test \#4, from patient L096) obtained by FBP, FBPConvNet, PWLS-EP, PWLS-ULTRA, FBPConvNet + EP, SUPER-ULTRA, and regular-dose FBP. The display window is [800 1200]~HU.}
	\label{fig:L096Slice170_image}
\end{figure*}

For training FBPConvNet+EP, we ran 10 epochs of FBPConvNet training (using ADAM) in each super layer to ensure that the learned 
networks capture layer-wise features. 
For the constituent PWLS-EP modules,  we ran 4 iterations of the relaxed OS-LALM algorithm with 4 subsets, and set $\delta=20$~HU and the regularization parameter $\beta=2^{15}$.
For training SUPER-ULTRA, we ran 10 epochs of FBPConvNet training in each super layer along with 4 outer iterations of PWLS-ULTRA with 5 inner iterations of the image update step that used the relaxed OS-LALM algorithm with 4 subsets. We set $\beta=5\times 10^3$ and $\gamma=20$ for the PWLS-ULTRA module.



\begin{figure}[!t]
	\centering
	\begin{overpic}[scale=0.19]{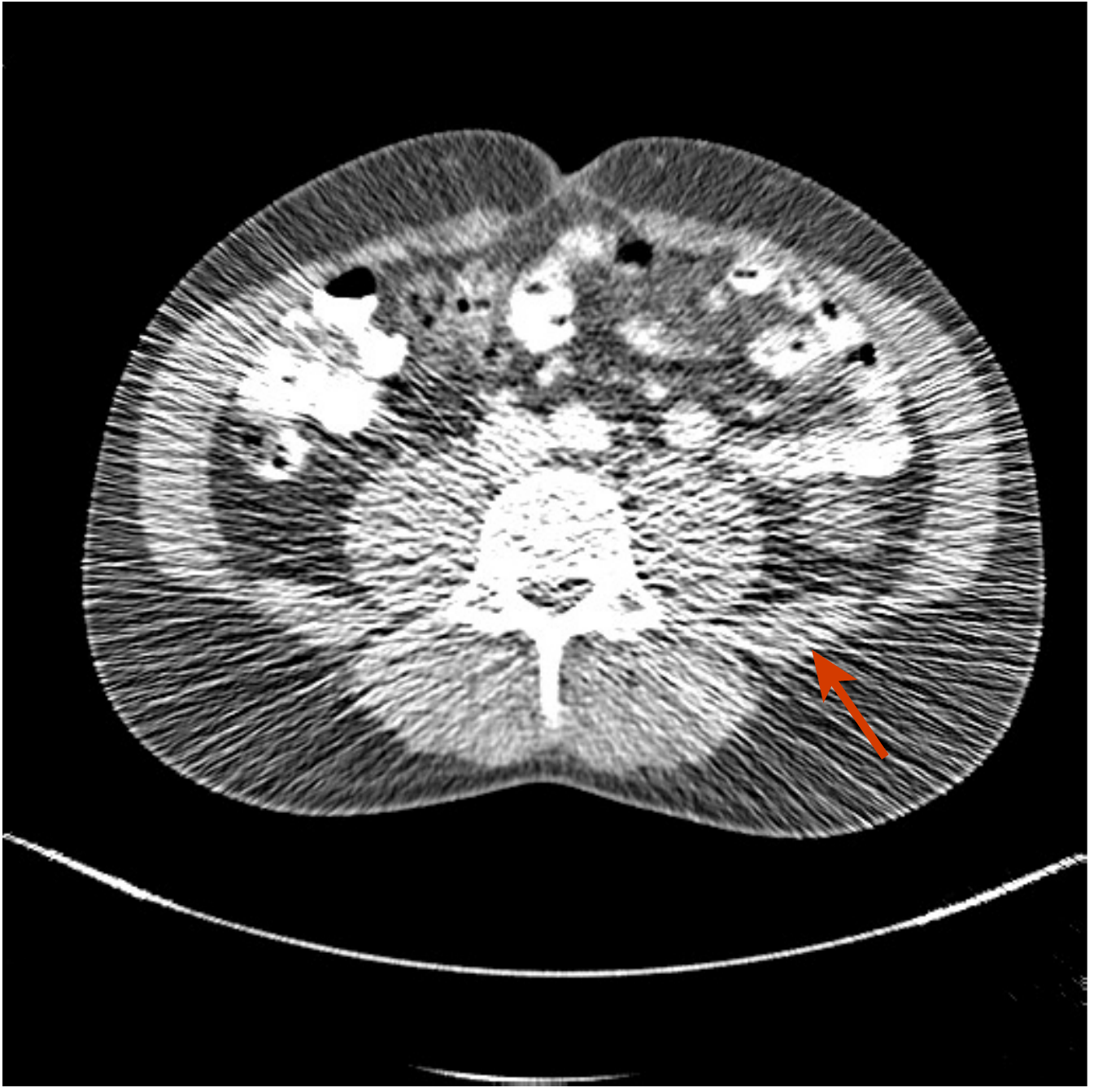} 
		\put(5,90){ \color{white}{\bf {PSNR:\color{white}{13.6}}} }	\end{overpic}
	\begin{overpic}[scale=0.19]{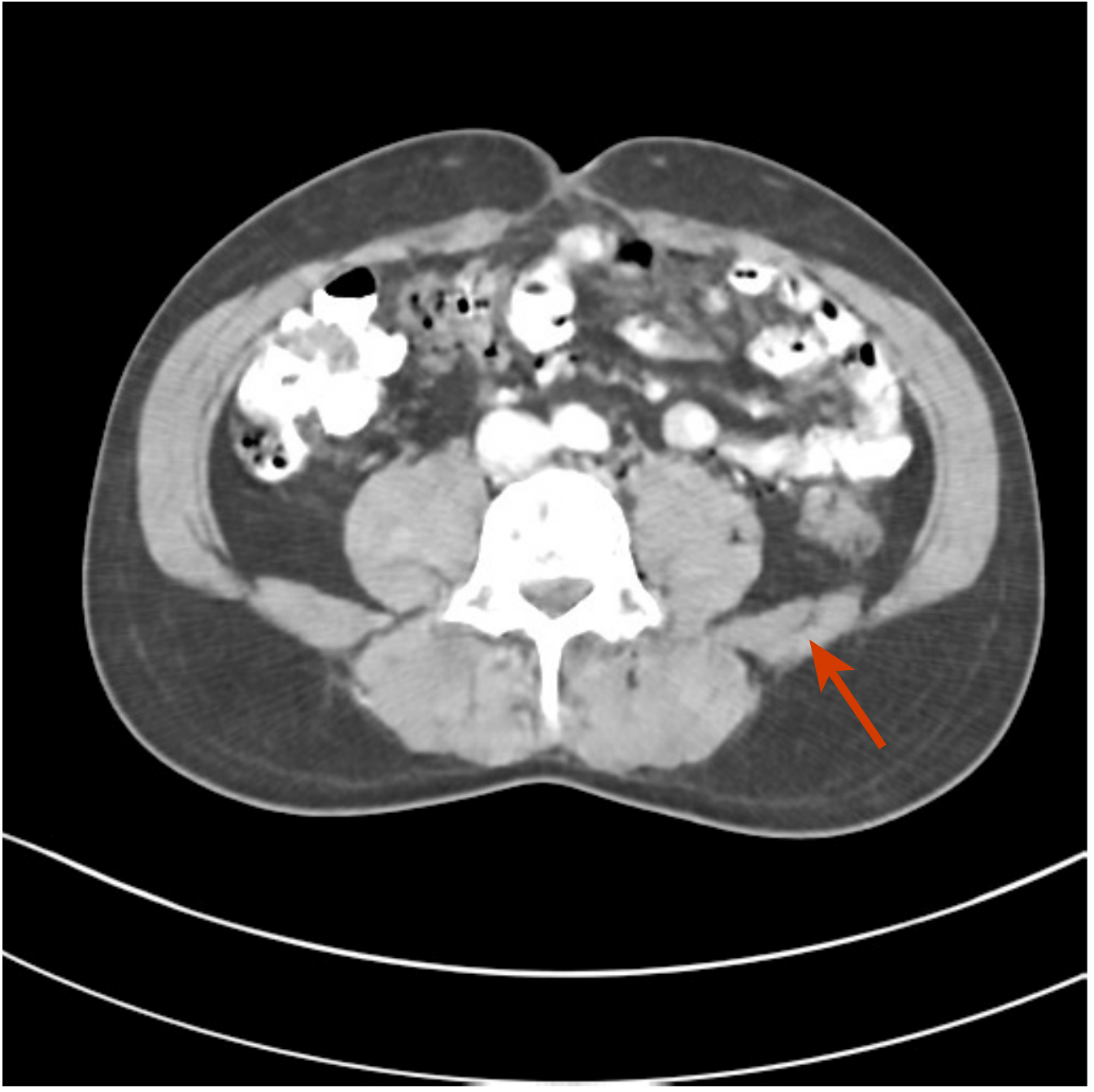} 
		\put(5,90){ \color{white}{\bf {PSNR: \color{white}{30.3}}} }	\end{overpic}
	\begin{overpic}[scale=0.19]{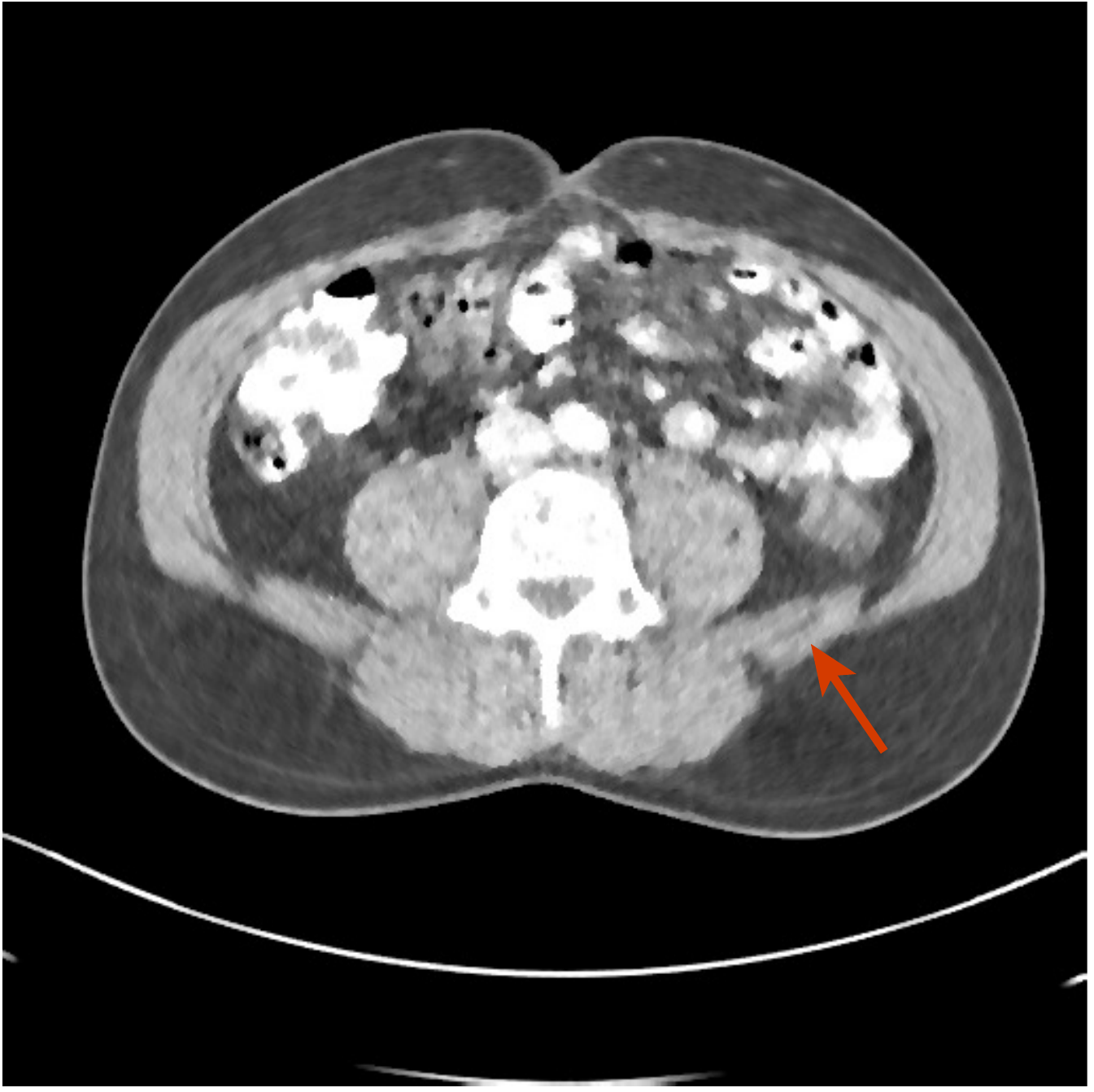} 
		\put(5,90){ \color{white}{\bf {PSNR: \color{white}{23.0}}}} 	\end{overpic}\\
	\begin{overpic}[scale=0.19]{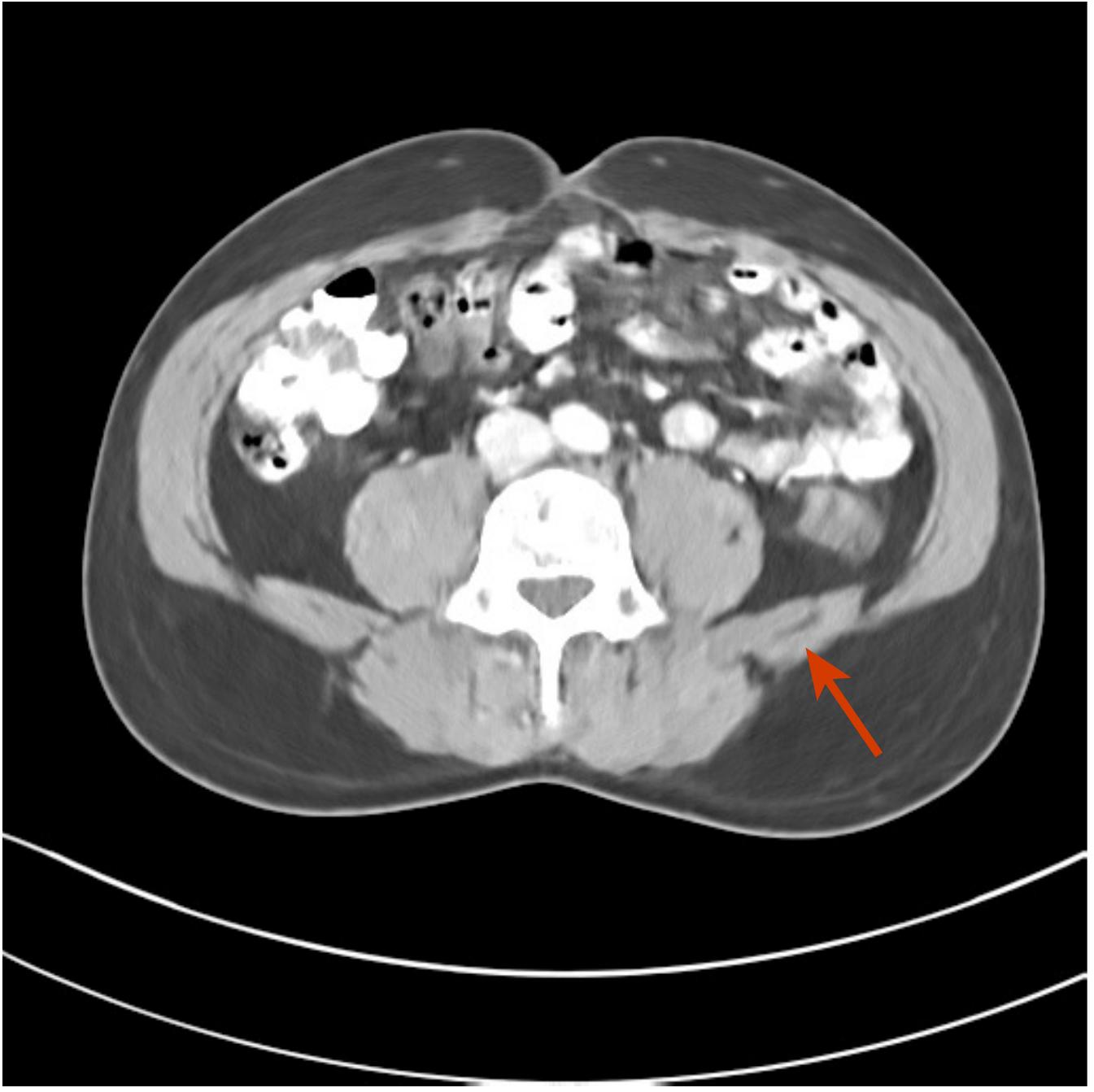} 
		\put(5,90){ \color{white}{\bf {PSNR: \color{white}{31.6}}}} 	\end{overpic}
	\begin{overpic}[scale=0.19]{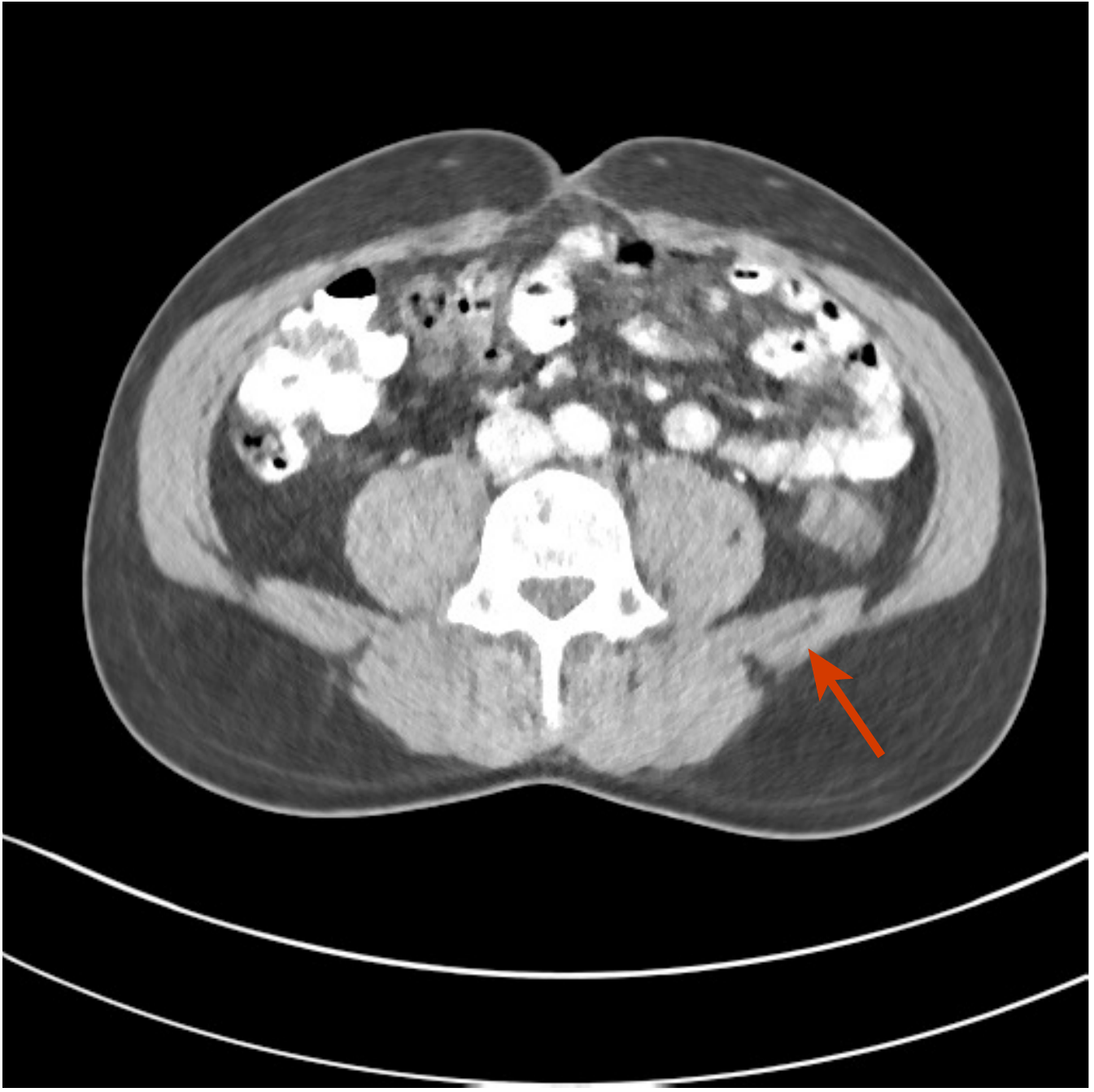} 
		\put(5,90){ \color{white}{\bf {PSNR: \color{white}{31.6}}}} 	\end{overpic}
	\begin{overpic}[scale=0.19]{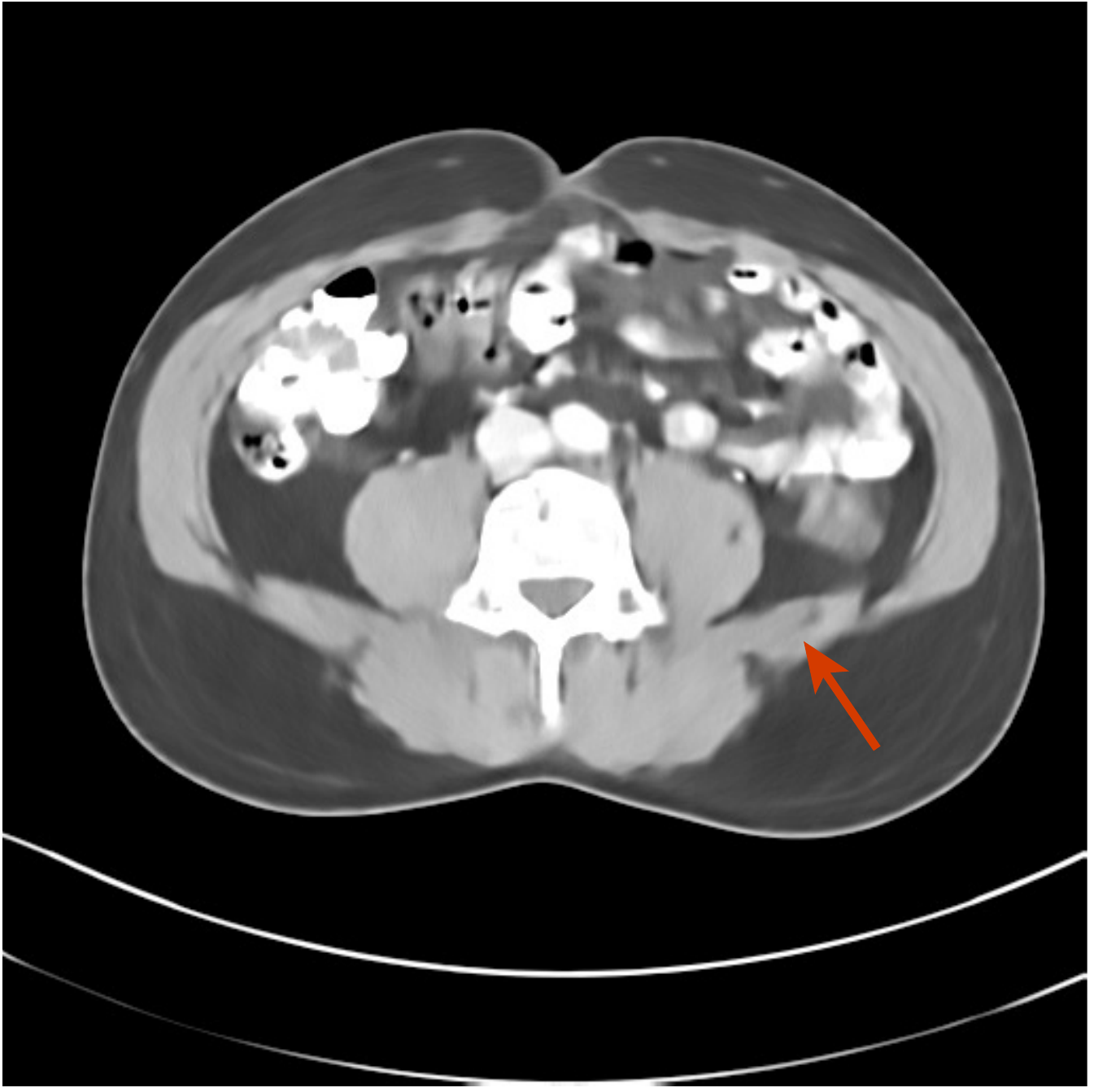} 
		\put(5,90){ \color{white}{\bf {PSNR: \color{white}{29.6}}}} 	\end{overpic} 
	\caption{Comparison of Test \#2 for FBP, FBPConvNet, PWLS-EP, PWLS-ULTRA, FBPConvNet + EP, and SUPER-ULTRA (clockwise top left). The display window is [800 1200]~HU.}
	\label{fig:L067Slice100_error}
	\vspace{-0.1in}
\end{figure}

We compared our learned models with the iterative schemes PWLS-EP and PWLS-ULTRA.
Since FBPConvNet+EP and SUPER-ULTRA already include the learned FBPConvNet modules, we used smaller regularization parameters for them compared to the usual or stand-alone PWLS-EP and PWLS-ULTRA, which worked well in our experiments.
For the stand-alone PWLS-EP iterative reconstruction algorithm that solves a convex problem, $\beta$ and $\delta$ were set as $2^{16}$ and $20$, respectively, and we ran 100 iterations of the OS-LALM algorithm to ensure convergence.
For the stand-alone PWLS-ULTRA, wherein the optimization problem is nonconvex, we set $\beta$ and $\gamma$ as $10^4$ and $25$, respectively, and ran $1000$ iterations of the alternating algorithm to ensure convergence. The above parameters provided optimal image quality in our experiments. Both PWLS-EP and PWLS-ULTRA were initialized with the simple FBP reconstructions.

\vspace{-0.01in}
\subsection{Evaluation Metrics}
\vspace{-0.01in}
To quantitatively evaluate the performances of the various reconstruction models, we chose three classic metrics, namely RMSE, peak signal to noise ratio (PSNR), and structural similarity index measure (SSIM)~\cite{wang:04:iqa}.
RMSE in Hounsfield units (HU) is defined as RMSE=$\sqrt{\sum_{i=1}^{N_p} \left(\hat{x}_i-x^*_i\right)^2/N_p}$, where $\x^*$ is the reference regular-dose CT image, $\hat{\x}$ is the reconstruction, and $N_p$ is the number of pixels.
We computed PSNR in decibels (dB).

\subsection{Results and Comparisons}
\label{sec:exper}

\vspace{-0.0in}
\subsubsection{Visual Results}
\vspace{-0.02in}
We applied the various methods
to six testing slices (three slices from L067 and three slices from L096) indexed as Test~\#1 to Test~\#6.
Figs.~\ref{fig:L096Slice170_image} and~\ref{fig:L067Slice100_error} show the \mbox{reconstructions} of
Test~\#4 (from patient L096) and Test~\#2 (from patient L067), respectively, using different methods.
Clearly, the traditional FBP reconstruction shows severe artifacts, while the results obtained by the other methods have much better image quality.
The stand-alone optimal PWLS-EP reconstruction still contains noise such as in the central soft-tissue area (Fig.~\ref{fig:L096Slice170_image}).
The (unsupervised) learning-based PWLS-ULTRA removes noise, but the edges of soft-tissues are more blurry in the results.
The supervised learning-based FBPConvNet achieves better trade-off between resolution and noise compared to PWLS-EP and PWLS-ULTRA, but many small structures were missed or distorted.
With the same training data as FBPConvNet, the learned FBPConvNet+EP and SUPER-ULTRA models provide much better reconstructions.
However, FBPConvNet+EP suffers from some streak artifacts in the central area (Fig.~\ref{fig:L096Slice170_image}) \DIFadd{as well as noise generally, and} SUPER-ULTRA mitigates these artifacts.
In both Figs.~\ref{fig:L096Slice170_image} and~\ref{fig:L067Slice100_error}, SUPER-ULTRA that combines supervised learned networks and transform learning-based iterative reconstructions achieved the best overall visual quality.
\footnote{\textcolor{black}{Additional comparisons between reconstructions for the other testing slices are included in the supplement.}}
\vspace{-0.15in}
\subsubsection{Quantitative Results}

\begin{figure}[!t]
		\includegraphics[width = 0.235\textwidth]{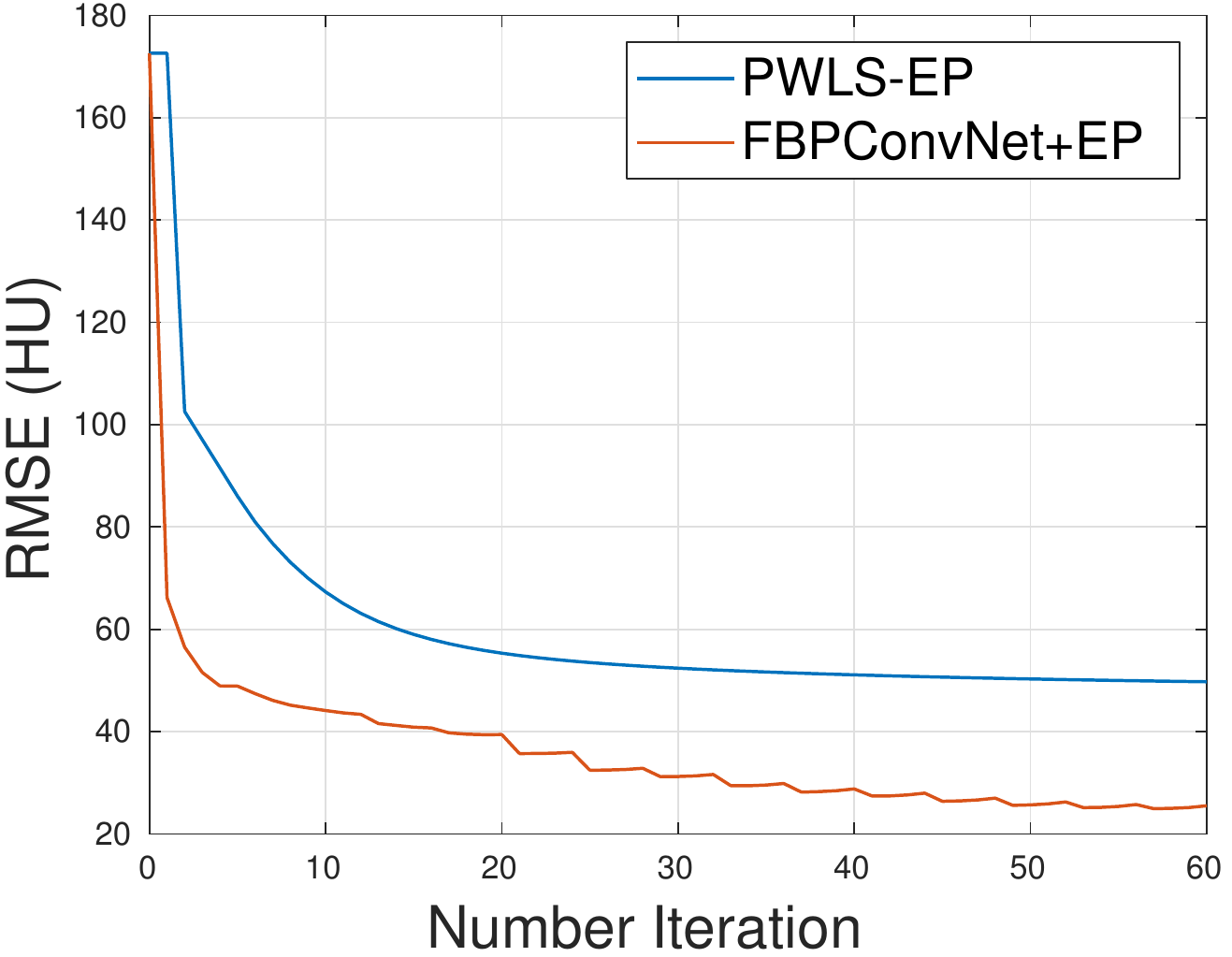}
		\includegraphics[width = 0.23\textwidth]{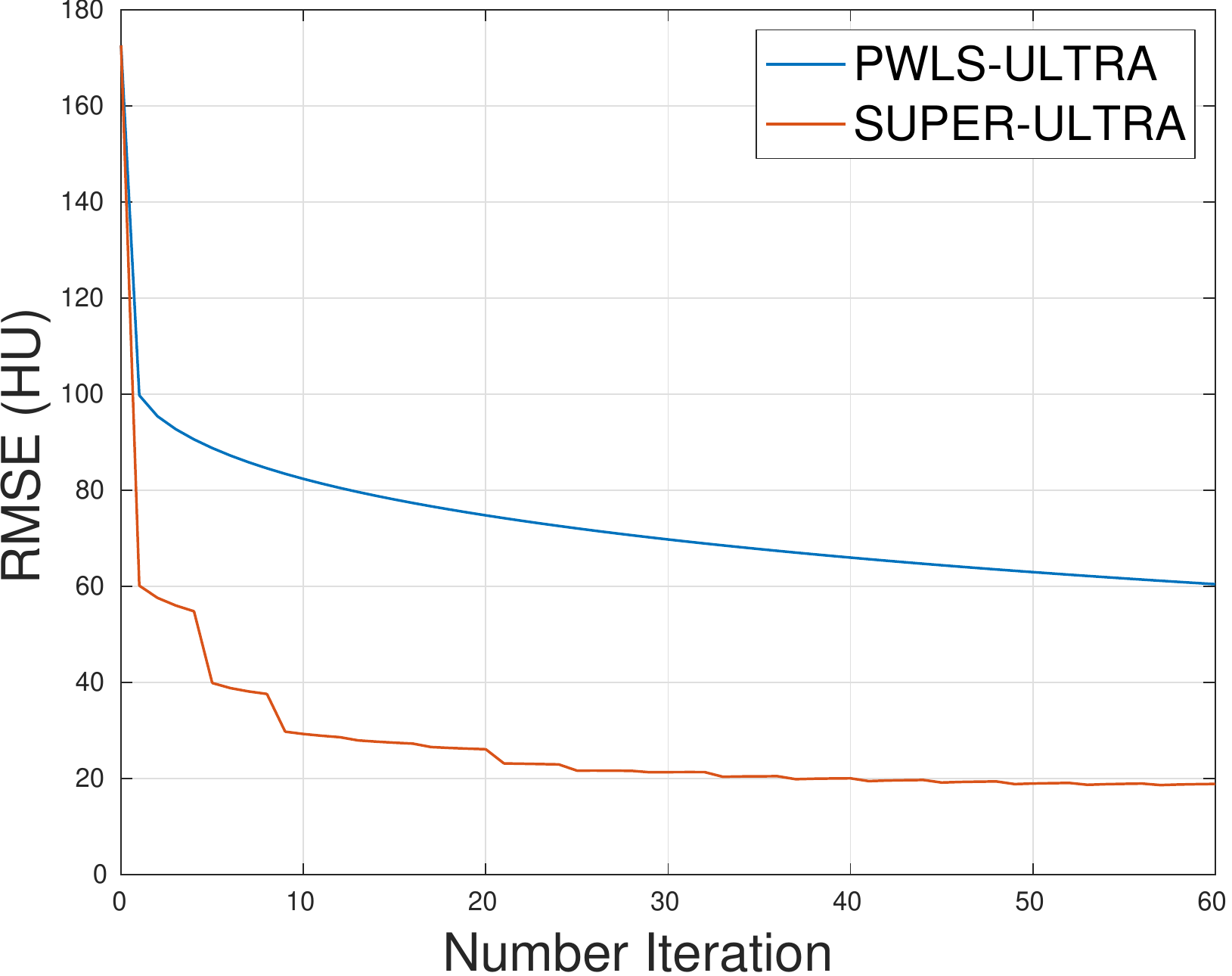} \\
		\includegraphics[width = 0.235\textwidth]{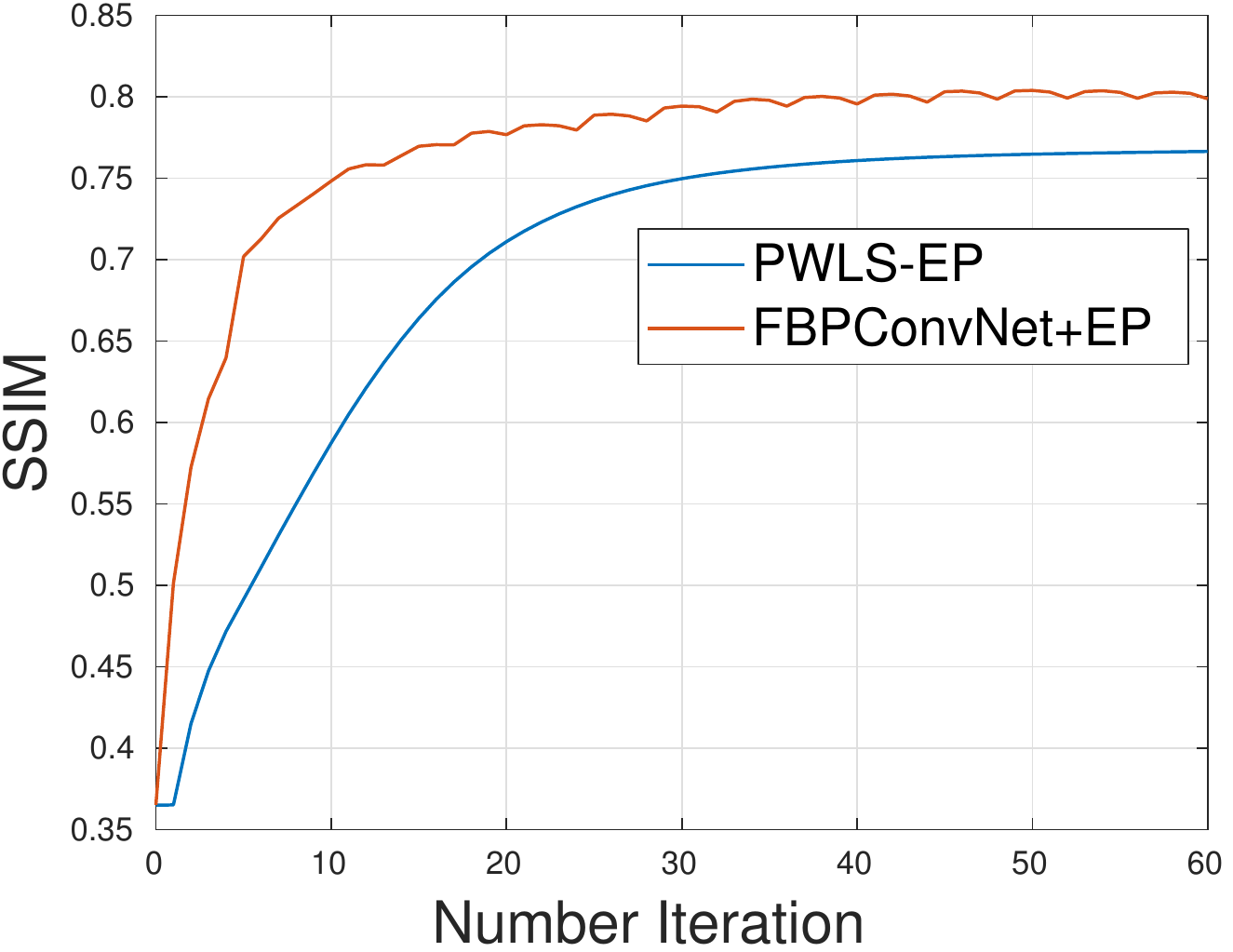}
		\includegraphics[width = 0.23\textwidth]{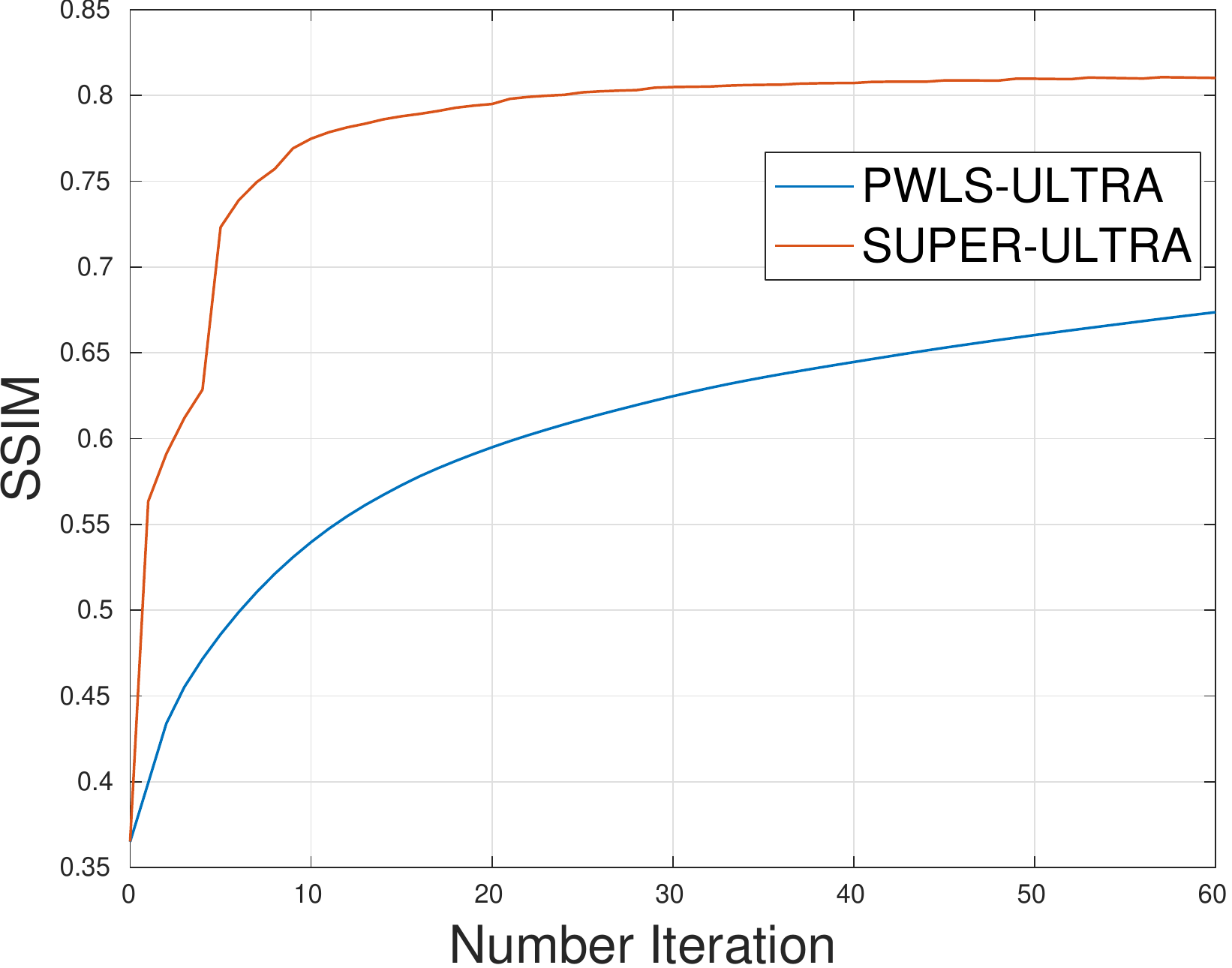}
		\vspace{-0.05in}
	\caption{RMSE (first row) and SSIM (second row) values for Test \#4 over the iterations of PWLS-EP, \DIFadd{FBPConvNet+EP}, PWLS-ULTRA, and SUPER-ULTRA.}
	\label{fig:L067Slice50_rmsessim_curve}
	\vspace{-0.0in}
\end{figure}

Fig.~\ref{fig:L067Slice50_rmsessim_curve} shows the RMSE and SSIM evolution for the stand-alone PWLS-EP and PWLS-ULTRA along with those for FBPConvNet+EP and SUPER-ULTRA.
The latter two involve 15 super layers with 4 (outer) iterations in the iterative module per layer, and thus, the RMSE and SSIM evolution is plotted over these individual iterations.
For PWLS-EP and PWLS-ULTRA, we show the evolution of the metrics over $60$ iterations.
FBPConvNet+EP and SUPER-ULTRA clearly achieve much lower RMSE values and higher SSIM values over layers than PWLS-EP and PWLS-ULTRA. The plots also show \DIFadd{faster} convergence for the SUPER models. 


\begin{table*}[htb]
	\centering  
	\caption{PSNR, RMSE, and SSIM of reconstructed test images for different methods.} 
	\vspace{-0.03in}
	\begin{tabular}{|c|c|c|c|c|c|c|c|c|} 
		\hline
		\multicolumn{3}{|c|}{} & FBP  & \tabincell{c}{FBPConvNet} & \tabincell{c}{PWLS-\\EP} & \tabincell{c}{PWLS-\\ULTRA} & \tabincell{c}{FBPConvNet+EP}  & \tabincell{c}{SUPER-ULTRA}\\
		\hline
		\multirow{9}{*}{L067} & \multirow{3}{*}{Test \#1} & PSNR & 11.0   & 29.8  & 23.5 &  29.3 & 30.5 & \textbf{30.9}\\
		\cline{3-9}
		& & RMSE & 245.5  & 26.2  & 54.1 & 28.3 & 24.5 &  \textbf{23.3}\\
		\cline{3-9}
		& & SSIM & 0.29  & 0.74  & 0.73 & 0.74 & 0.77 & \textbf{0.77}\\  \cline{2-9}
		& \multirow{3}{*}{Test \#2} & PSNR & 13.6   & 30.3  & 23.0 & 29.6 & 31.6 & \textbf{31.6}\\
		\cline{3-9}
		& & RMSE & 170.3  & 22.4  & 52.2 & 24.5 & 19.5 & \textbf{19.3}\\
		\cline{3-9}
		& & SSIM & 0.40  & 0.79  & 0.78 & 0.79 & 0.81 & \textbf{0.81}\\ 	
		\cline{2-9}
		& \multirow{3}{*}{Test \#3} & PSNR & 9.5  & 19.9  & 24.4 & 28.6 & \textbf{32.2}  & 32.1\\
		\cline{3-9}
		& & RMSE & 299.5  & 81.7  & 47.9 & 29.6 & \textbf{19.6}  & 19.8\\
		\cline{3-9}
		& & SSIM & 0.24  & 0.56  & 0.69 & 0.69 & 0.72 & \textbf{0.72}\\ 	\hline
		\multirow{9}{*}{L096} & \multirow{3}{*}{Test \#4} & PSNR & 13.3   & 20.3  & 23.4 & 27.1 & 29.0 & \textbf{31.7}\\
		\cline{3-9}
		& & RMSE & 172.6  & 70.7  & 48.7 & 31.9 & 25.5 & \textbf{18.9}\\
		\cline{3-9}
		& & SSIM & 0.37 & 0.67 & 0.77 & 0.79 & 0.80 & \textbf{0.81}\\  
		\cline{2-9}
		& \multirow{3}{*}{Test \#5} & PSNR & 9.3 & 29.7  & 23.3 & 25.7 & 30.6 & \textbf{30.8}\\
		\cline{3-9}
		& & RMSE & 304.4  & 25.8  & 53.8 & 40.6& 23.3 & \textbf{22.7}\\
		\cline{3-9}
		& & SSIM & 0.20  & 0.71  & 0.70 & 0.70 & 0.74 & \textbf{0.75}\\  \cline{2-9}
		& \multirow{3}{*}{Test \#6} & PSNR & 9.7   & 27.6  & 23.6 & 28.1 & 30.7 & \textbf{32.3}\\
		\cline{3-9}
		& & RMSE & 274.2  & 32.3  & 48.6 & 29.0 & 21.5 & \textbf{17.8}\\
		\cline{3-9}
		& & SSIM & 0.23  & 0.66  & 0.73 & 0.74 & 0.75 & \textbf{0.76}\\ \hline
	\end{tabular}  
	\label{Tab:snrrmsessim_comp}	 
	\vspace{-0.1in}
\end{table*}


Table.~\ref{Tab:snrrmsessim_comp} shows the RMSE, PSNR, and SSIM values with various methods for the testing slices.
The proposed SUPER-ULTRA \DIFadd{typically} achieves significant improvements in RMSE, SSIM, and PSNR over the other methods for all testing slices. 
Importantly, FBPConvNet+EP and SUPER-ULTRA perform much better than PWLS-EP and PWLS-ULTRA, respectively, and also provide more promising results than FBPConvNet, demonstrating that the combination of the supervised module and iterative modules in the SUPER model works well and outperforms the individual modules. 

\vspace{-0.0in}
\subsubsection{Visual Quality over SUPER layers}
\begin{figure}[!t]
	\centering
	\begin{overpic}[scale=0.295]{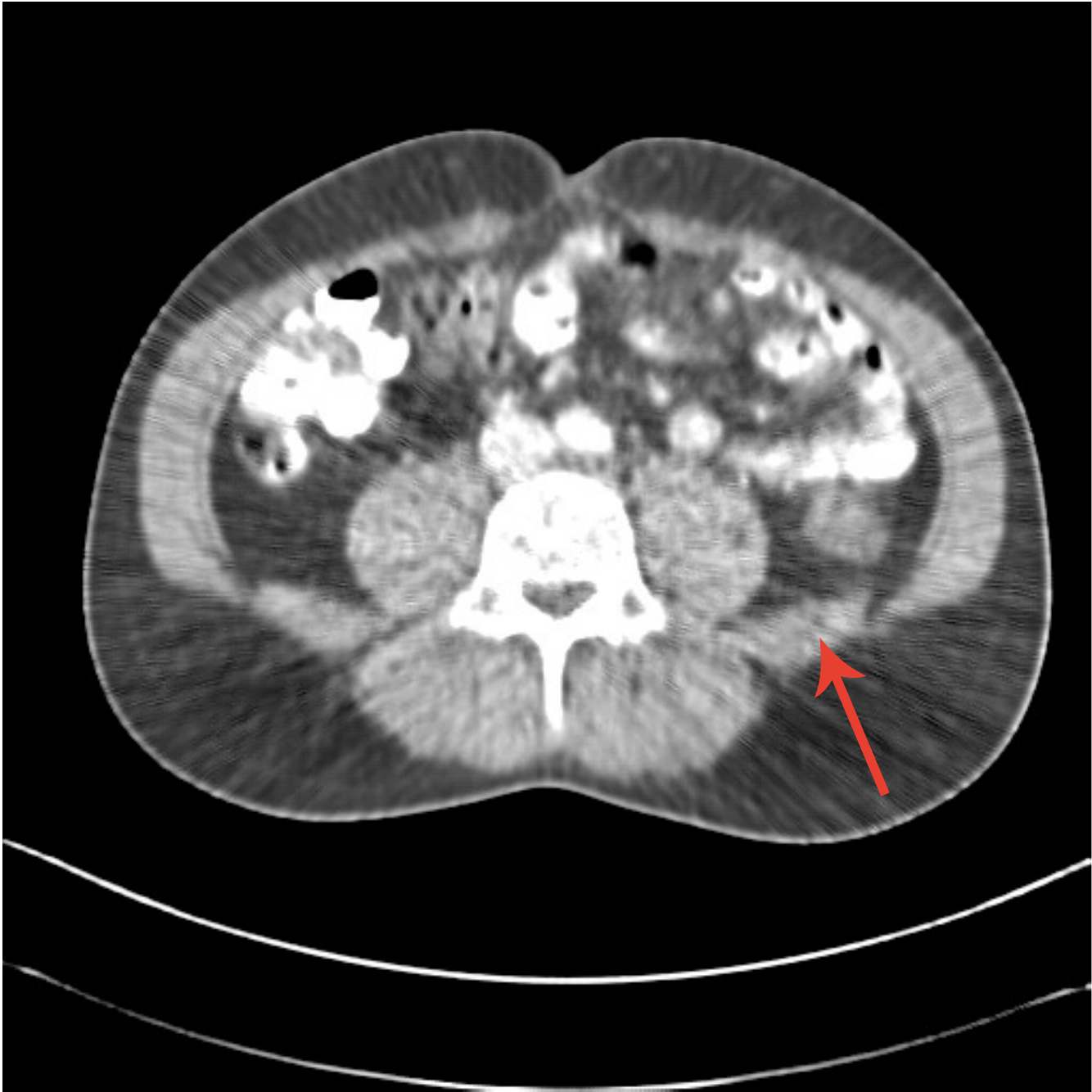}
	\put(20,90){ \color{white}{\bf \large{Super Layer 1}}} 	
	\end{overpic}
	\begin{overpic}[scale=0.295]{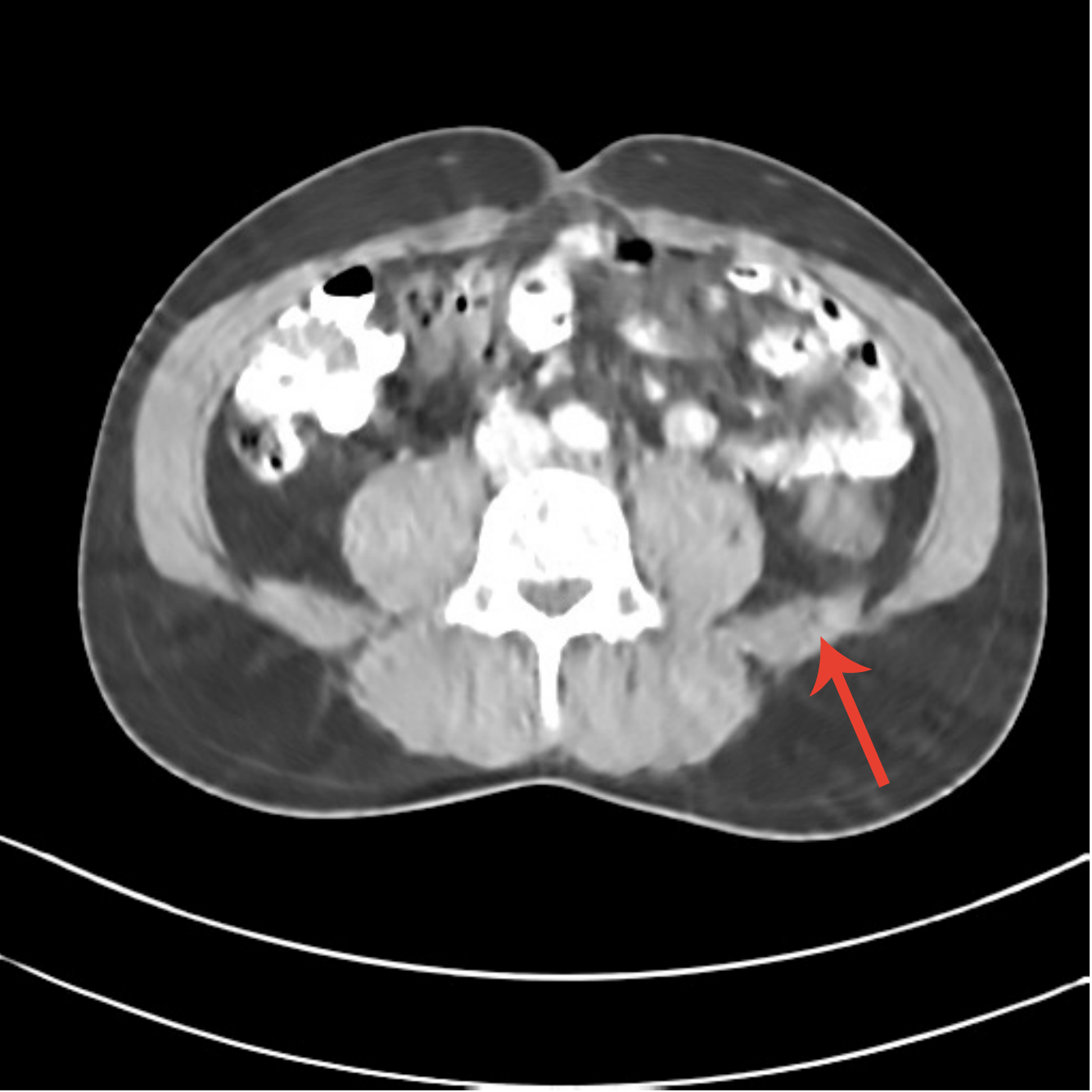}
		\put(20,90){ \color{white}{\bf \large{Super Layer 3}}} 	
	\end{overpic}  \\
	\begin{overpic}[scale=0.295]{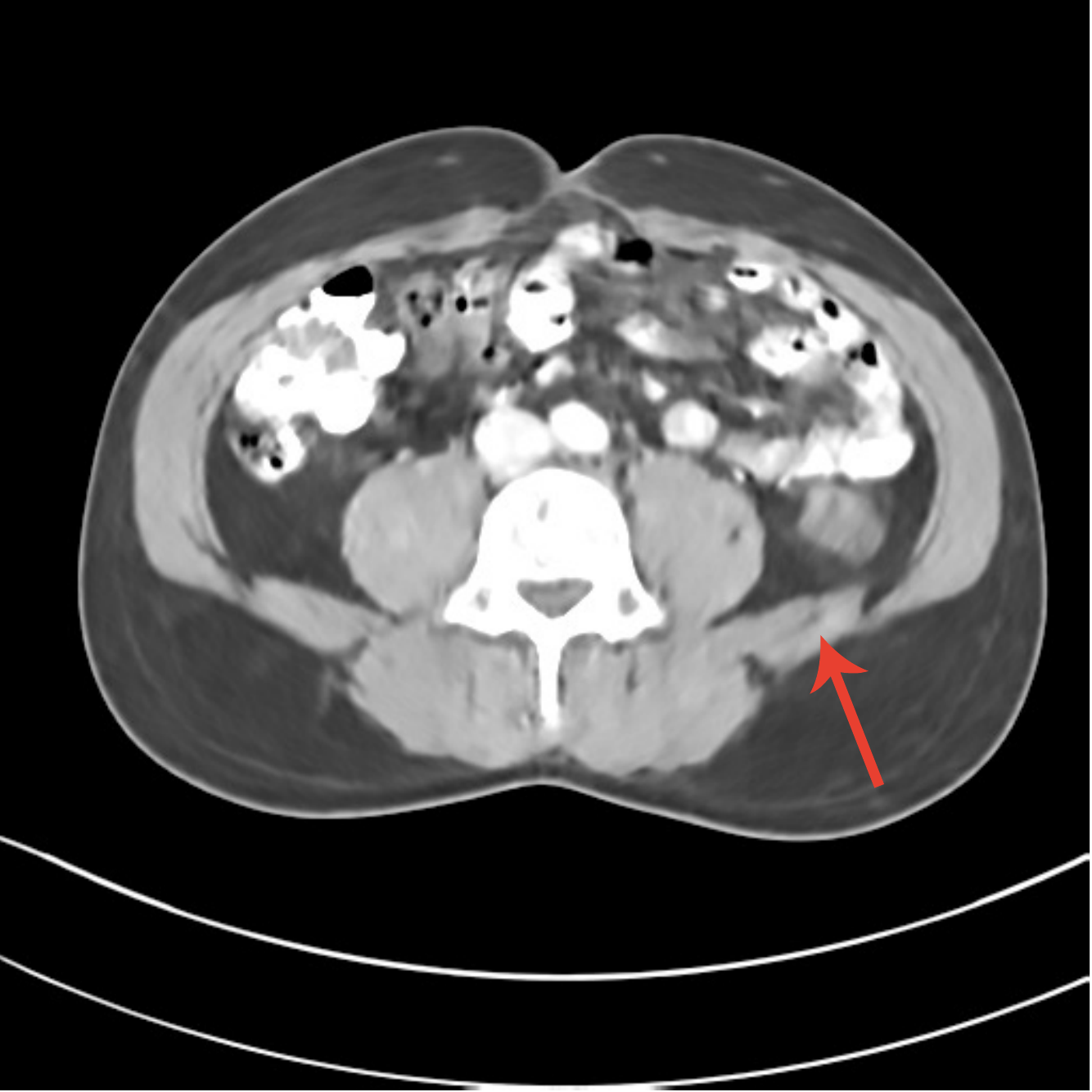}
		\put(20,90){ \color{white}{\bf \large{Super Layer 7}}} 	
	\end{overpic}
	\begin{overpic}[scale=0.295]{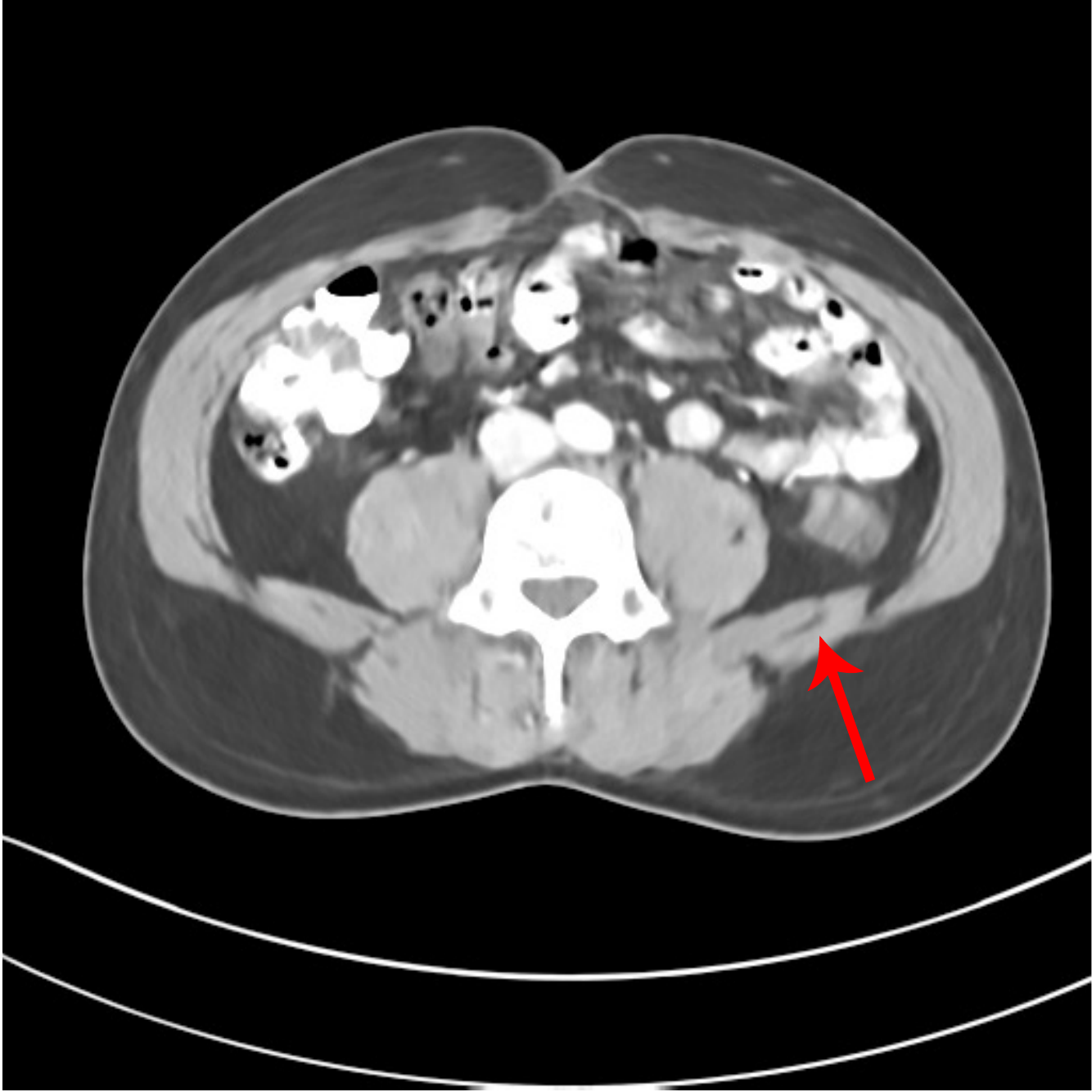}
		\put(17,90){ \color{white}{\bf \large{Super Layer 13}}} 	
	\end{overpic}
	\caption{Outputs of 1st, 3rd, 7th, and 13th SUPER-ULTRA layers with display window [800 1200]~HU. The reconstruction is visually converging.  }
	\label{fig:L067Slice100_layer_res}
	\vspace{-0.0in}
\end{figure}

\vspace{-0.01in}
Fig.~\ref{fig:L067Slice100_layer_res} shows the output of SUPER-ULTRA after different numbers of SUPER layers.
The initial FBP image and the final output (after $15$ layers) were shown in Fig.~\ref{fig:L067Slice100_error}. 
The first several layers of SUPER-ULTRA mainly remove severe noise and artifacts, while the later layers recover some structural details.

\vspace{-0.1in}
\subsubsection{Behavior of the ULTRA model in the SUPER architecture}
\vspace{-0.01in}
\begin{figure*}[!t]
	\centering
	\includegraphics[scale=0.35]{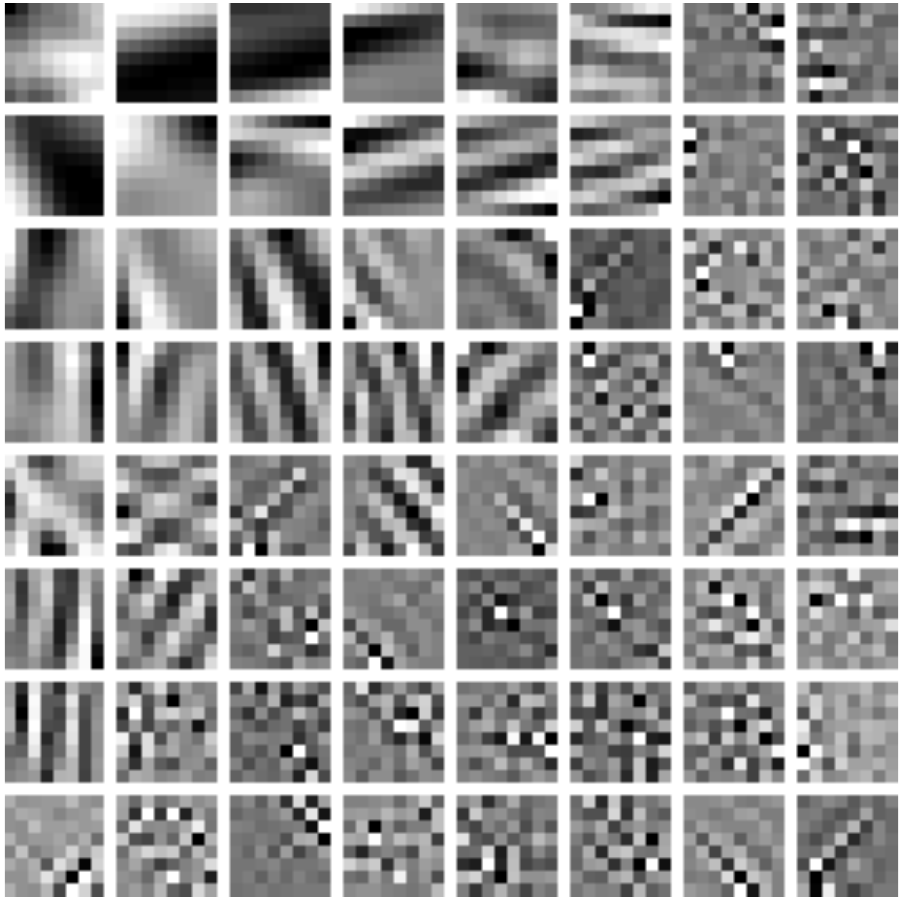}
	\includegraphics[scale=0.35]{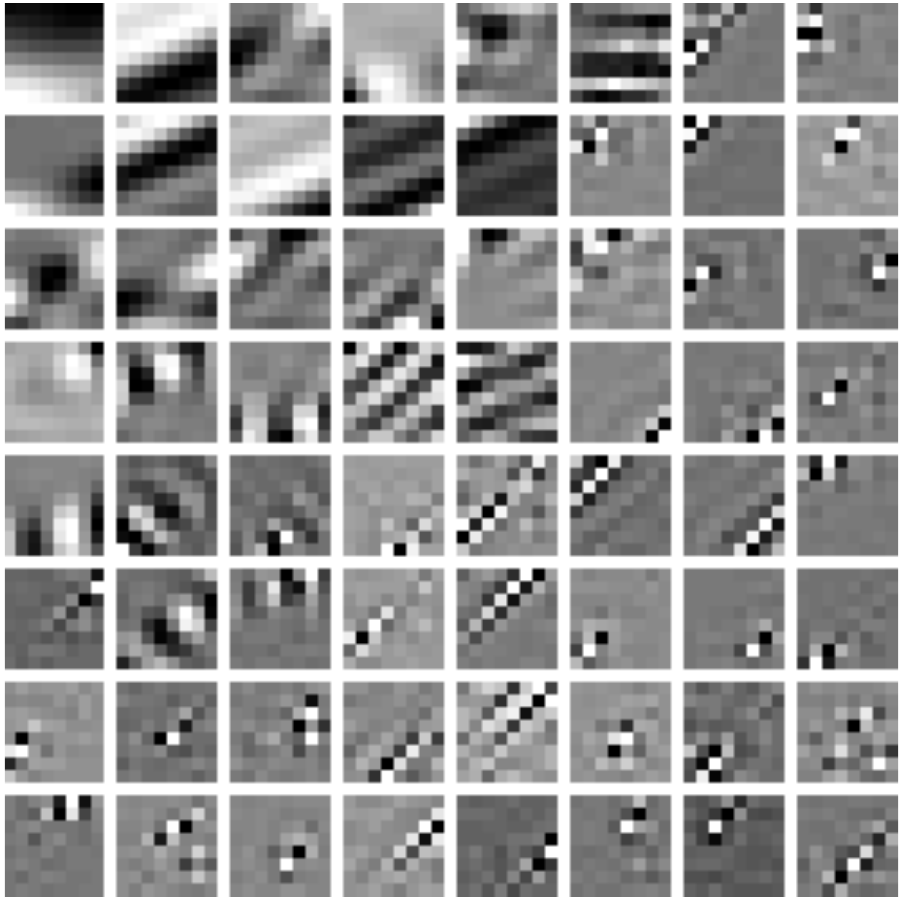}
	\includegraphics[scale=0.35]{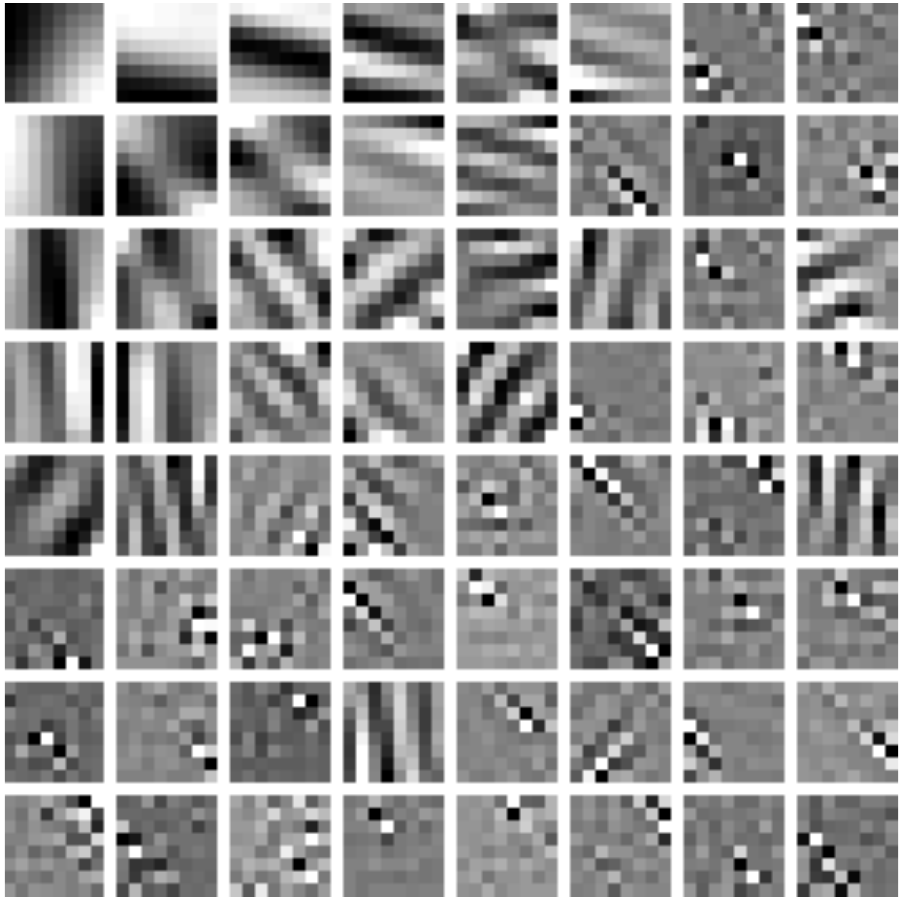}
	\includegraphics[scale=0.35]{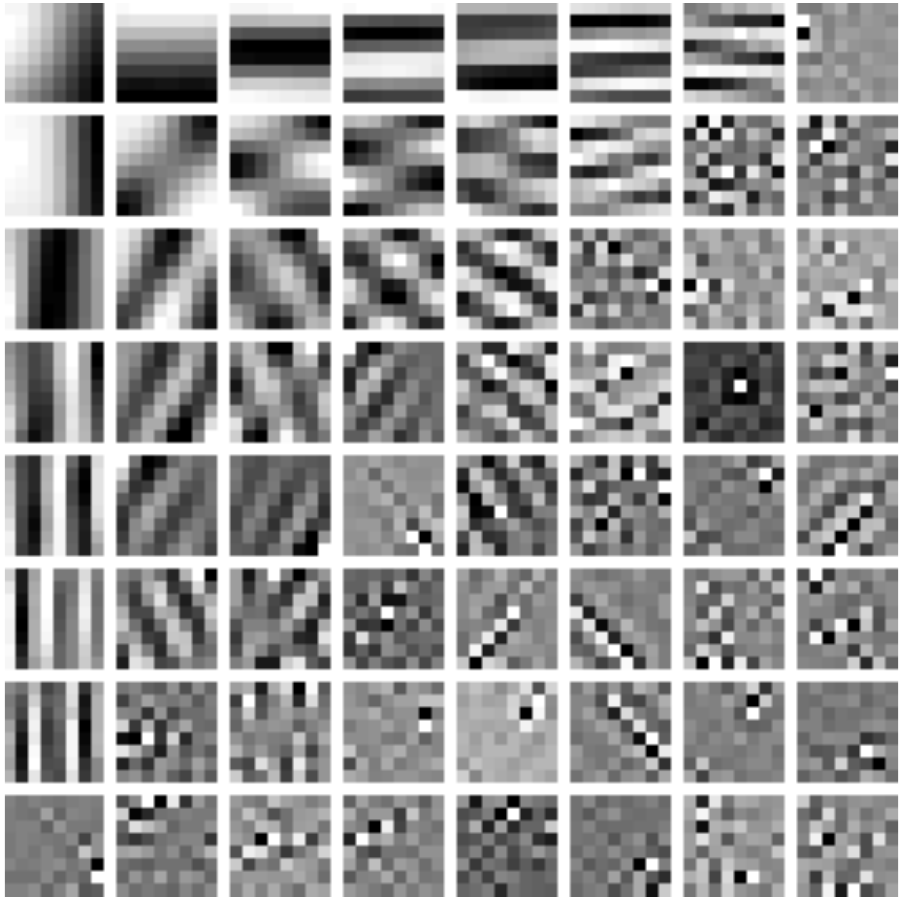}
	\includegraphics[scale=0.35]{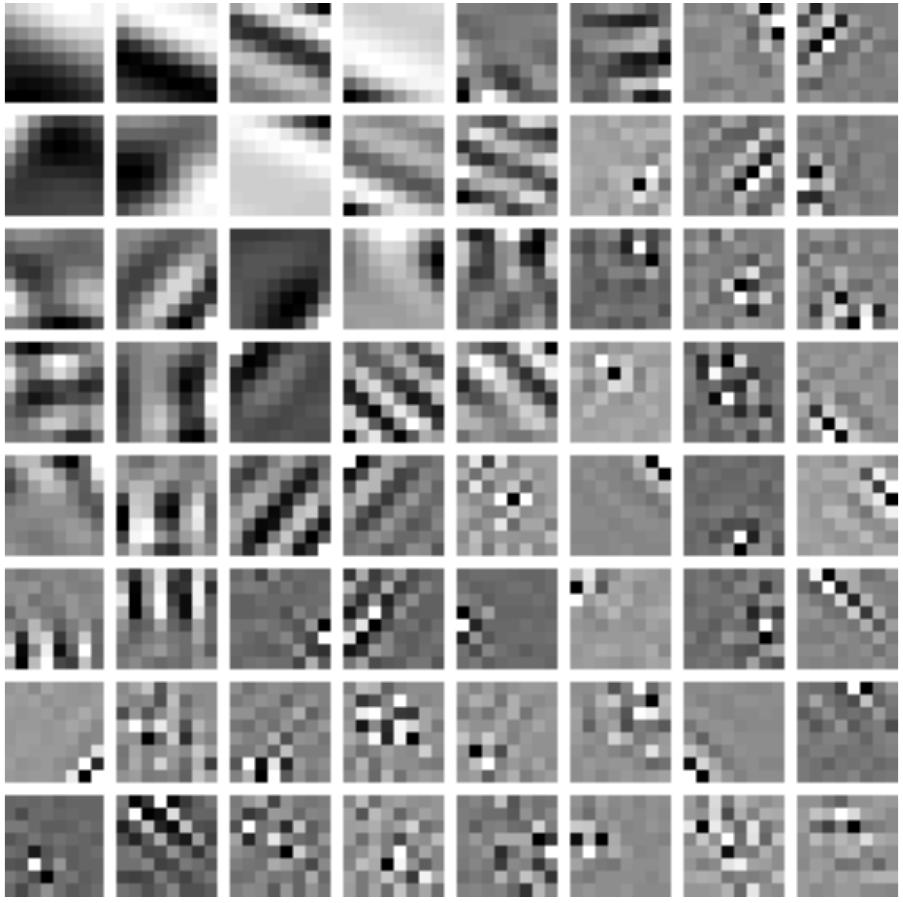} \\
		\begin{overpic}[scale=0.235]{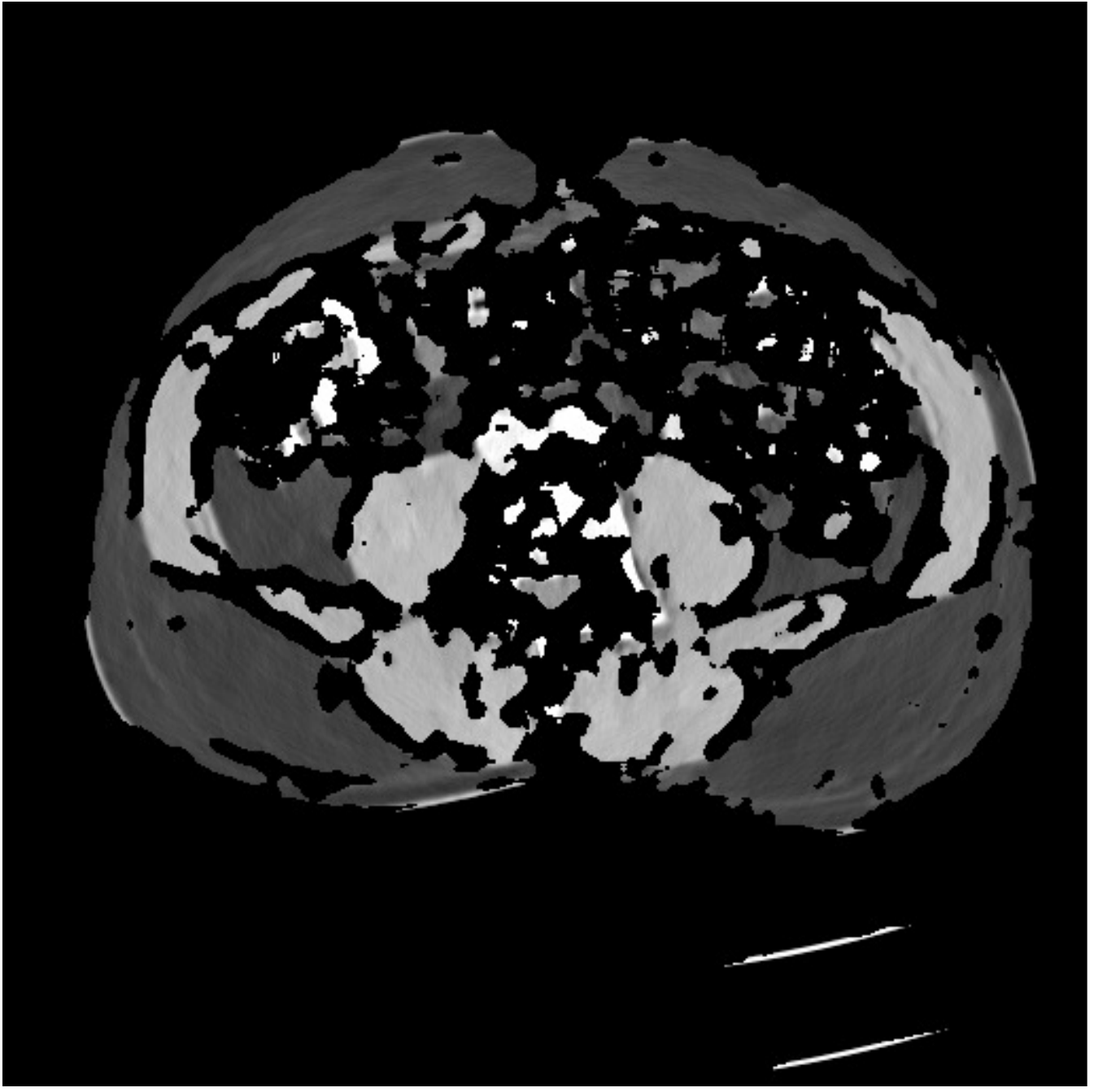}\put(30,12){ \color{white}{\bf \large{Class 1}}} 	\end{overpic}
		\begin{overpic}[scale=0.235]{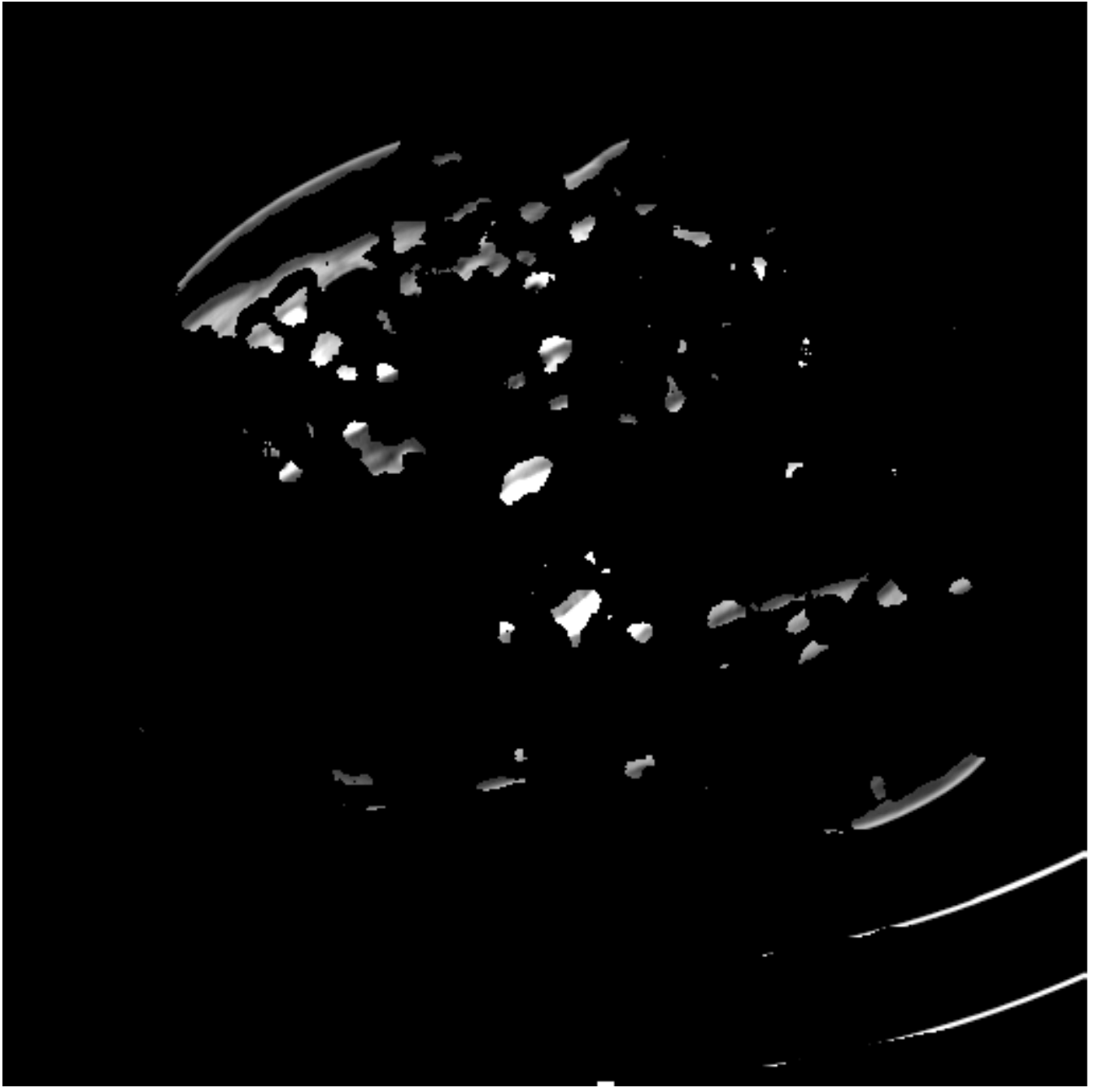}
		\put(30,12){ \color{white}{\bf \large{Class 2}}} 	\end{overpic}
		\begin{overpic}[scale=0.235]{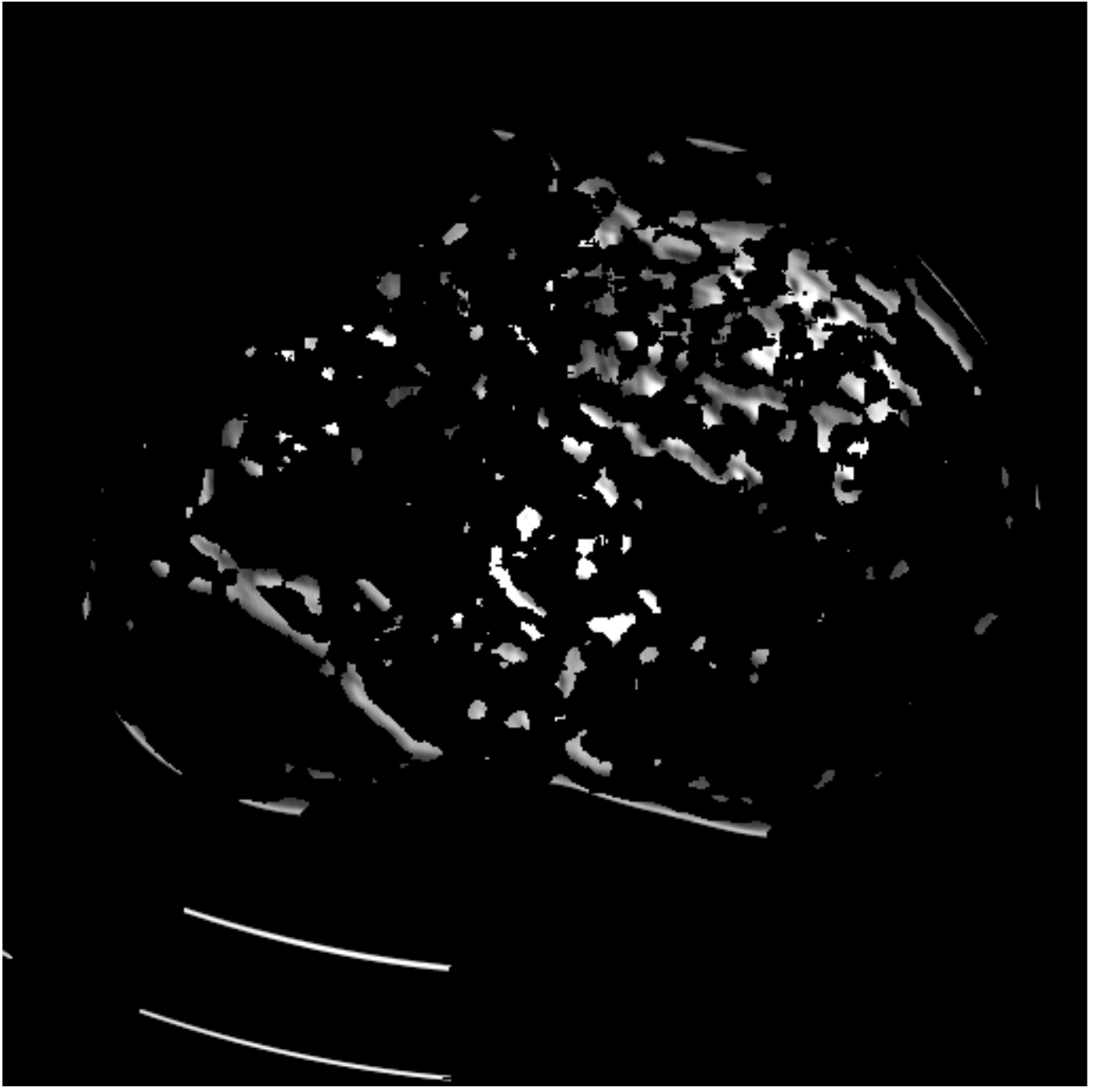}
		\put(30,12){ \color{white}{\bf \large{Class 3}}} 	\end{overpic}
		\begin{overpic}[scale=0.235]{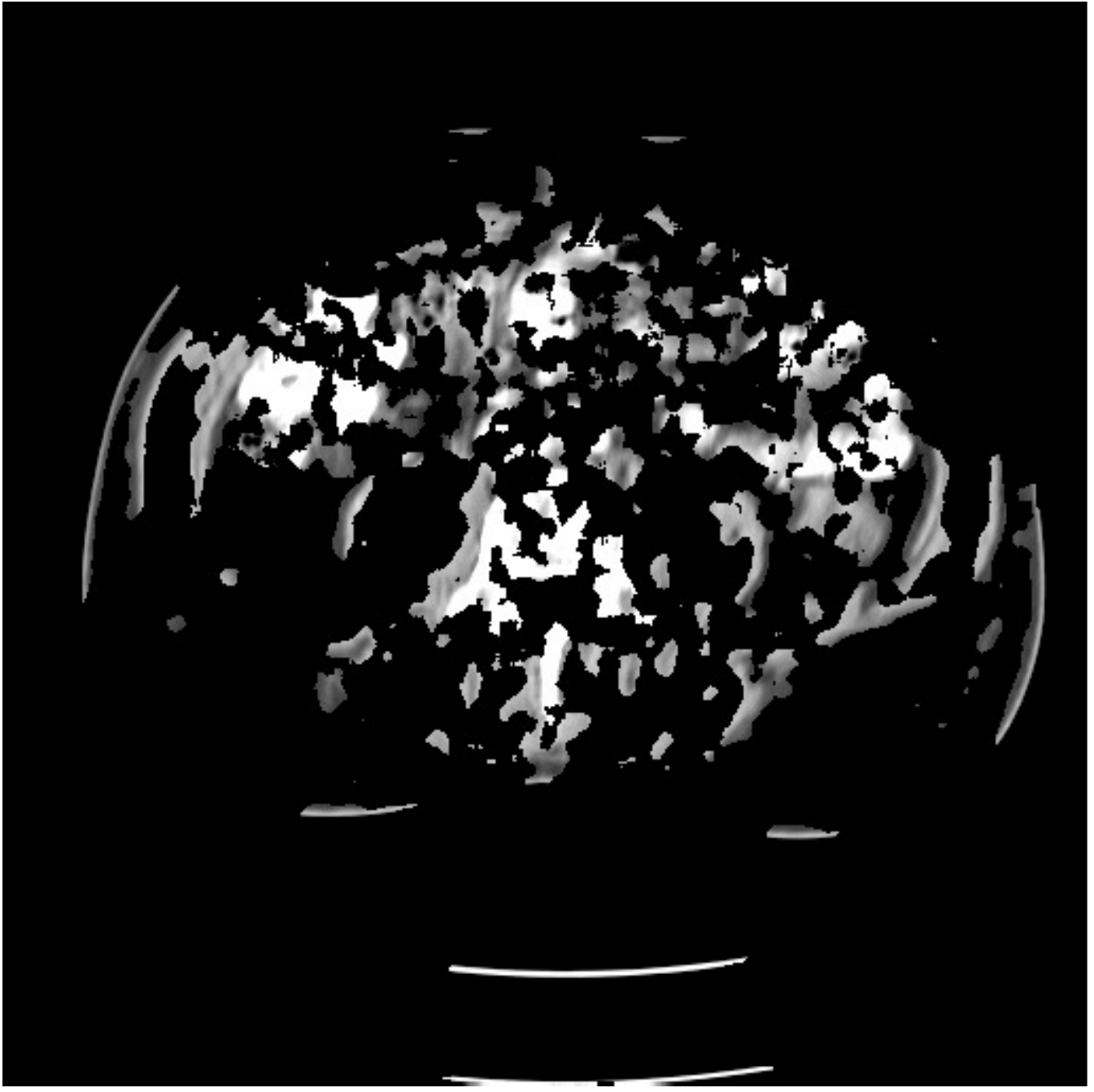}
		\put(30,12){ \color{white}{\bf \large{Class 4}}} 	\end{overpic}
		\begin{overpic}[scale=0.235]{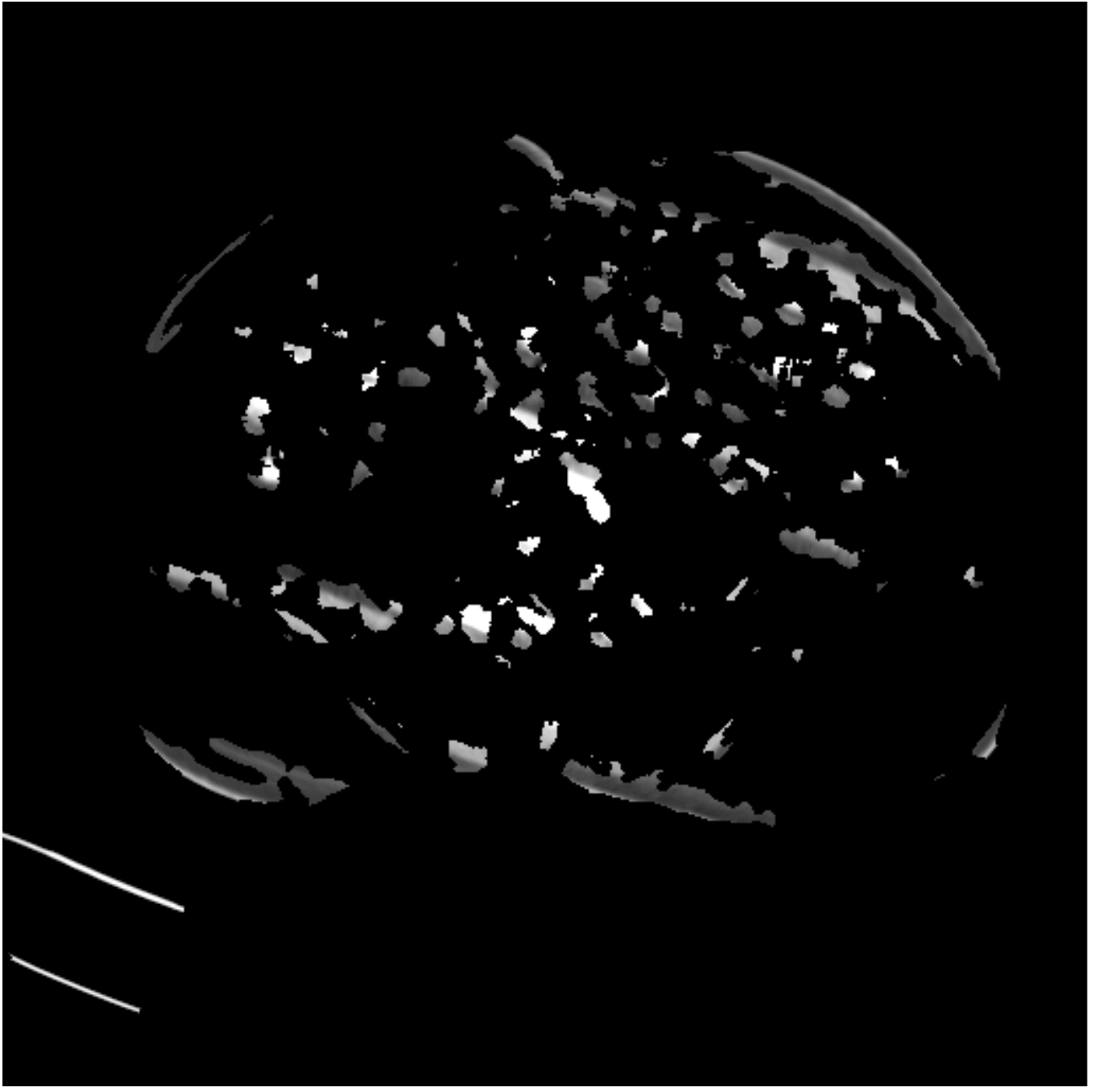}
		\put(30,12){ \color{white}{\bf \large{Class 5}}} 	\end{overpic}
		\vspace{-0.02in}
	\caption{Pixel-level clustering results in Test \#2. The top row shows the transforms, with the transform rows shown as $8\times8$ patches. The bottom row shows the clustering results of SUPER-ULTRA with display window [800 1200]~HU.}
	\label{fig:clstering}
	\vspace{-0.2in}
\end{figure*}

Next, to better illustrate the image-adaptive learned clustering in the SUPER-ULTRA model, Fig.~\ref{fig:clstering} shows pixel-level clustering results from the last super layer for test slice \#1.
Since the ULTRA modules cluster image patches into specific classes, we cluster each pixel here using a majority vote among the patches overlapping the pixel.
Class 1 contains most of the more uniform soft-tissues; Classes 2, 3, and 5 contain many oriented edges (e.g., at 45-degree and 135-degree orientation); and class 4 contains most of the vertical edges and some horizontal edges as well as most of the bones.
The latter classes help provide sharper SUPER-ULTRA reconstructions.
The pre-learned transforms corresponding to each class are also shown in Fig.~\ref{fig:clstering}, and contain various directional and edge-like features.

\vspace{-0.1in}
\section{Conclusions}
\vspace{-0.03in}
This paper presented a new framework that combined supervised learned networks and unsupervised iterative algorithms for low-dose CT reconstruction. 
The proposed SUPER framework effectively combines various kinds of priors and learning methods.
In particular, we studied SUPER-ULTRA that combines (supervised) deep learning (FBPConvNet) and the recent iterative (unsupervised) PWLS-ULTRA, as well as FBPConvNet+EP (or SUPER-EP).
Both methods showed better performance and faster convergence compared to their individual modules. FBPConvNet+EP substantially improved the performance of PWLS-EP, while SUPER-ULTRA \DIFadd{typically} performed the best by effectively leveraging deep learning and transform learning.
While SUPER model learning can exploit a variety of architectures and algorithms for the supervised and iterative modules, a more detailed study of various such architectures is left for future work.
We also plan to explore layer-dependent parameter selection for the iterative modules to further improve performance in future work.

	
{\small
\bibliographystyle{ieee}
\bibliography{egbib}
}


\newpage
\clearpage



{
\twocolumn[
\begin{center}
 \Huge SUPER Learning: A Supervised-Unsupervised Framework for Low-Dose CT Image Reconstruction -- Supplementary Material
\vspace{0.2in}
\end{center}]
}










\newcommand{\DIFaddadd}{\textcolor{white}}








\setcounter{section}{4} 
\section{Additional Experimental Results} 
\setcounter{figure}{7}

\begin{figure}[htb]
	\centering
	\begin{overpic}[scale=0.23]{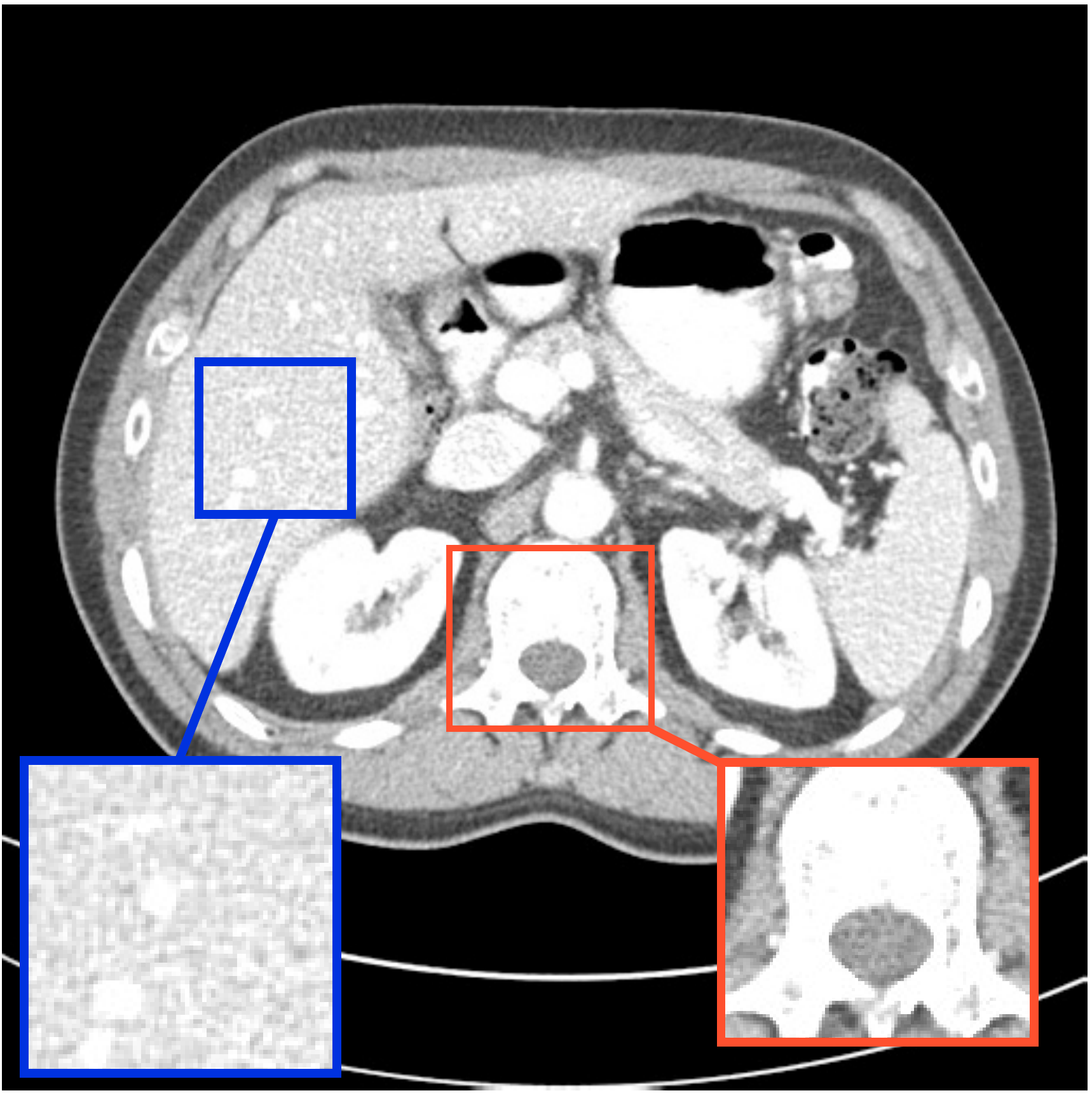} 
		\put(25,90){ \color{white}{\bf \large{}}} 	\end{overpic}	
	\begin{overpic}[scale=0.23]{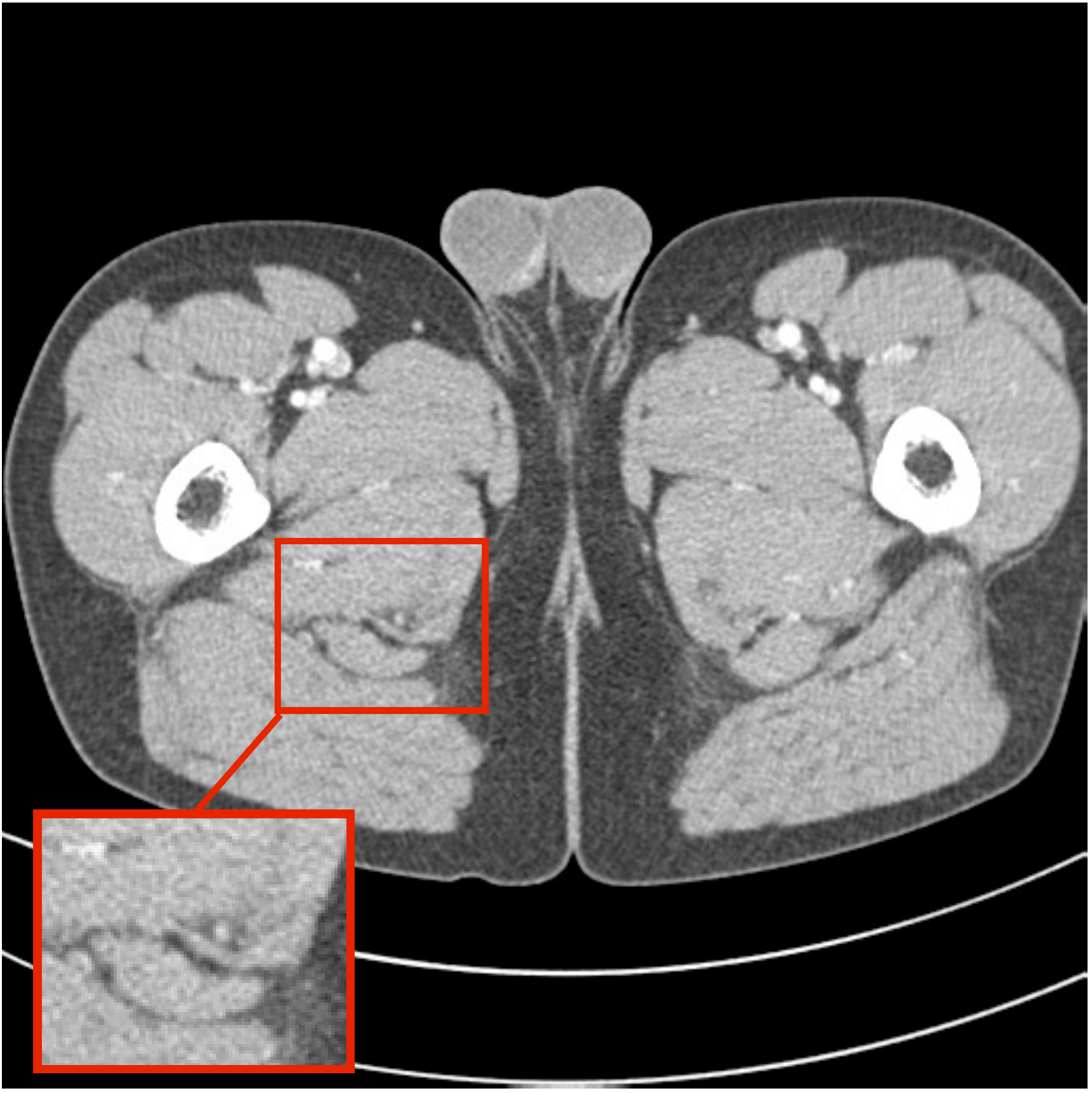} 
		\put(20,88){ \color{white}{\bf \large{}}} 	\end{overpic}
	\begin{overpic}[scale=0.23]{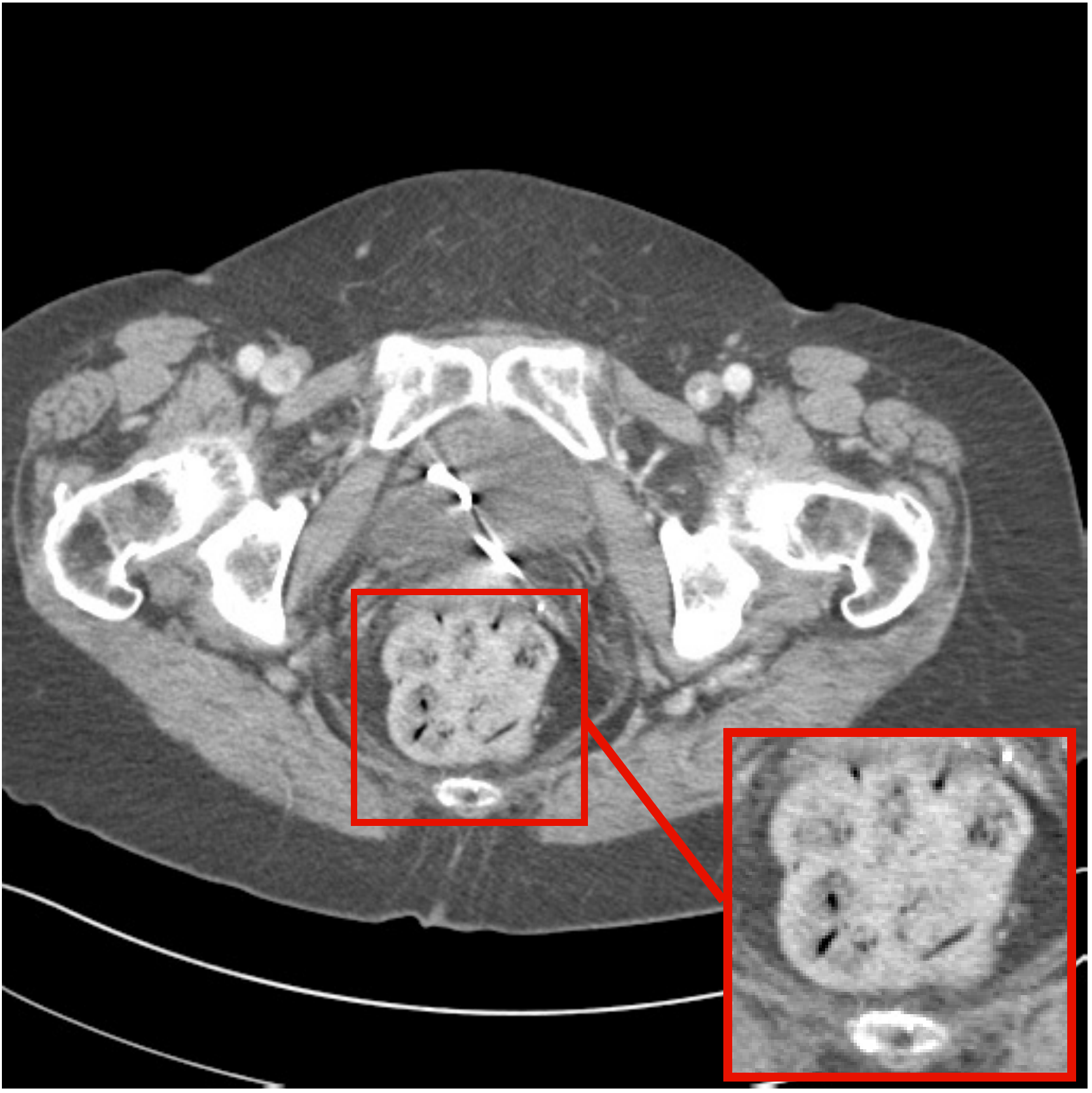} 
		\put(25,90){ \color{white}{\bf \large{}}} 	\end{overpic}
	\begin{overpic}[scale=0.23]{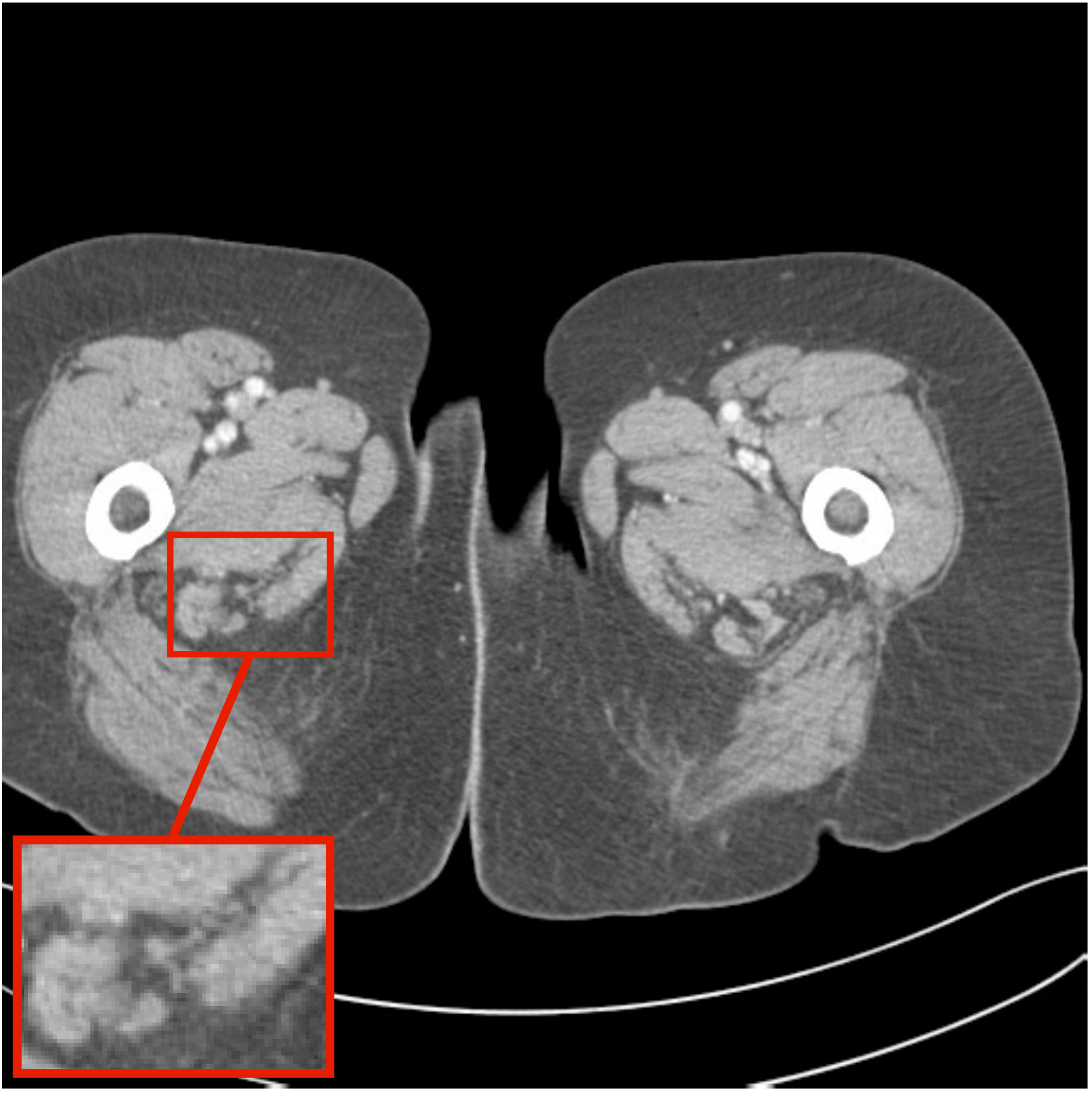} 
		\put(25,90){ \color{white}{\bf \large{}}} 	\end{overpic}
	\caption{Regular-dose FBP images of four testing slices: Test \#1 (top left), \#3 (top right), \#5 (bottom left), and \#6 (bottom right). The display window is [800 1200]~HU.} 
	\label{fig:ref}
\end{figure}

\begin{figure}[htb]
	\centering
	\begin{overpic}[scale=0.22]{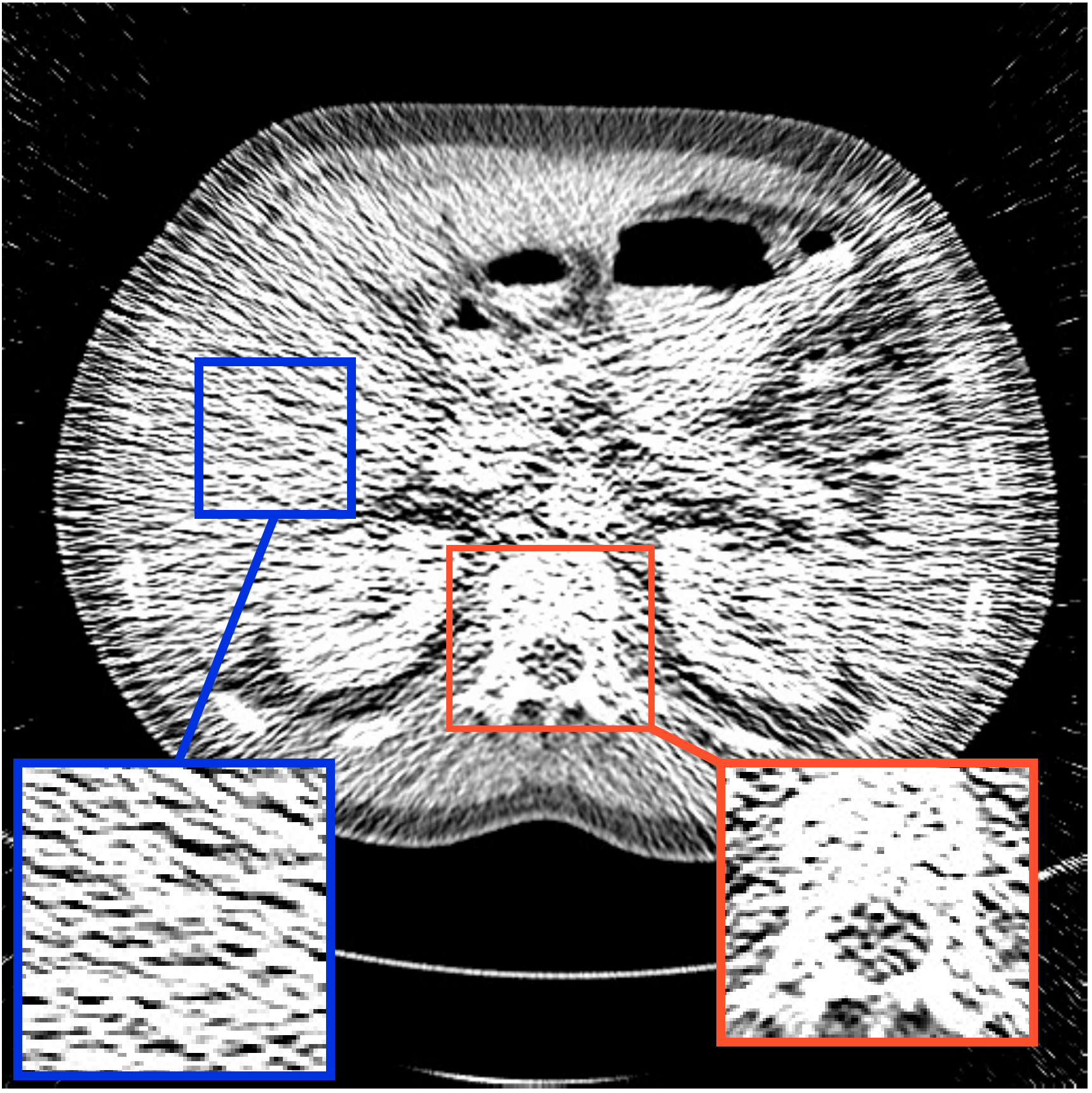} 
		\put(20,92){ \color{white}{\bf \normalsize{PSNR: 11.0}}} 	\end{overpic}
	\begin{overpic}[scale=0.22]{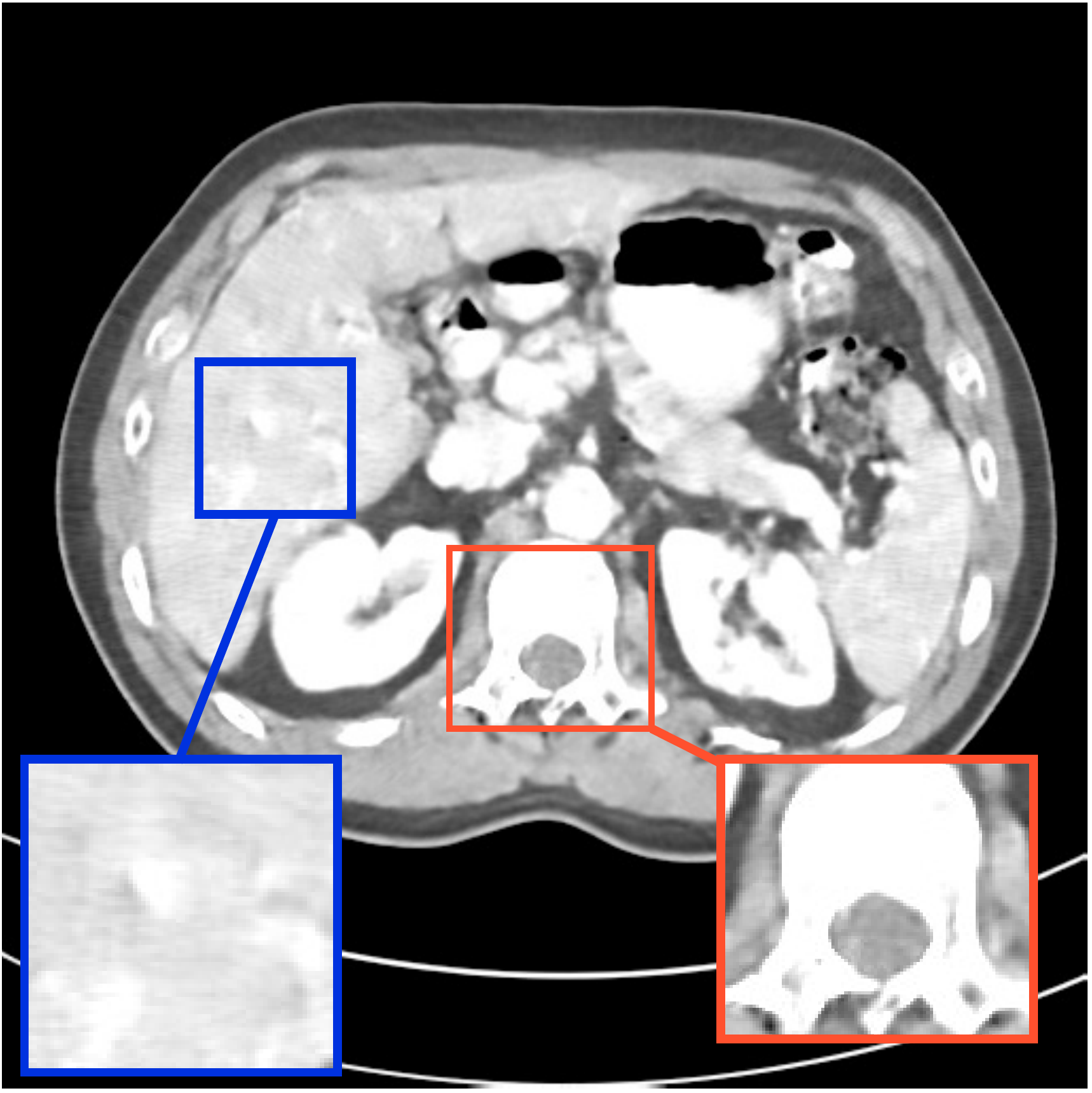} 
		\put(20,92){ \color{white}{\bf \normalsize{PSNR: 29.8}}} 	\end{overpic} \\
	\begin{overpic}[scale=0.22]{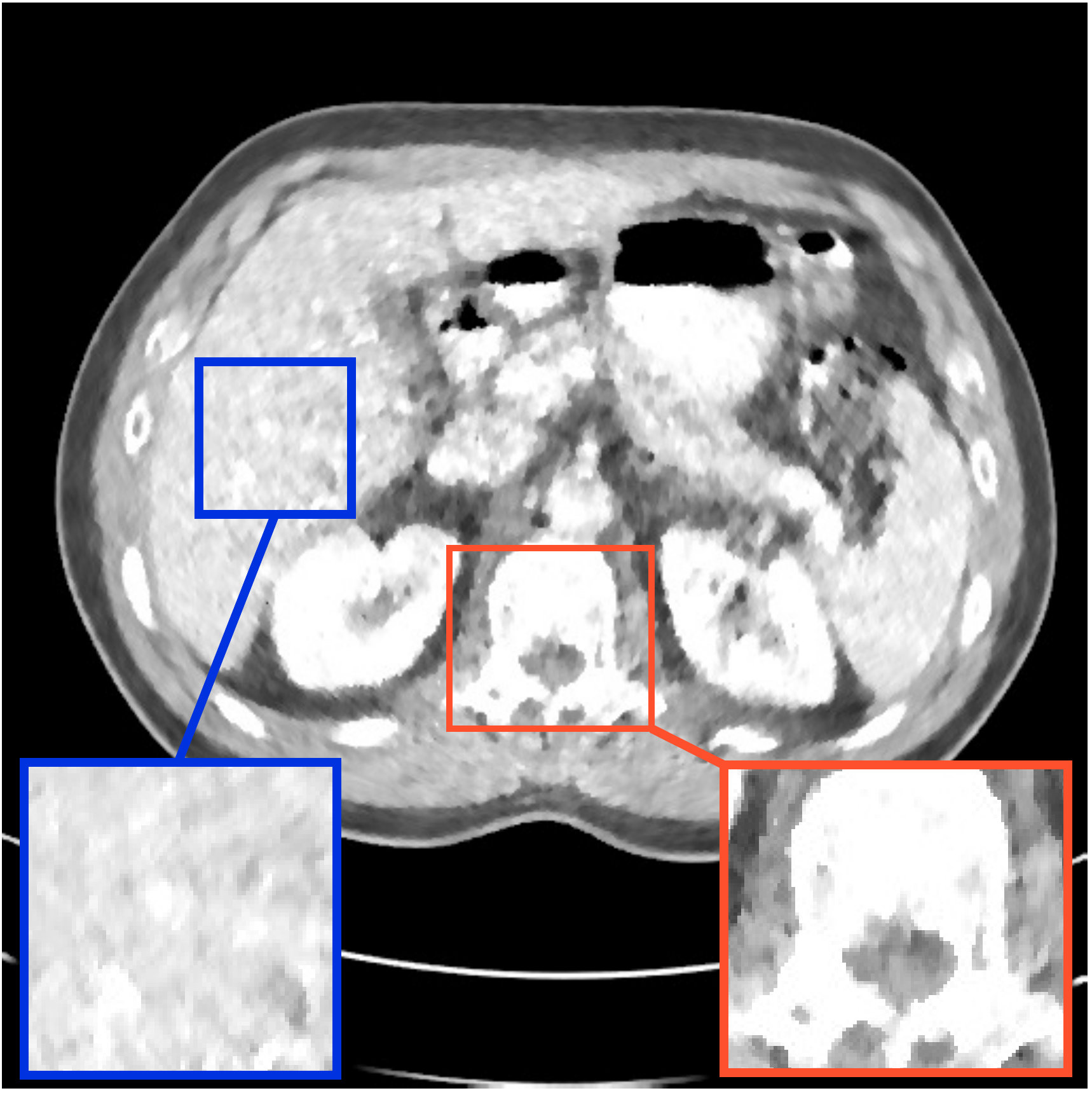} 
		\put(20,92){ \color{white}{\bf \normalsize{PSNR: 23.5}}} 	\end{overpic}
	\begin{overpic}[scale=0.22]{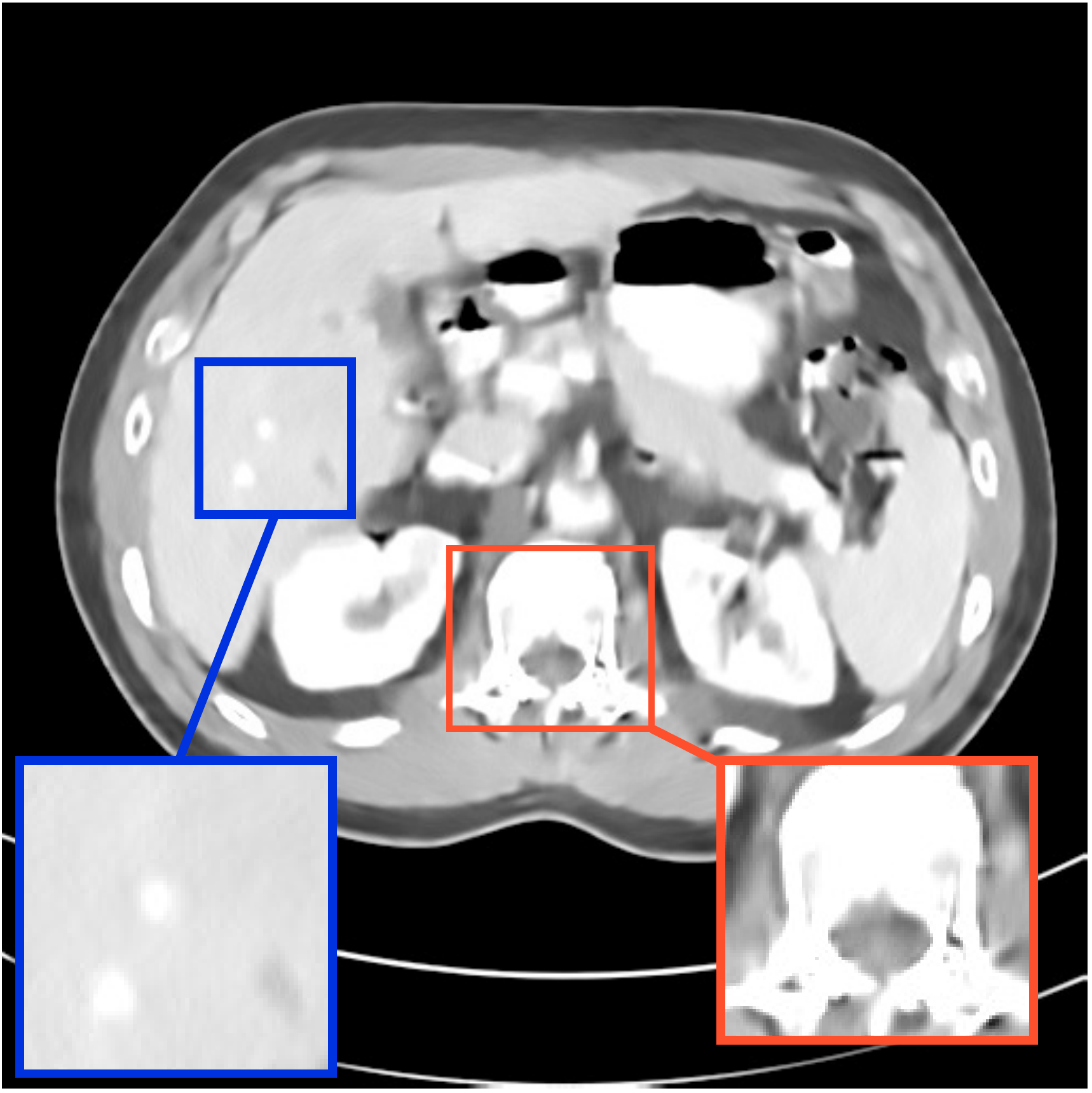} 
		\put(20,92){ \color{white}{\bf \normalsize{PSNR: 29.3}}} 	\end{overpic} \\ 
	\begin{overpic}[scale=0.22]{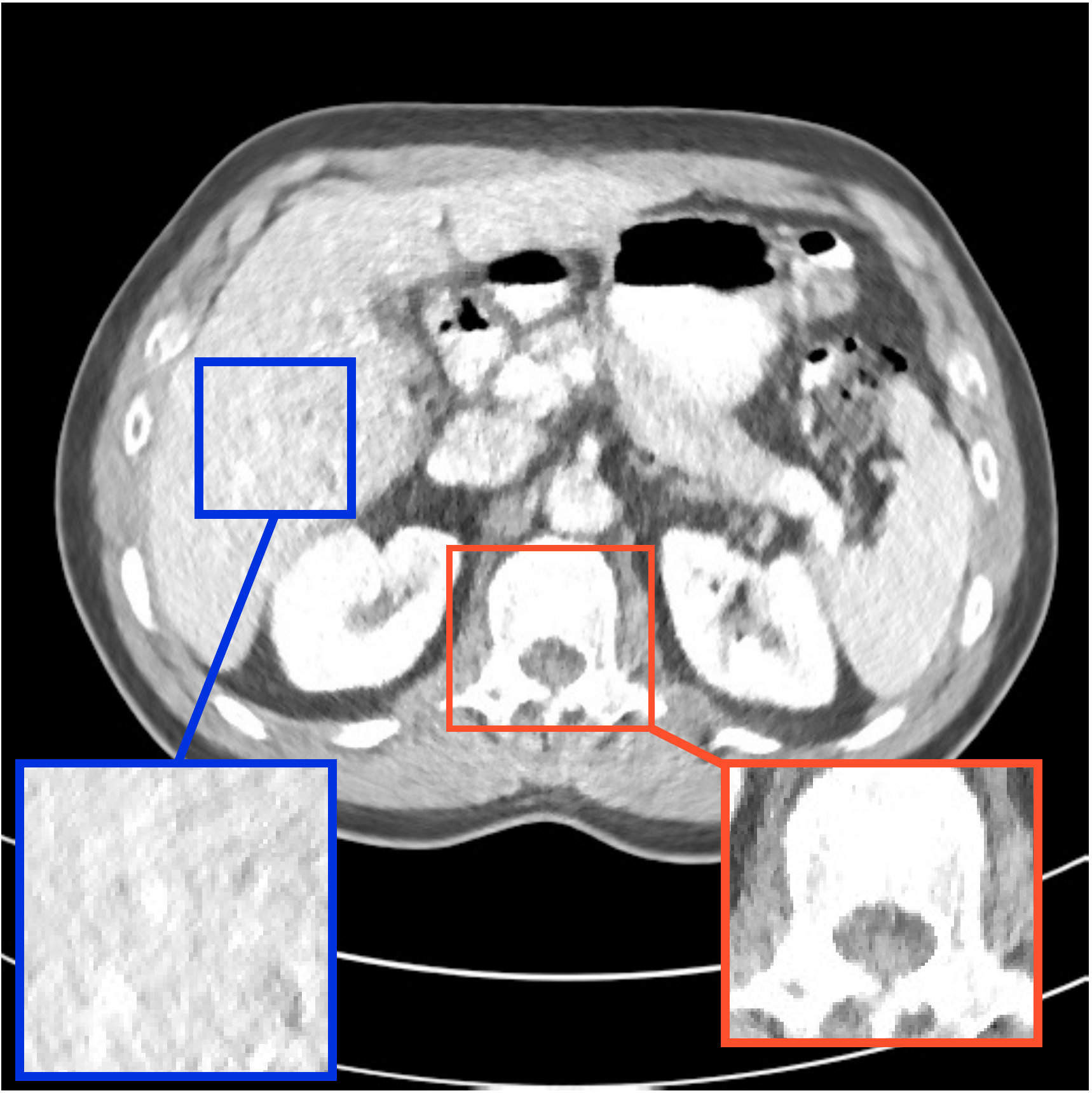} 
		\put(20,92){ \color{white}{\bf \normalsize{PSNR: \DIFaddadd{30.5}}}} 	\end{overpic}
	\begin{overpic}[scale=0.22]{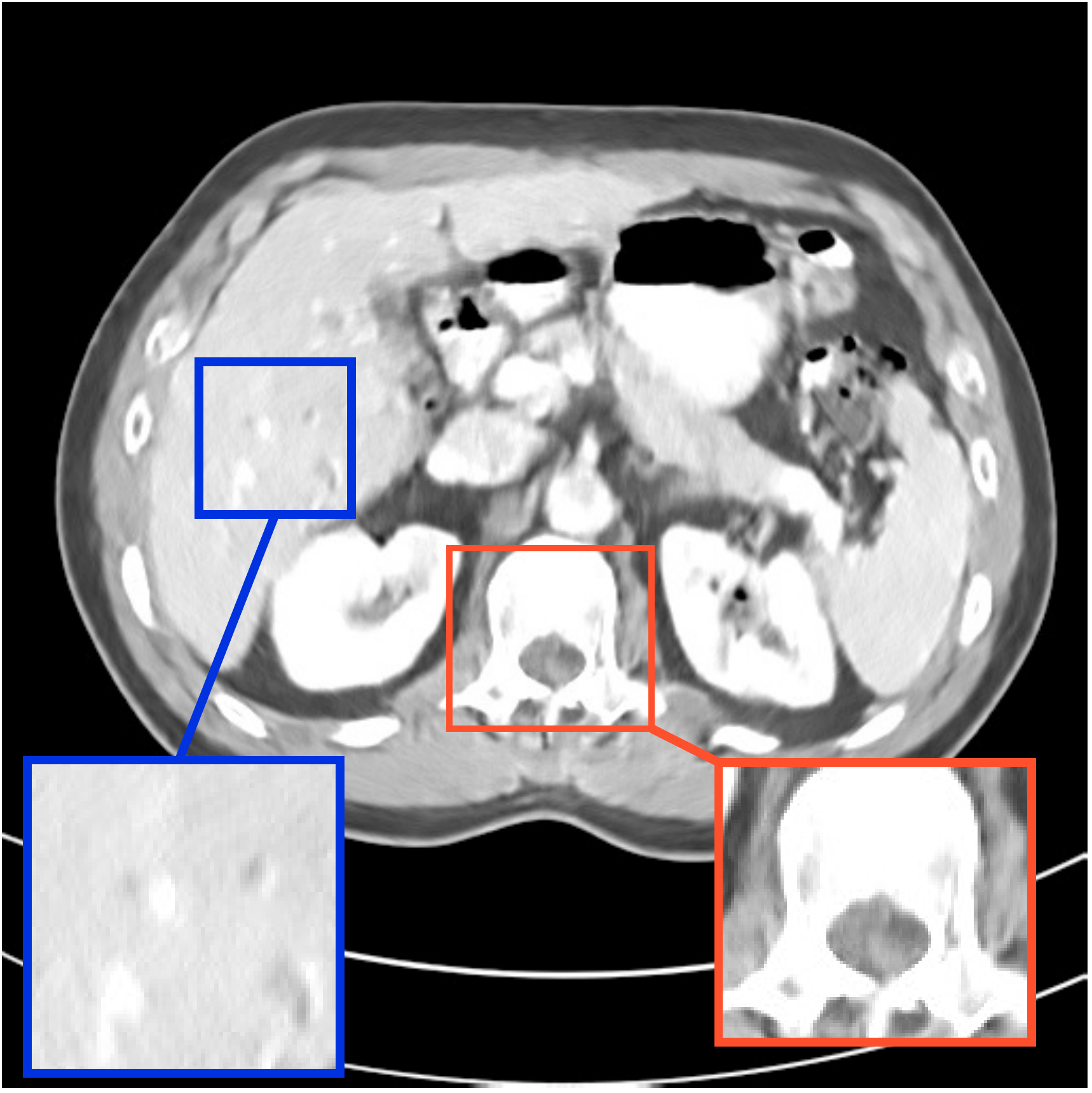} 
		\put(20,92){ \color{white}{\bf \normalsize{PSNR: 30.9}}} 	\end{overpic}
	\caption{Reconstructed testing image (Test \#1, from patient L067) obtained by FBP (top left), FBPConvNet (top right), PWLS-EP (middle left), PWLS-ULTRA (middle right), FBPConvNet + EP (bottom left), and SUPER-ULTRA (bottom right). The display window is [800 1200]~HU.}
	\label{fig:L067Slice50_image}
	\vspace{-0.2in}
\end{figure}

\begin{figure}[!t]
	\centering
	\begin{overpic}[scale=0.22]{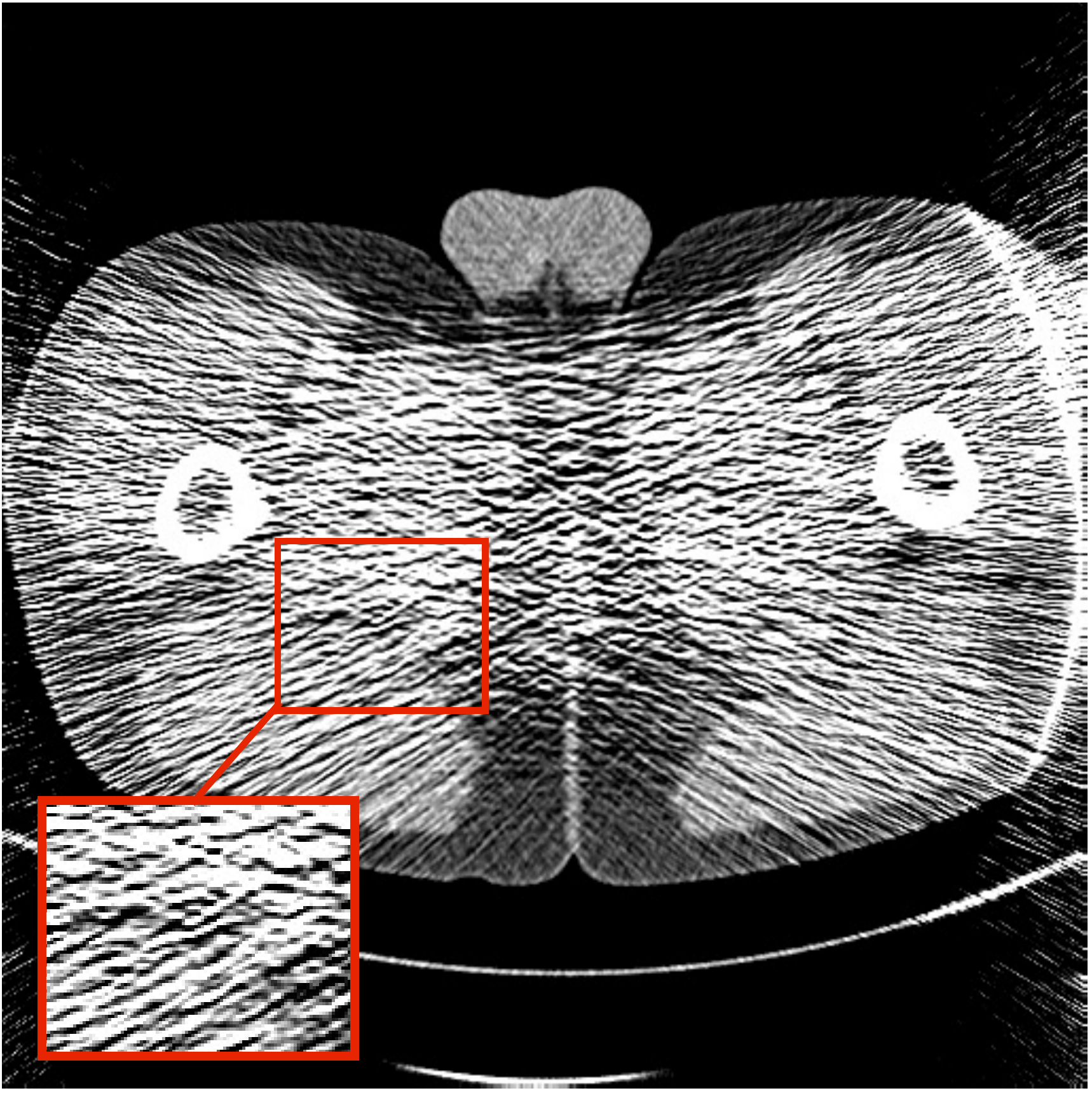} 
		\put(20,88){ \color{white}{\bf \normalsize{PSNR: 9.5}}} 	\end{overpic}
	\begin{overpic}[scale=0.22]{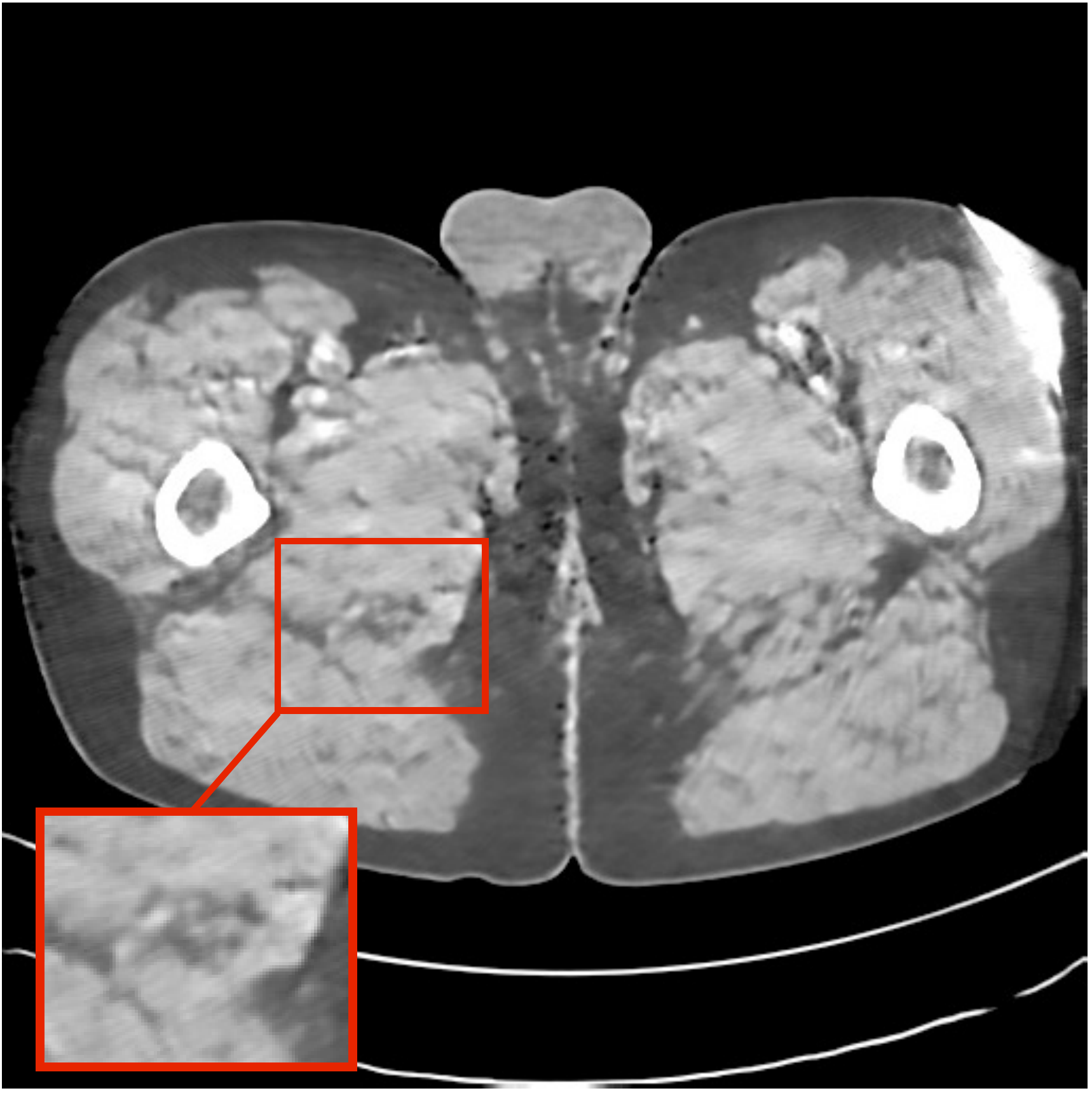} 
		\put(20,88){ \color{white}{\bf \normalsize{PSNR: 19.9}}} 	\end{overpic} \\
	\begin{overpic}[scale=0.22]{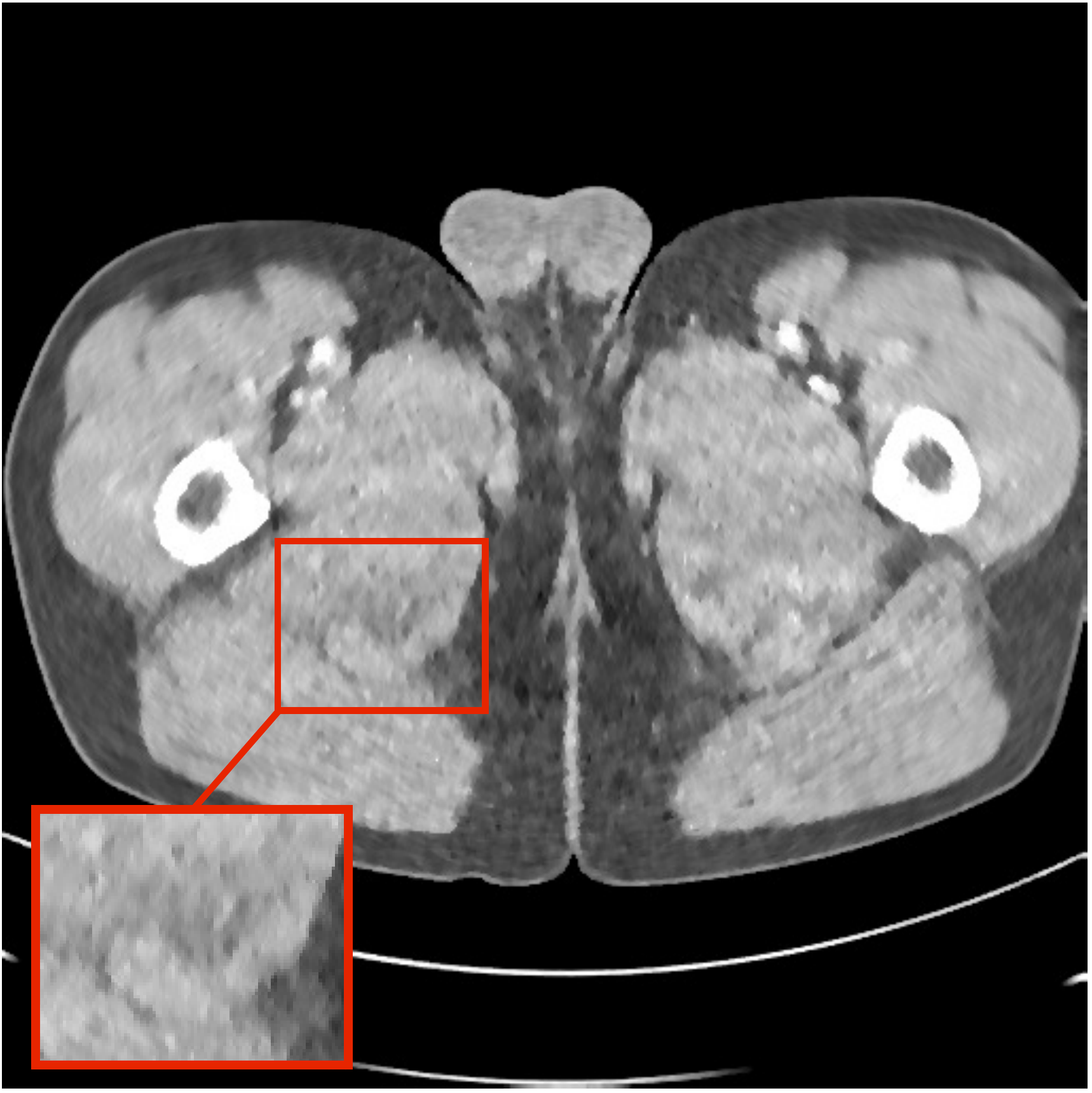} 
		\put(20,88){ \color{white}{\bf \normalsize{PSNR: 24.4}}} 	\end{overpic} 
	\begin{overpic}[scale=0.22]{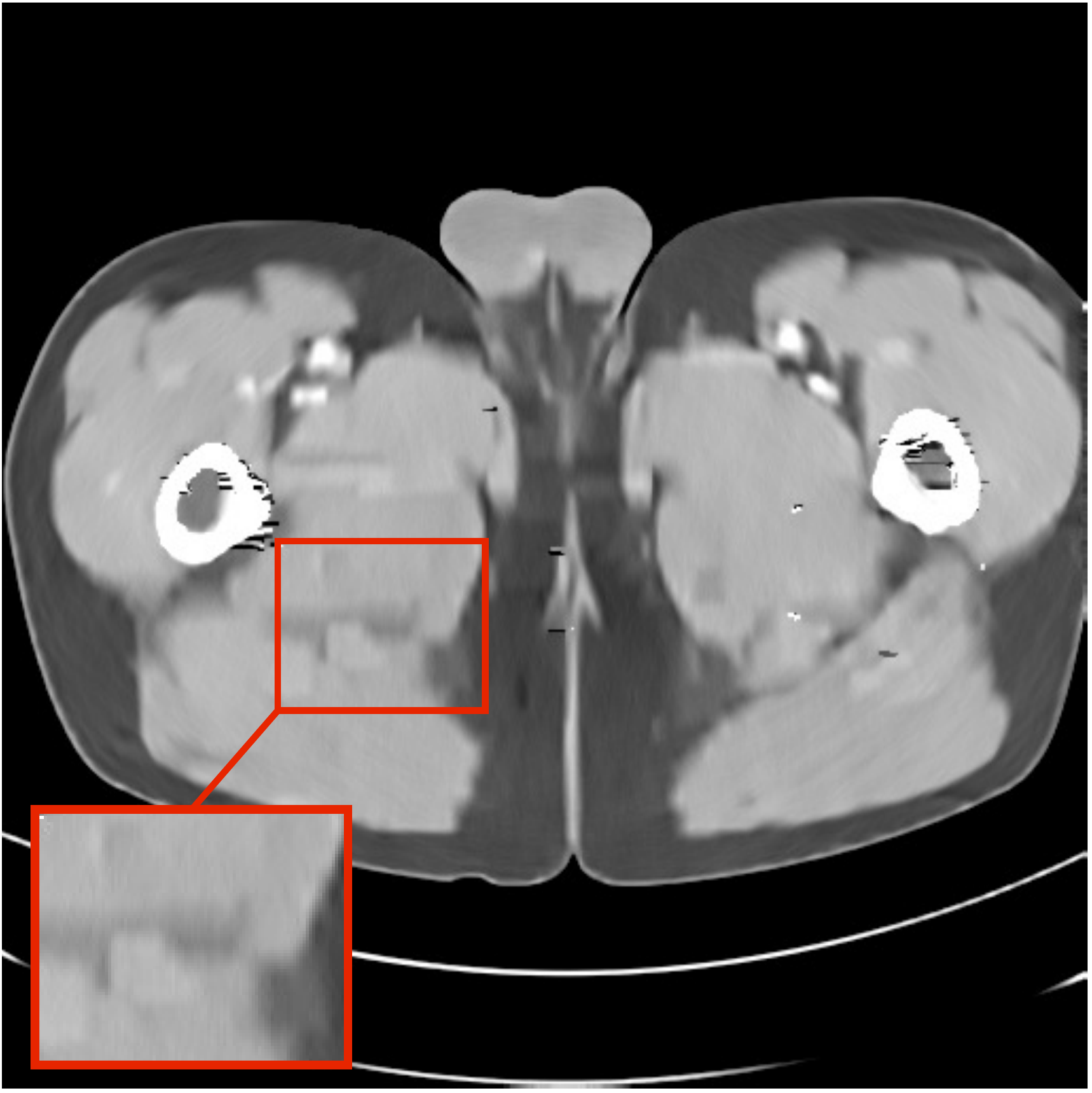} 
		\put(20,88){ \color{white}{\bf \normalsize{PSNR: \DIFaddadd{28.6}}}} 	\end{overpic} \\
	\begin{overpic}[scale=0.22]{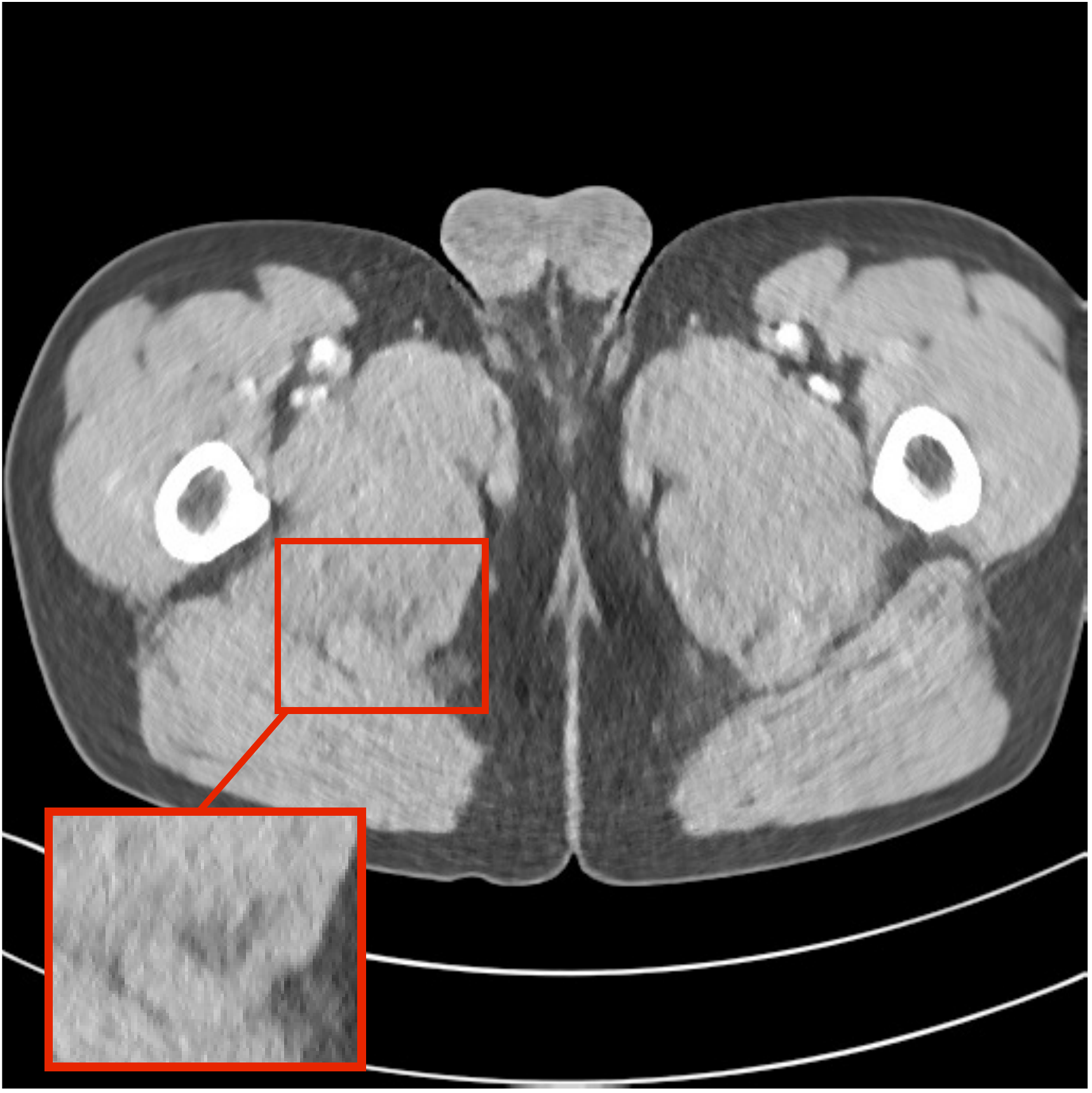} 
		\put(20,88){ \color{white}{\bf \normalsize{PSNR: \DIFaddadd{32.2}}}} 	\end{overpic}
	\begin{overpic}[scale=0.22]{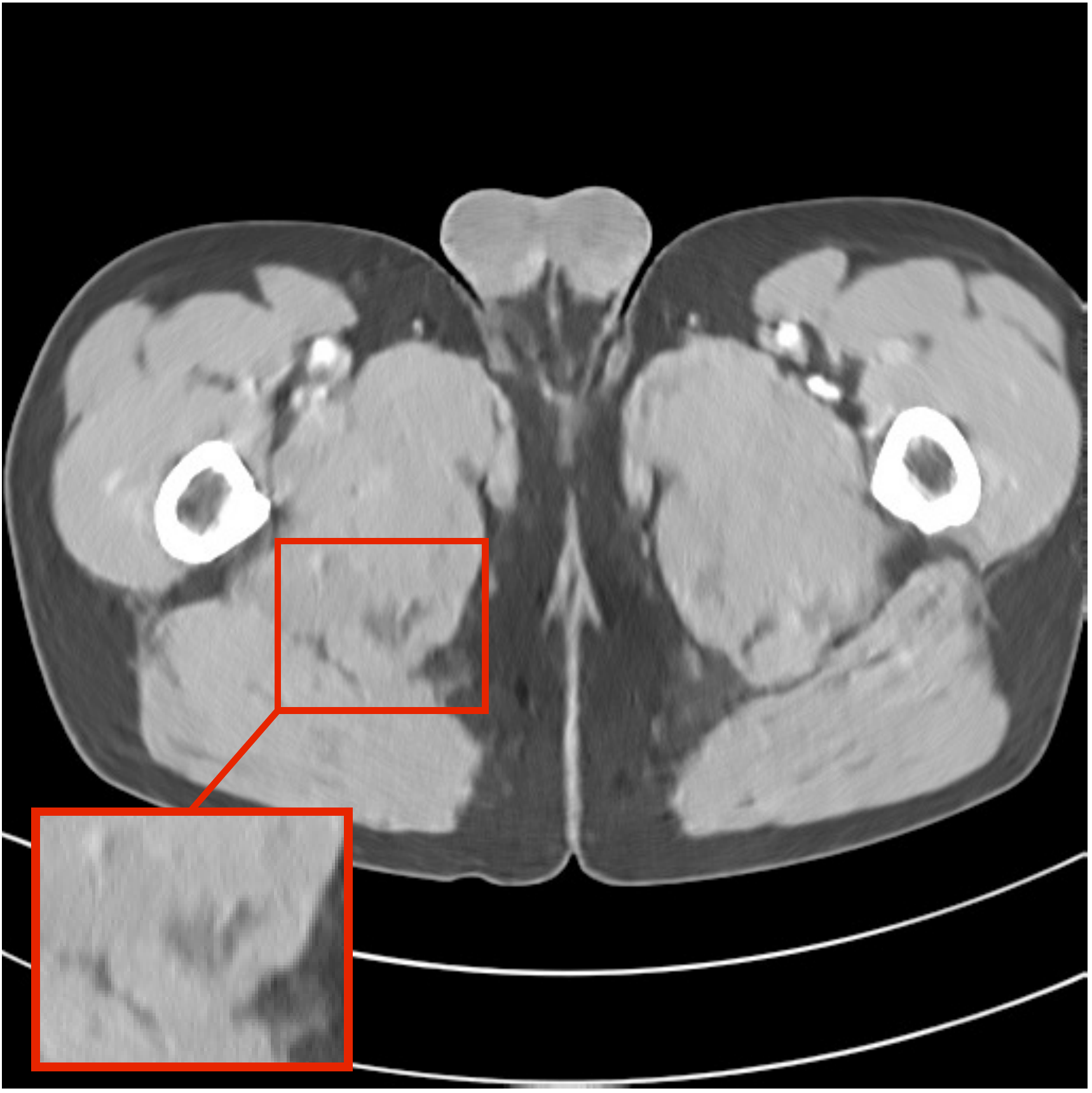} 
		\put(20,88){ \color{white}{\bf \normalsize{PSNR: 32.1}}} 	\end{overpic}
	\caption{Reconstructed testing image (Test \#3, from patient L067) obtained by FBP (top left), FBPConvNet (top right), PWLS-EP (middle left), PWLS-ULTRA (middle right), FBPConvNet + EP (bottom left), and SUPER-ULTRA (bottom right). The display window is [800 1200]~HU.}
	\label{fig:L067Slice210_image}
	\vspace{-0.2in}
\end{figure}

\begin{figure}[!t]
	\centering
	\begin{overpic}[scale=0.22]{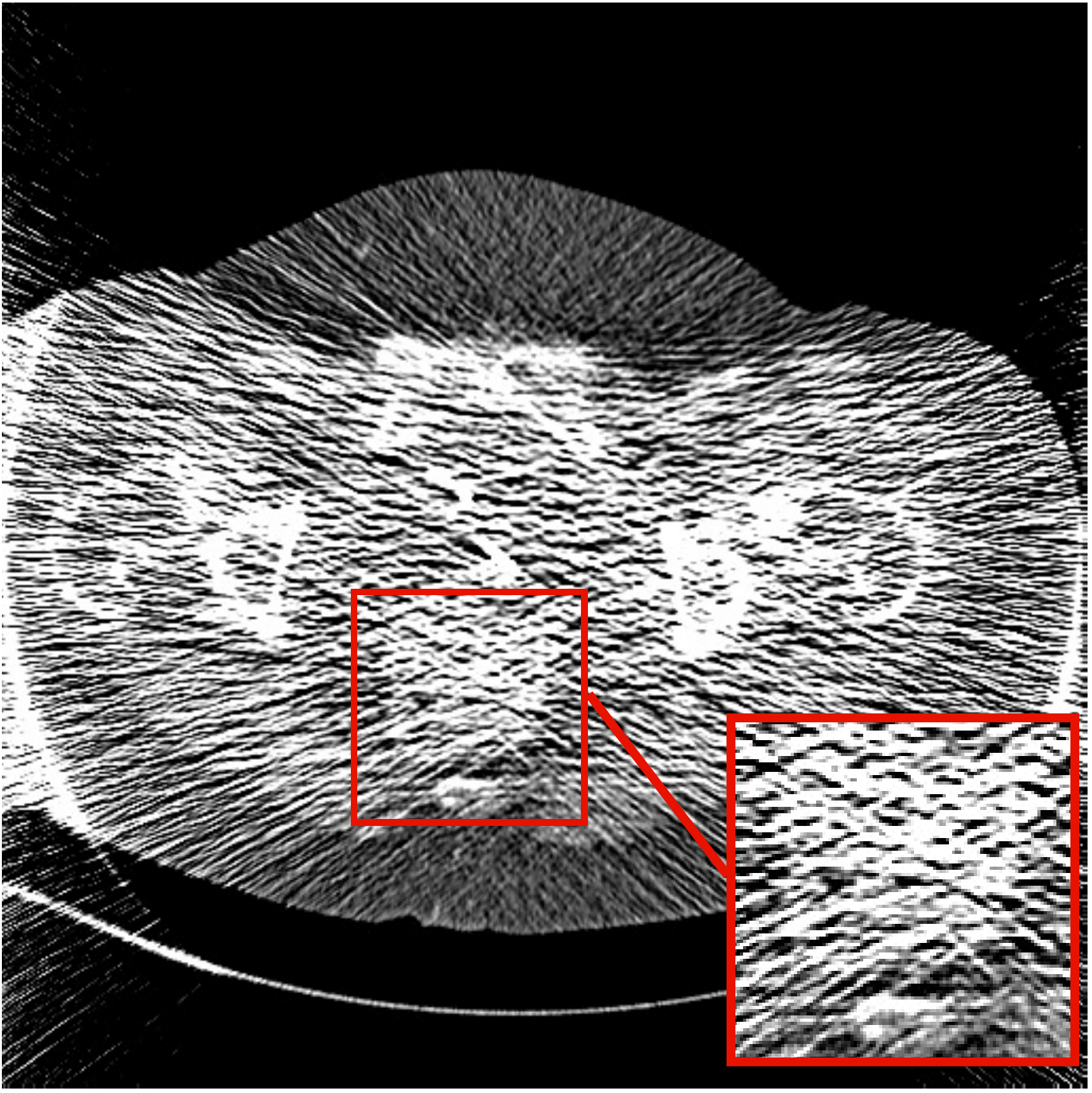} 
		\put(20,88){ \color{white}{\bf \normalsize{PSNR: 9.3}}} 	\end{overpic}
	\begin{overpic}[scale=0.22]{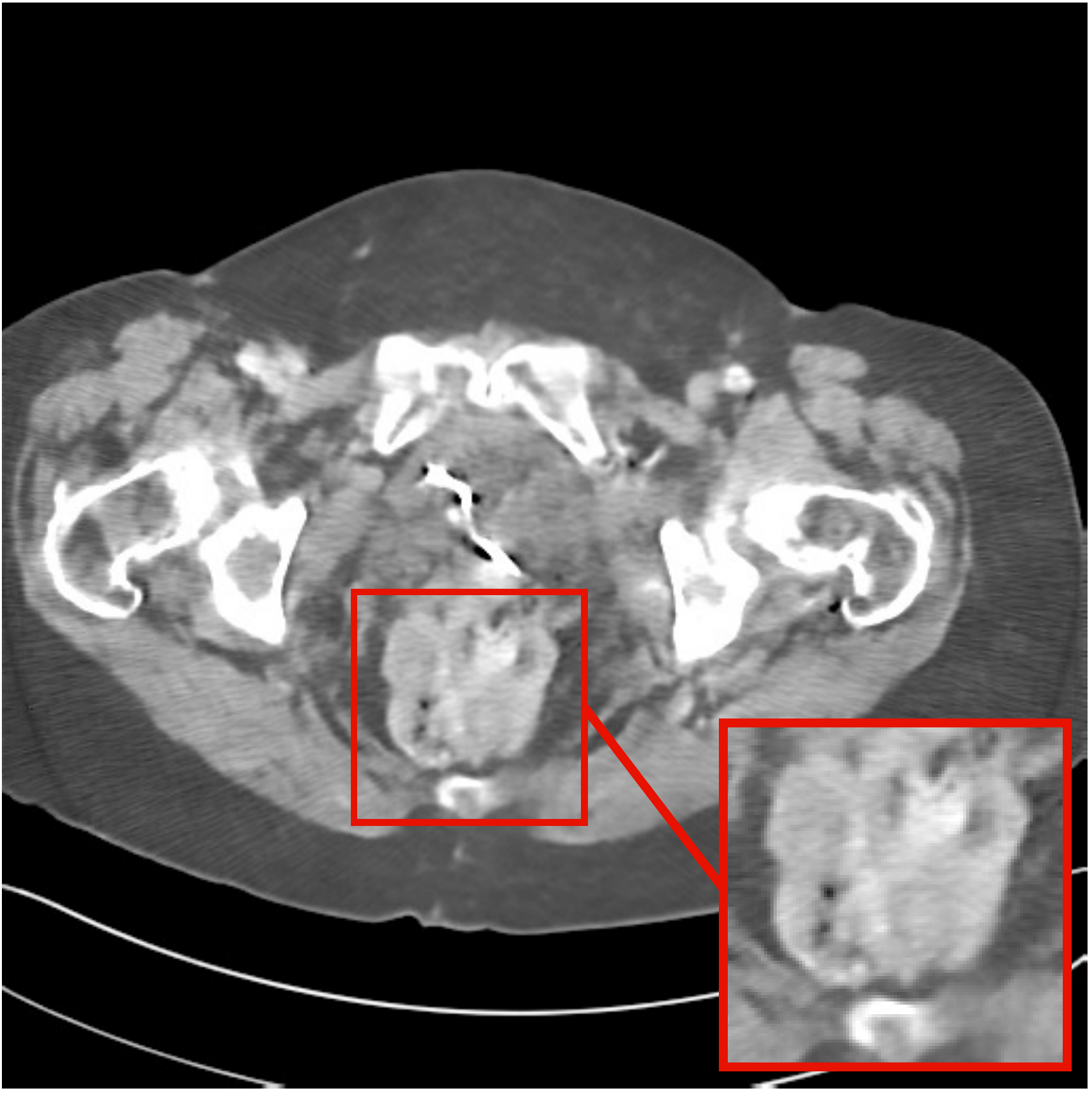} 
		\put(20,88){ \color{white}{\bf \normalsize{PSNR: 29.7}}} 	\end{overpic} \\
	\begin{overpic}[scale=0.22]{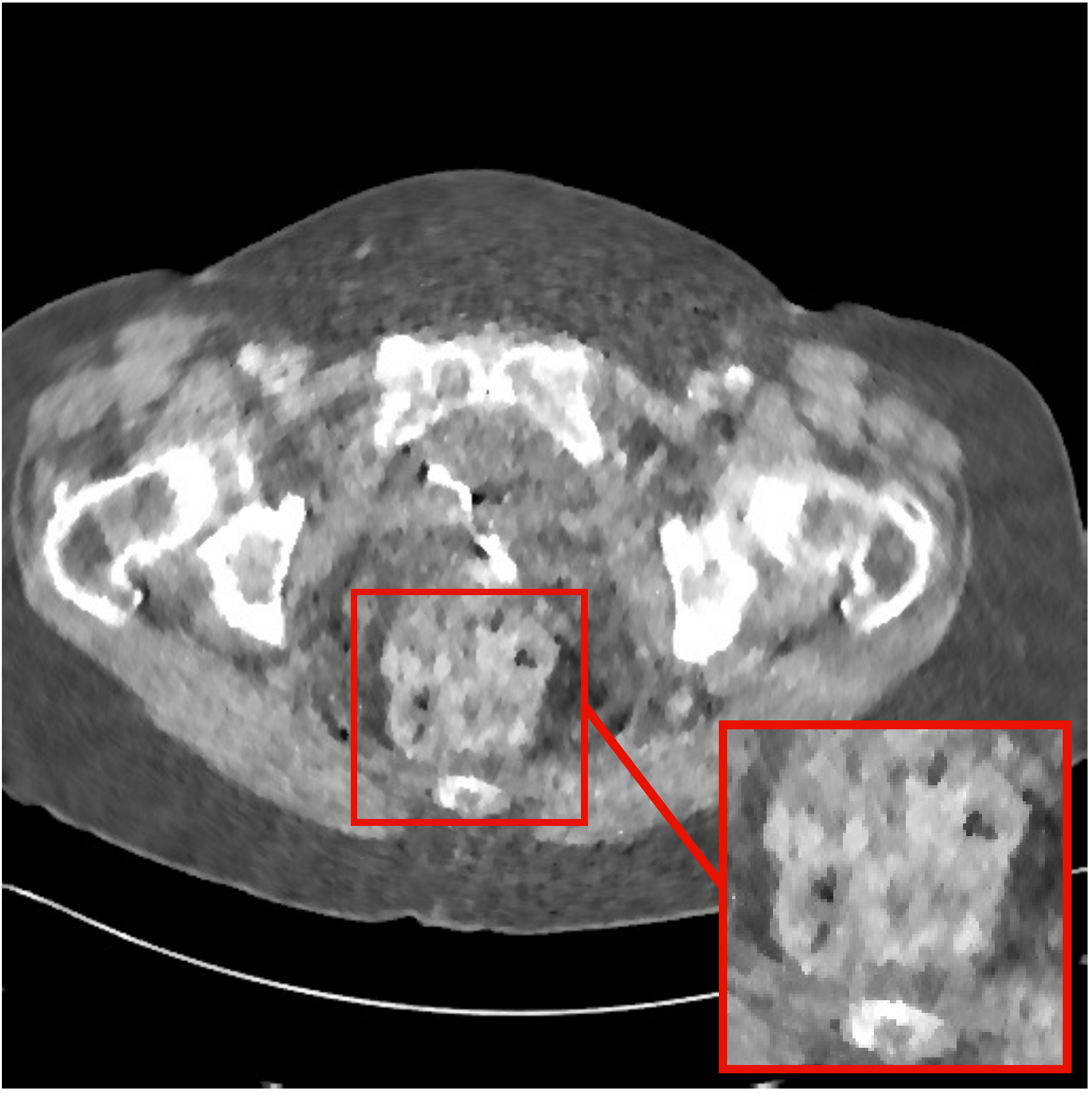} 
		\put(20,88){ \color{white}{\bf \normalsize{PSNR: 23.3}}} 	\end{overpic} 
	\begin{overpic}[scale=0.22]{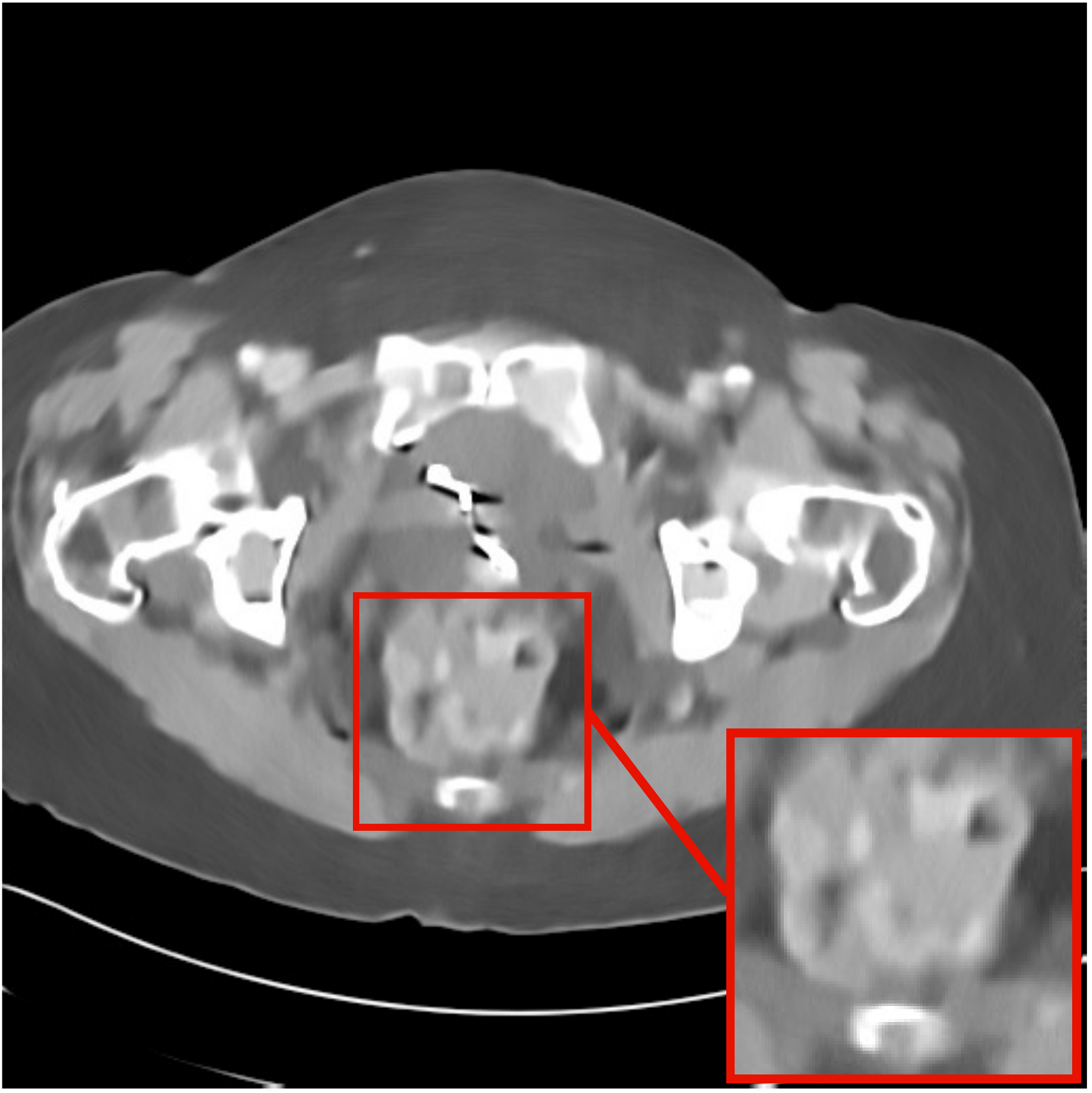} 
		\put(20,88){ \color{white}{\bf \normalsize{PSNR: 25.7}}} 	\end{overpic} \\
	\begin{overpic}[scale=0.22]{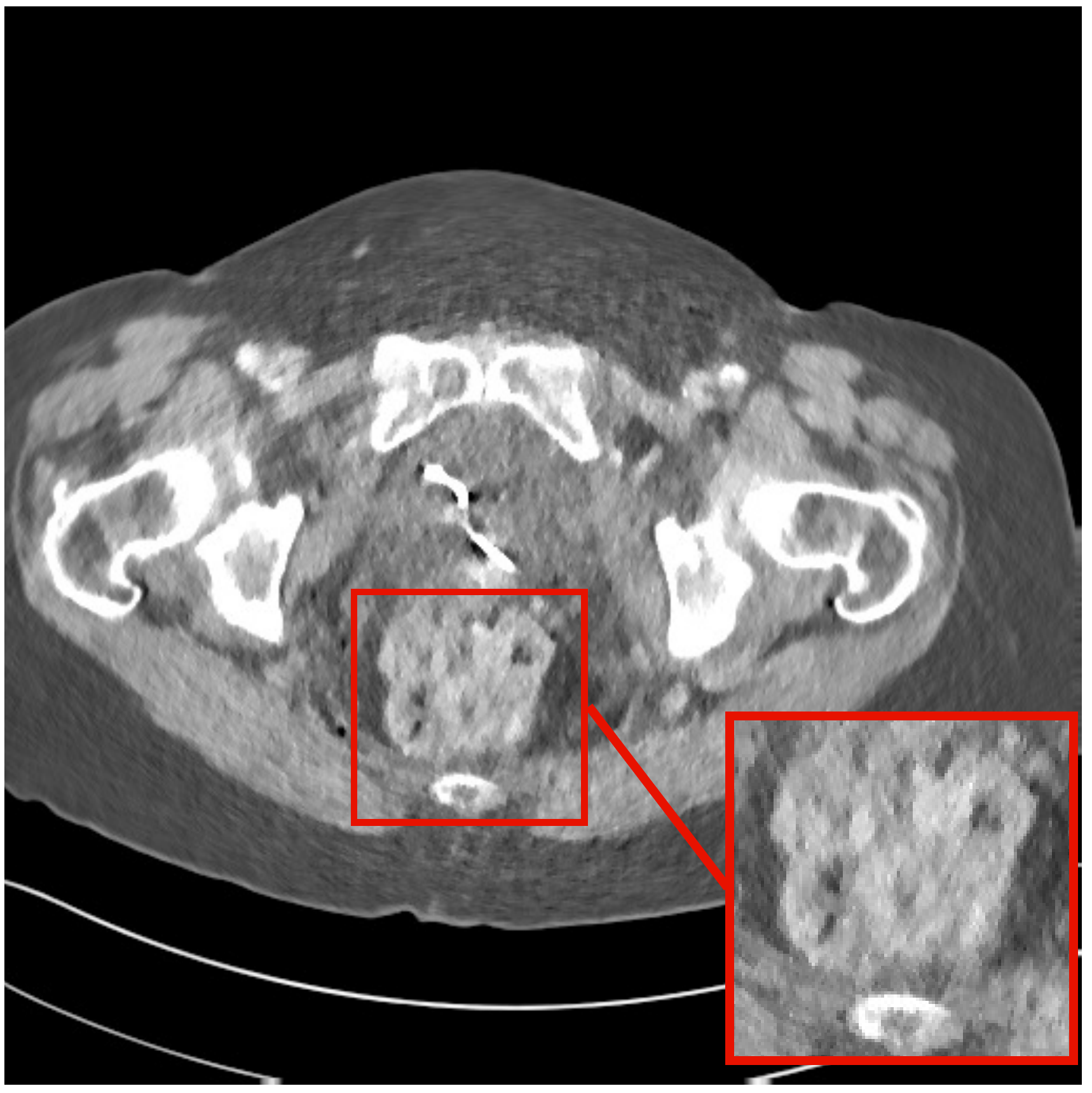} 
		\put(20,88){ \color{white}{\bf \normalsize{PSNR: \DIFaddadd{30.6}}}} 	\end{overpic}
	\begin{overpic}[scale=0.22]{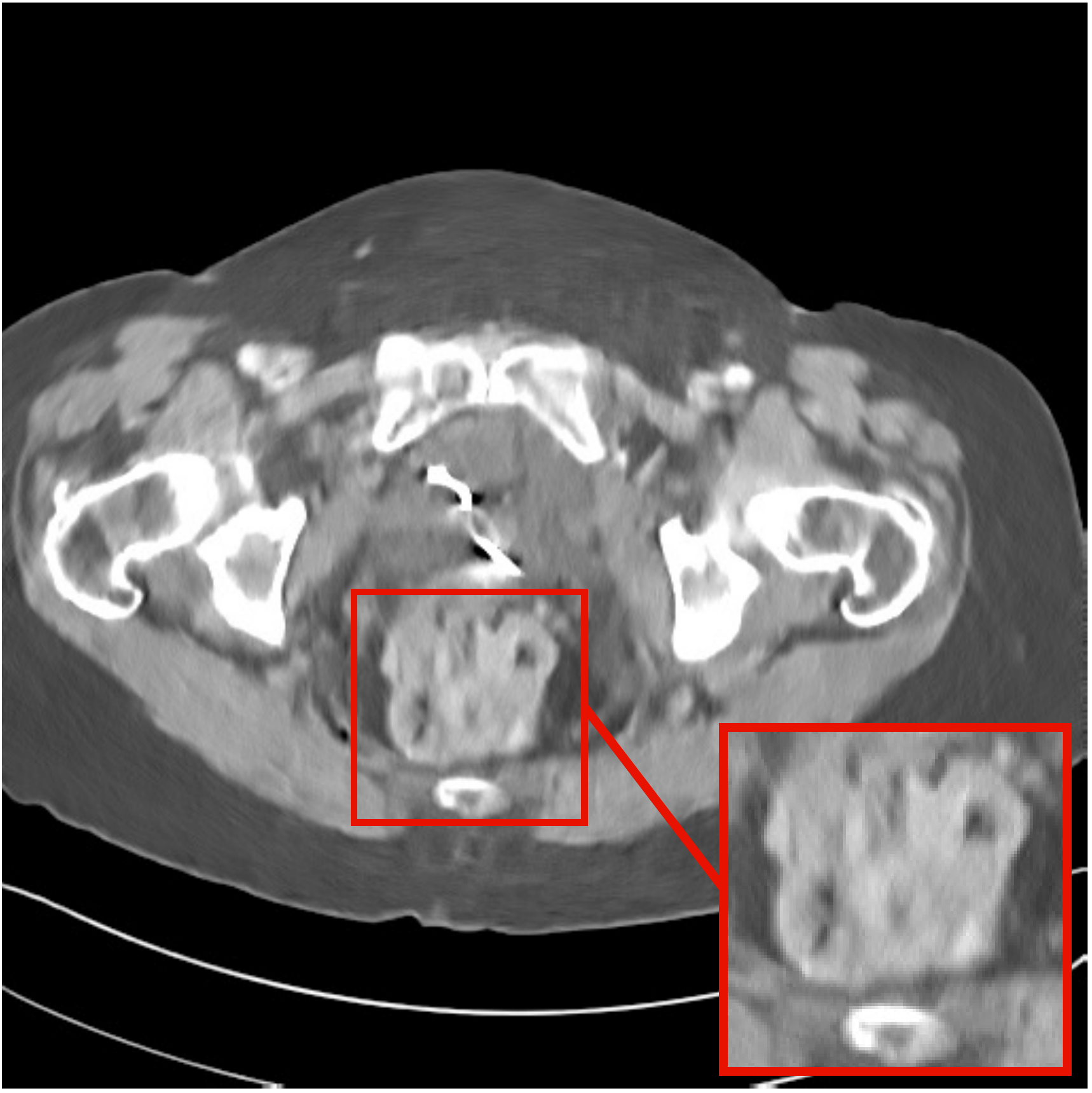} 
		\put(20,88){ \color{white}{\bf \normalsize{PSNR: 30.8}}} 	\end{overpic}
	\caption{Reconstructed testing image (Test \#5, from patient L096) obtained by FBP (top left), FBPConvNet (top right), PWLS-EP (middle left), PWLS-ULTRA (middle right), FBPConvNet + EP (bottom left), and SUPER-ULTRA (bottom right). The display window is [800 1200]~HU.}
	\label{fig:L096Slice291_image}
\end{figure}

\begin{figure*}[htb]
	\centering
	\begin{overpic}[scale=0.26]{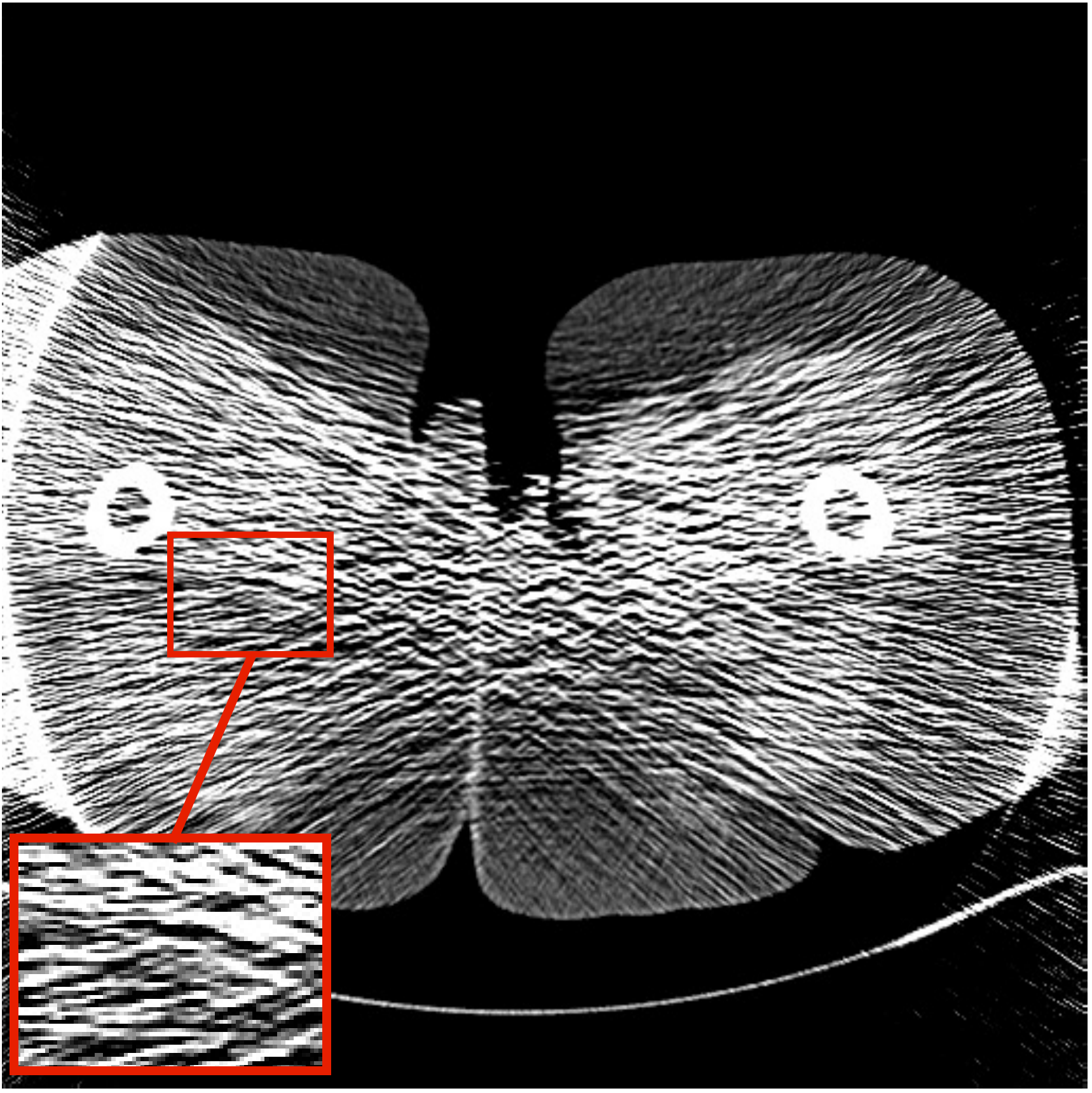} 
		\put(20,88){ \color{white}{\bf \normalsize{PSNR: 9.7}}} 	\end{overpic}
	\begin{overpic}[scale=0.26]{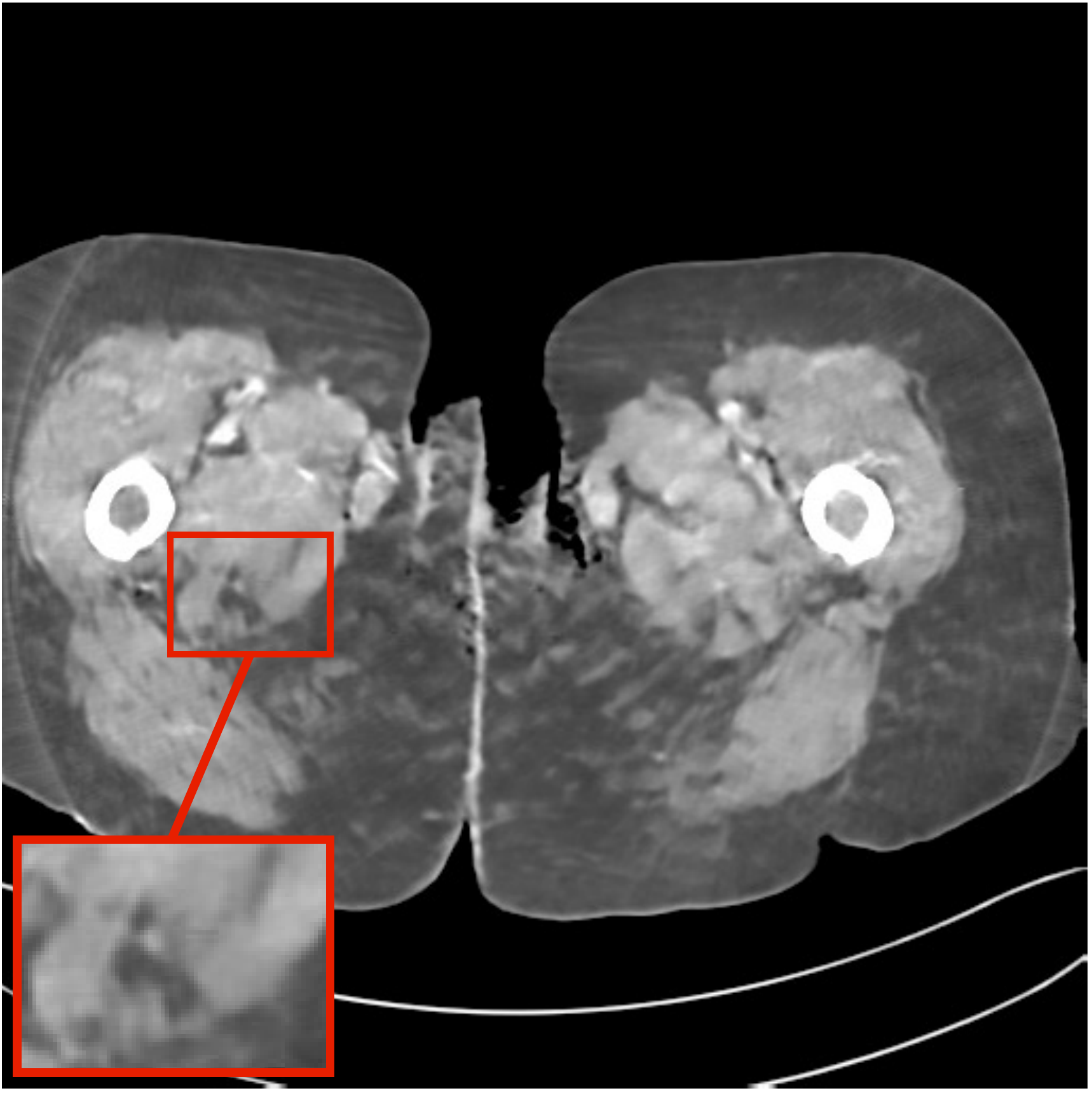} 
		\put(20,88){ \color{white}{\bf \normalsize{PSNR: 27.6}}} 	\end{overpic} 
	\begin{overpic}[scale=0.26]{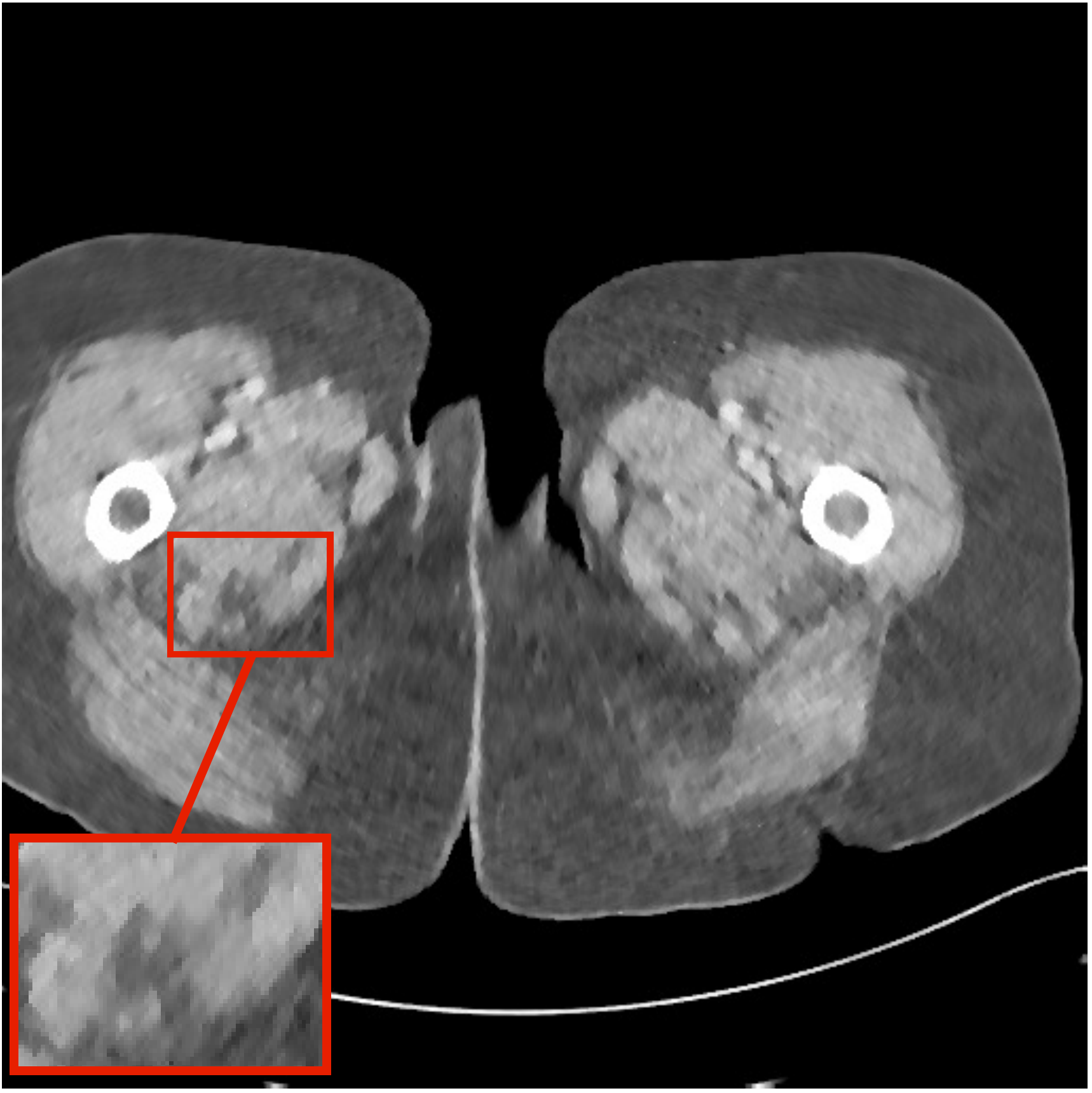} 
		\put(20,88){ \color{white}{\bf \normalsize{PSNR: 23.6}}} 	\end{overpic} \\
	\begin{overpic}[scale=0.26]{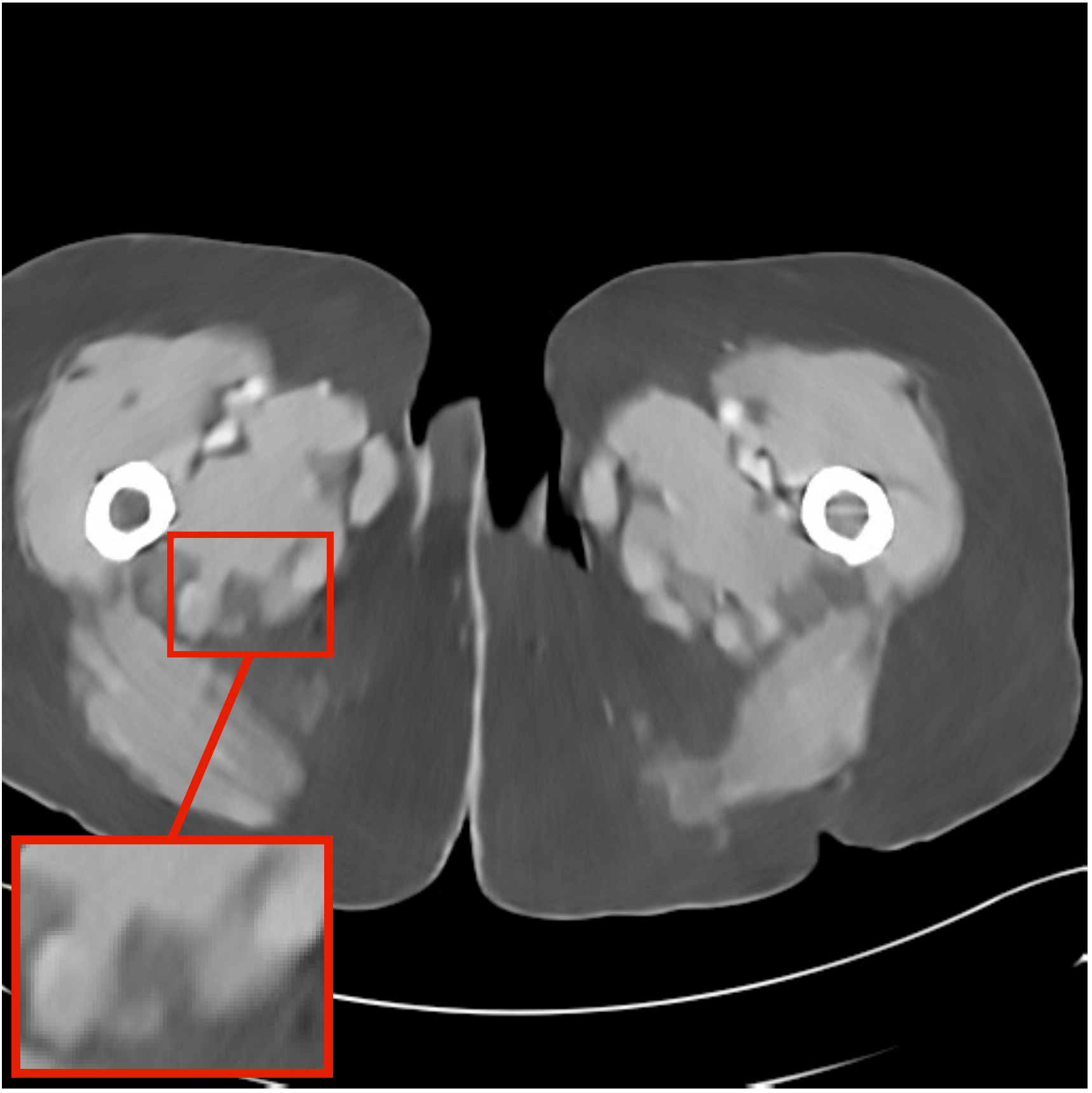} 
		\put(20,88){ \color{white}{\bf \normalsize{PSNR: 28.1}}} 	\end{overpic} 
	\begin{overpic}[scale=0.26]{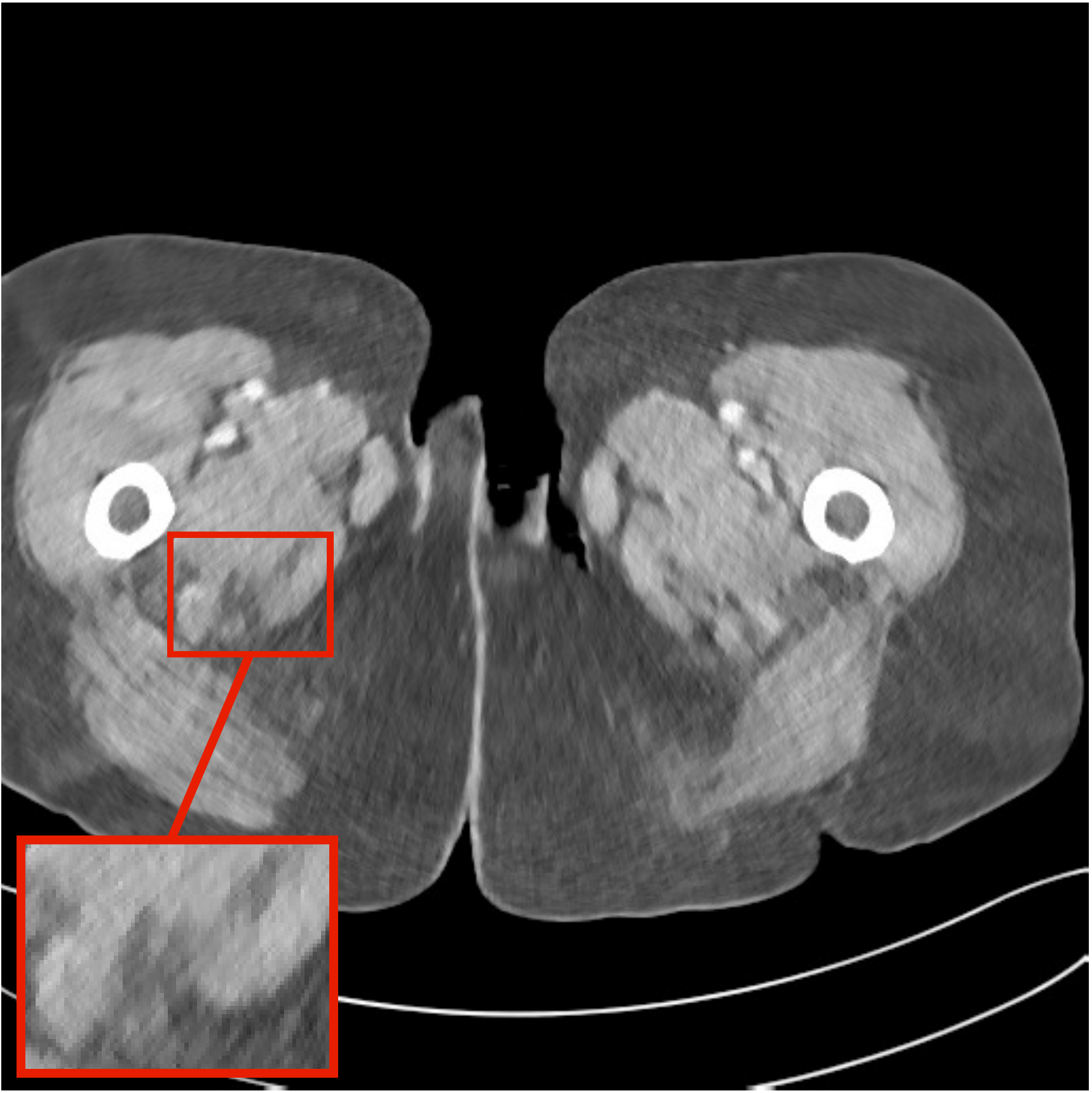} 
		\put(20,88){ \color{white}{\bf \normalsize{PSNR: \DIFaddadd{30.7}}}} 	\end{overpic}
	\begin{overpic}[scale=0.26]{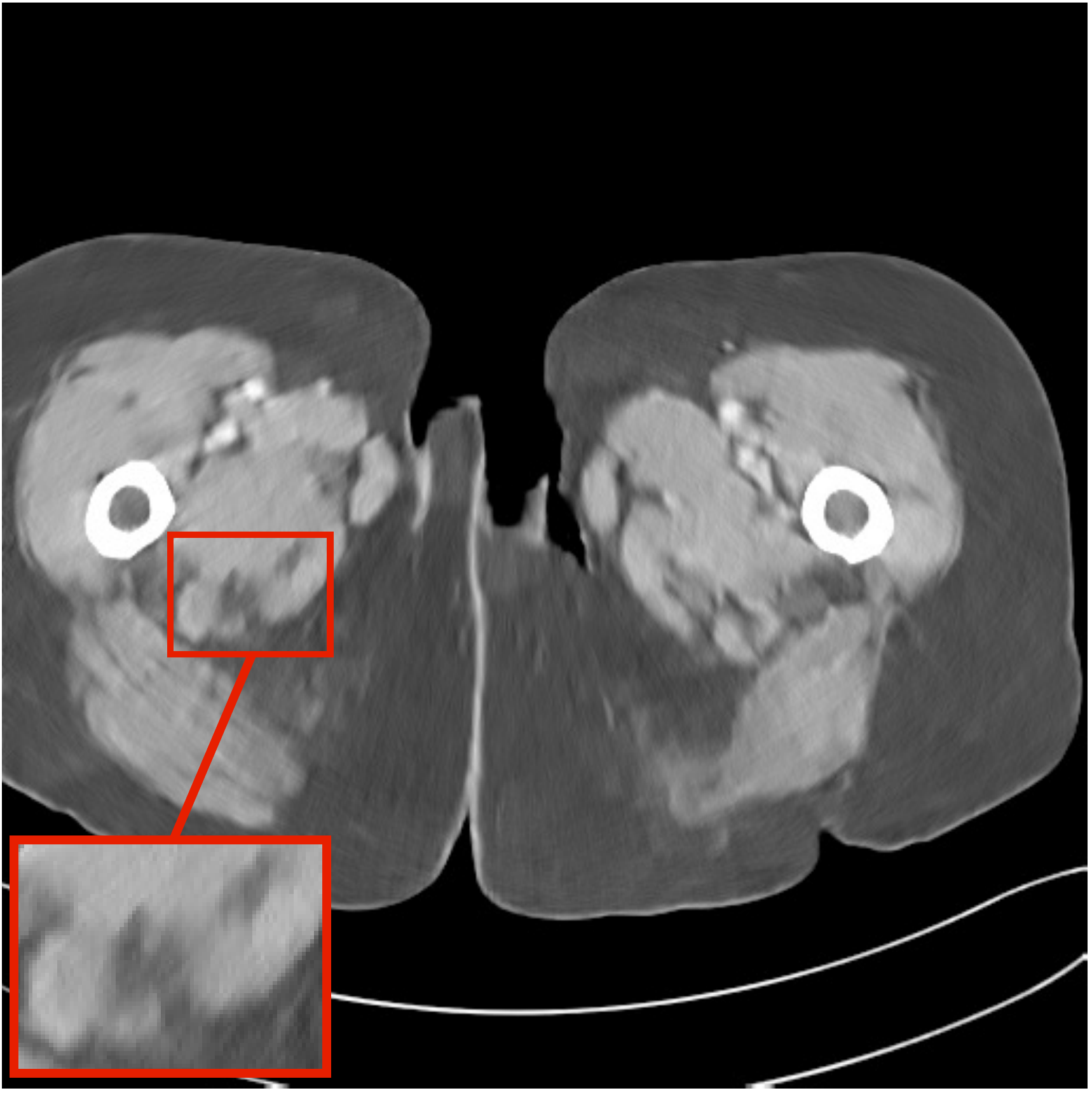} 
		\put(20,88){ \color{white}{\bf \normalsize{PSNR: 32.3}}} 	\end{overpic}
	\caption{Reconstructed testing image (Test \#6, from patient L096) obtained by FBP (top left), FBPConvNet (top middle), PWLS-EP (top right), PWLS-ULTRA (bottom left), FBPConvNet + EP (bottom middle), and SUPER-ULTRA (bottom right). The display window is [800 1200]~HU.}
	\label{fig:L096Slice330_image}
\end{figure*}

Here, we show the regular-dose FBP images for the other four testing slices (Test \#1, \#3, \#5, and \#6) in Fig.~\ref{fig:ref}, and their reconstructions in Figs.~\ref{fig:L067Slice50_image}, \ref{fig:L067Slice210_image}, \ref{fig:L096Slice291_image}, and \ref{fig:L096Slice330_image}.
\vspace{-0.07in}


SUPER-ULTRA and FBPConvNet+EP obviously reduce artifacts and noise compared to FBP, PWLS-EP, and FBPConvNet.
Furthermore, SUPER-ULTRA also significantly improves the sharpness in the soft-tissue compared to PWLS-ULTRA.
Lastly, compared to FBPConvNet+EP, SUPER-ULTRA reduces noise while producing sharp reconstructions.

\begin{figure*}[!b]
	\centering
	\begin{overpic}[scale=0.195]{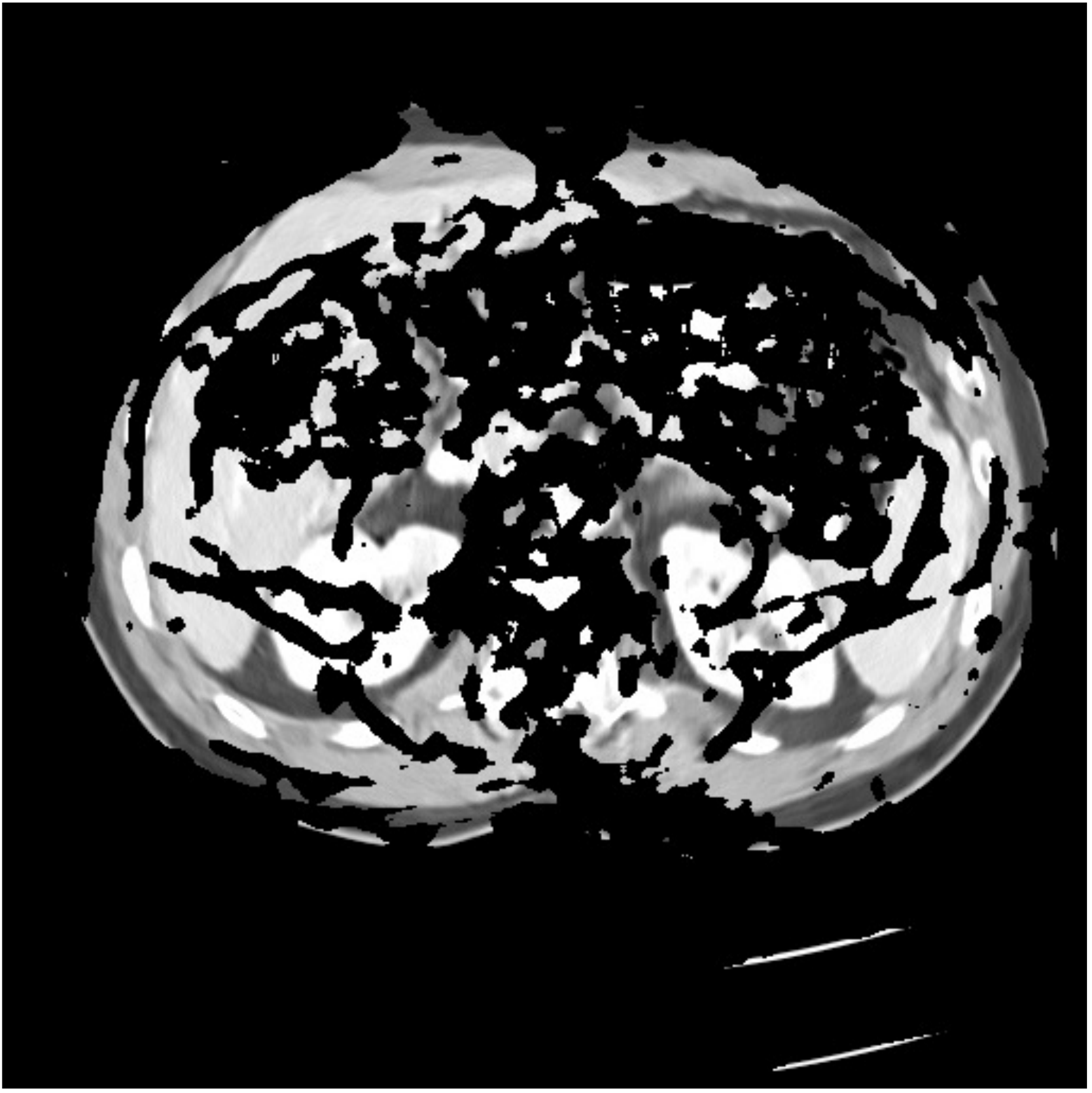}
		\put(30,90){ \color{white}{\bf \large{Class 1}}} 
	\end{overpic}
	\begin{overpic}[scale=0.195]{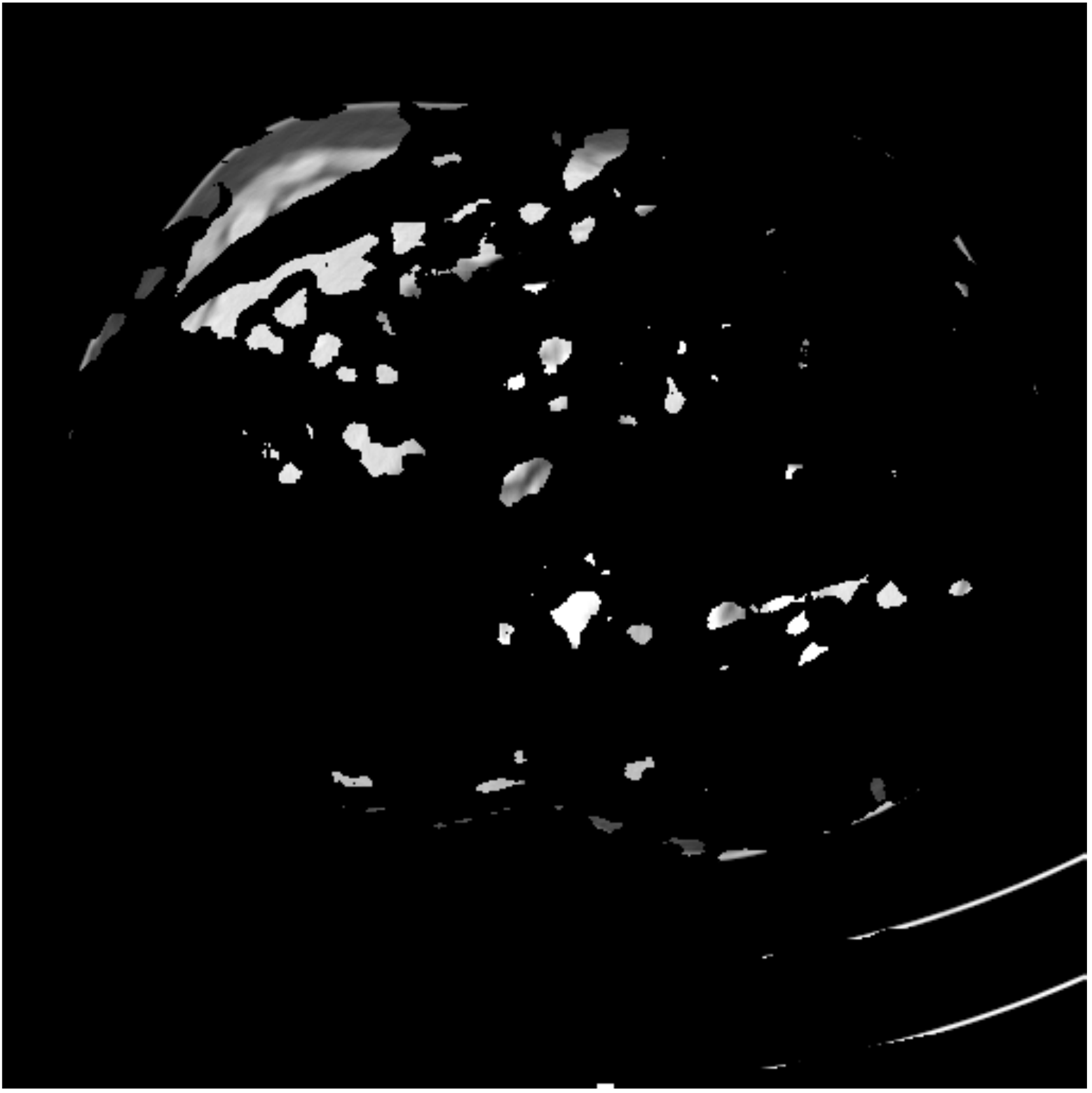}
		\put(30,90){ \color{white}{\bf \large{Class 2}}} 
	\end{overpic}
	\begin{overpic}[scale=0.195]{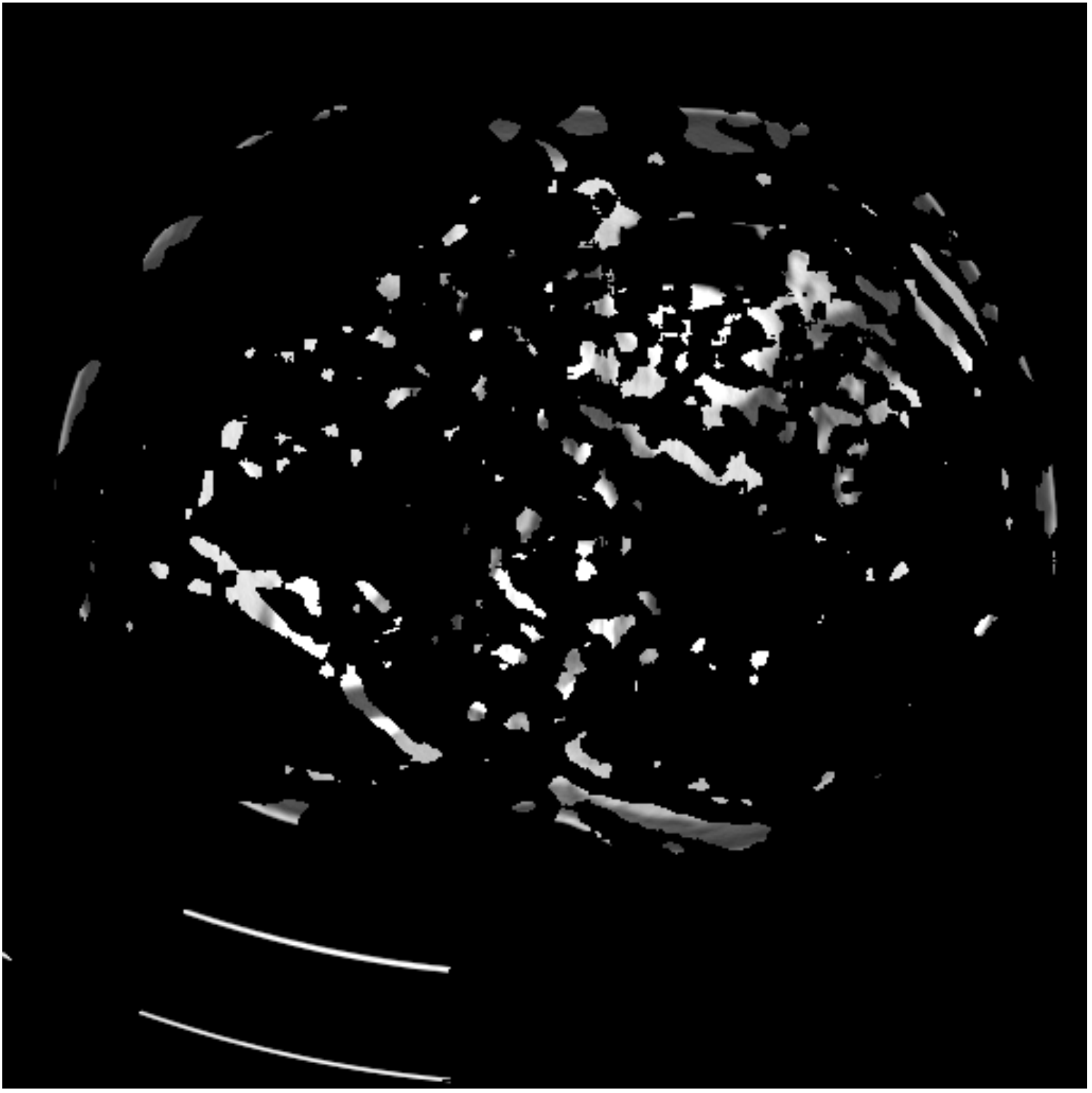}
		\put(30,90){ \color{white}{\bf \large{Class 3}}} 
	\end{overpic}
	\begin{overpic}[scale=0.195]{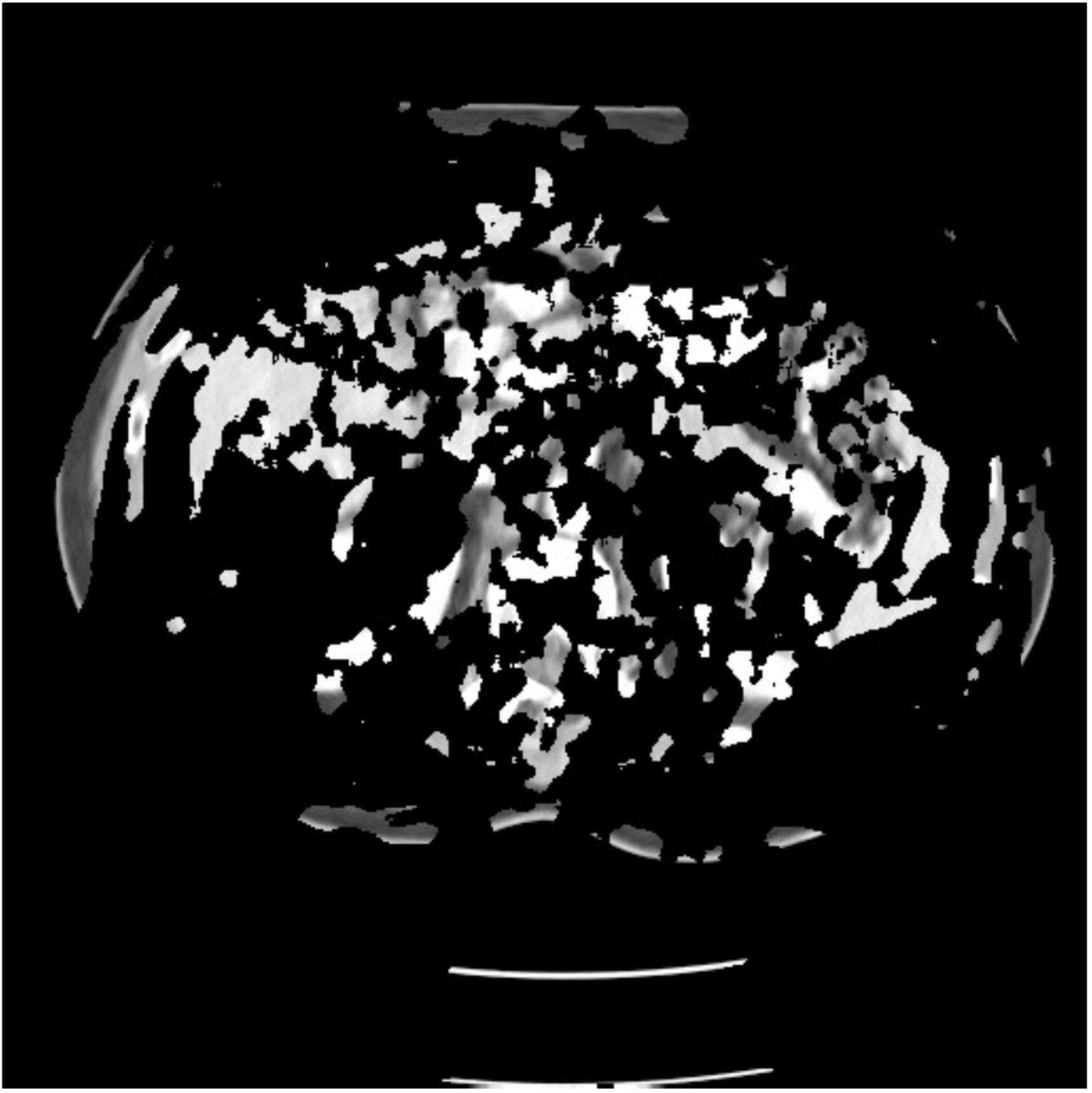}
		\put(30,90){ \color{white}{\bf \large{Class 4}}} 
	\end{overpic}
	\begin{overpic}[scale=0.195]{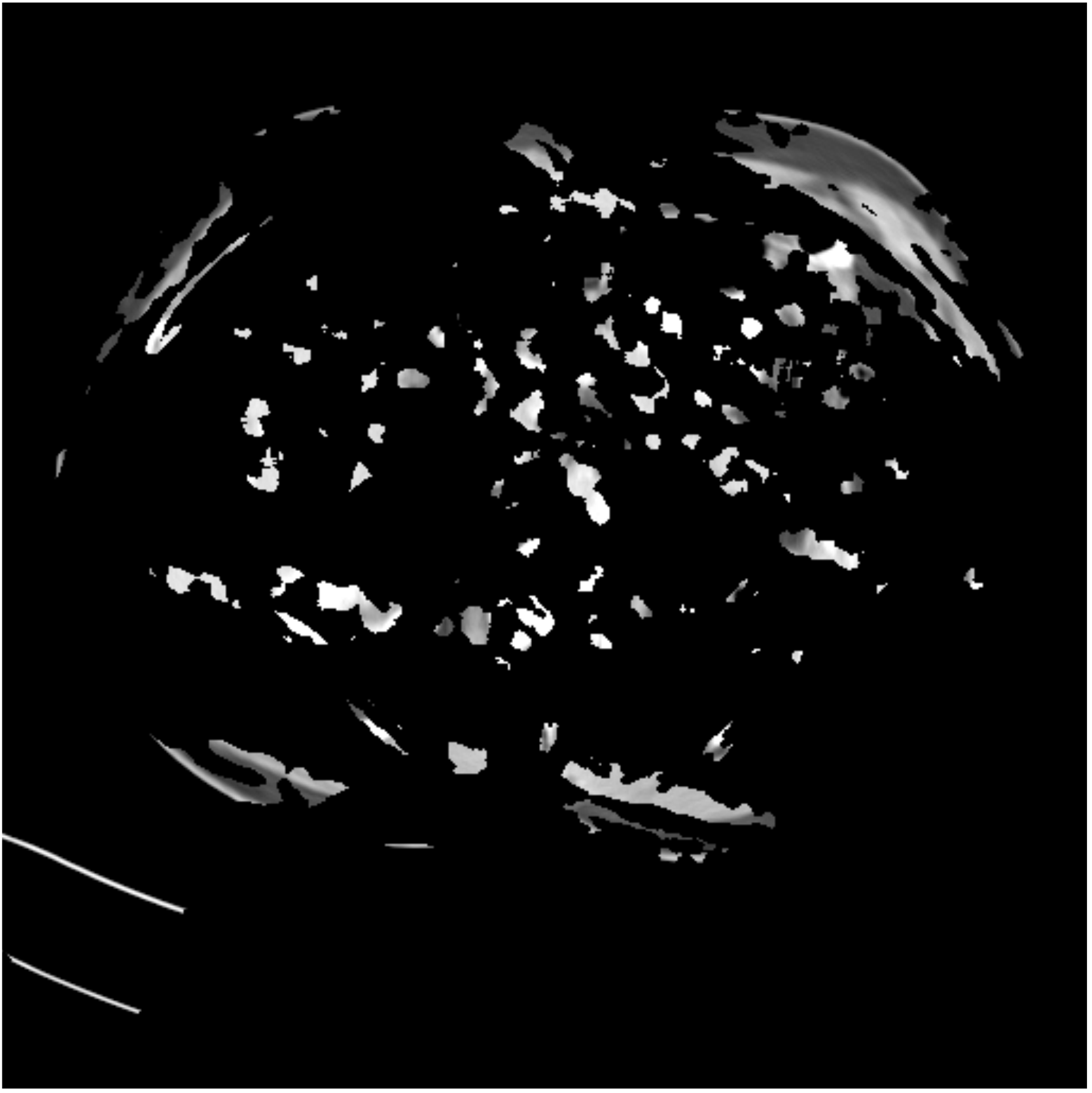}
		\put(30,90){ \color{white}{\bf \large{Class 5}}} 
	\end{overpic}\\
	\begin{overpic}[scale=0.195]{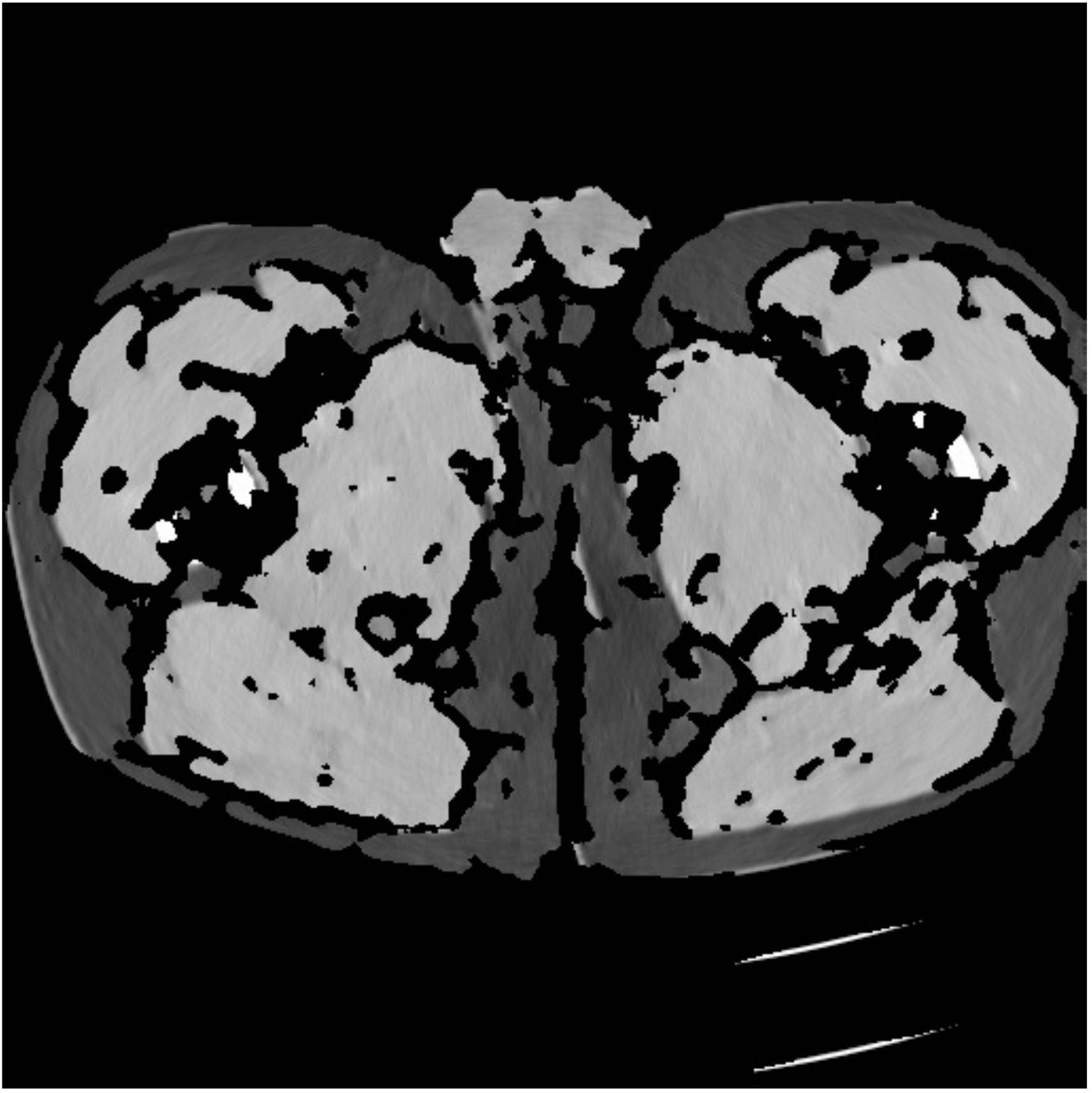}
		\put(30,87){ \color{white}{\bf \large{Class 1}}} 
	\end{overpic}
	\begin{overpic}[scale=0.195]{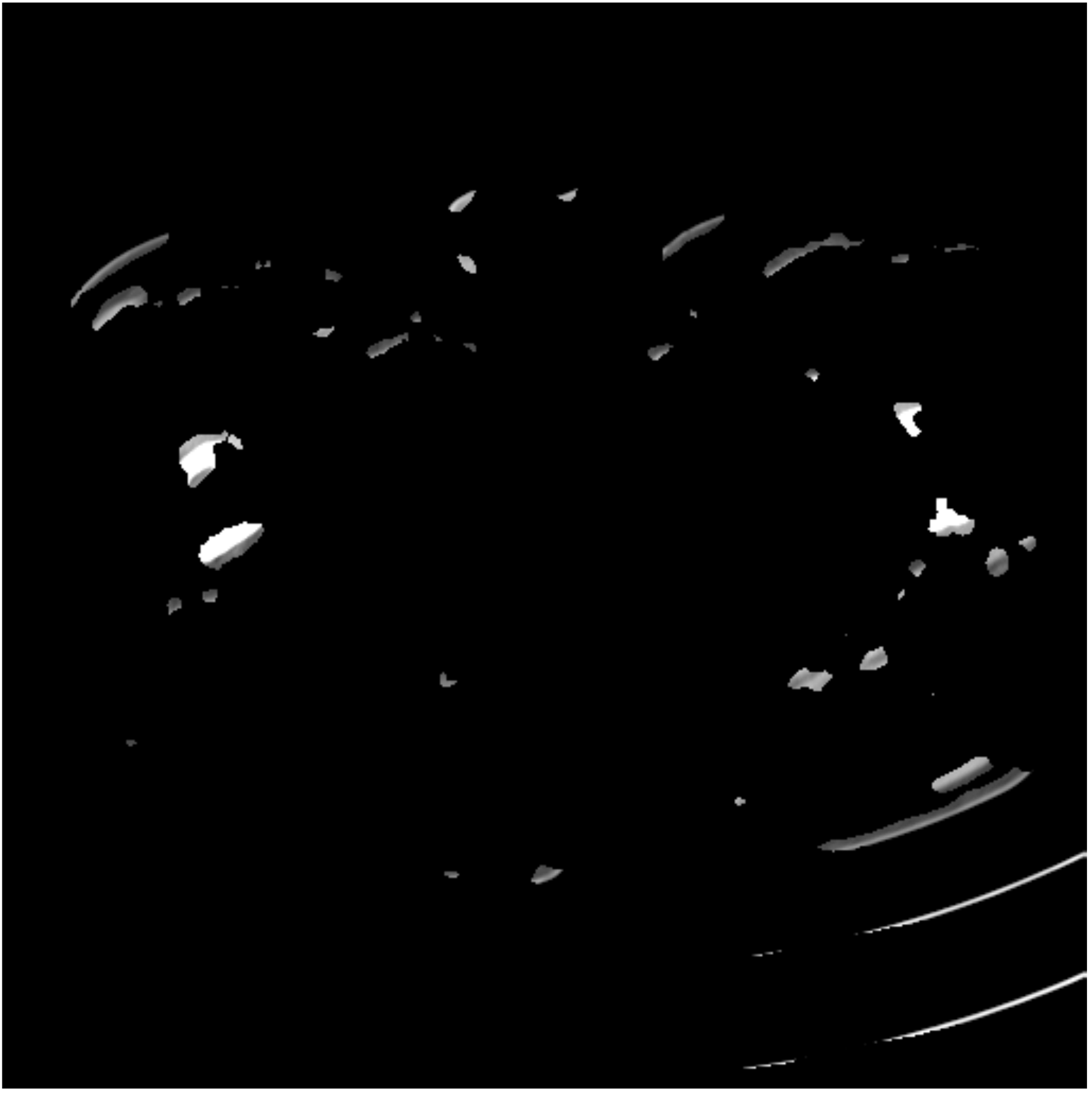}
		\put(30,85){ \color{white}{\bf \large{Class 2}}} 
	\end{overpic}
	\begin{overpic}[scale=0.195]{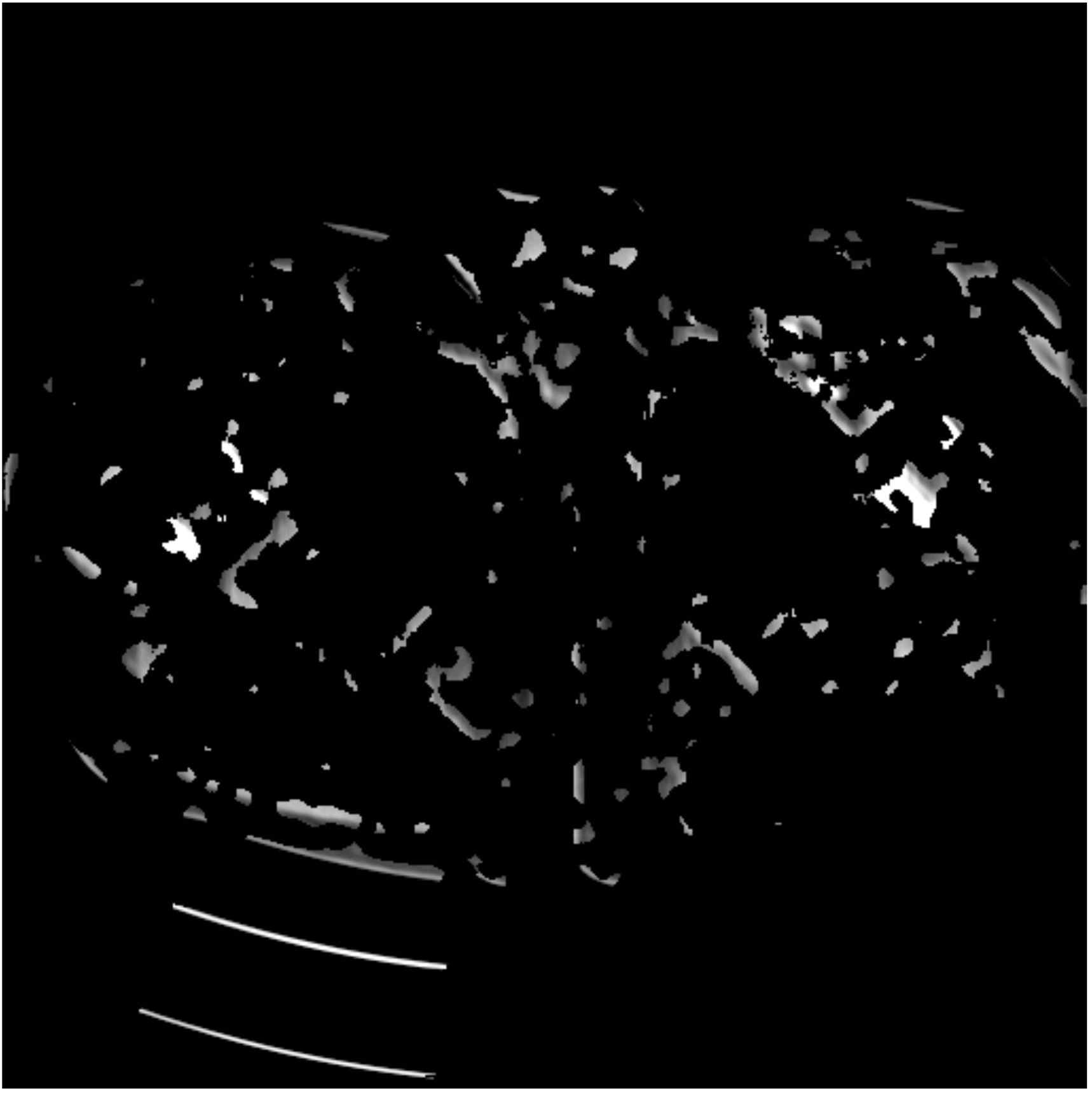}
		\put(30,85){ \color{white}{\bf \large{Class 3}}} 
	\end{overpic}
	\begin{overpic}[scale=0.195]{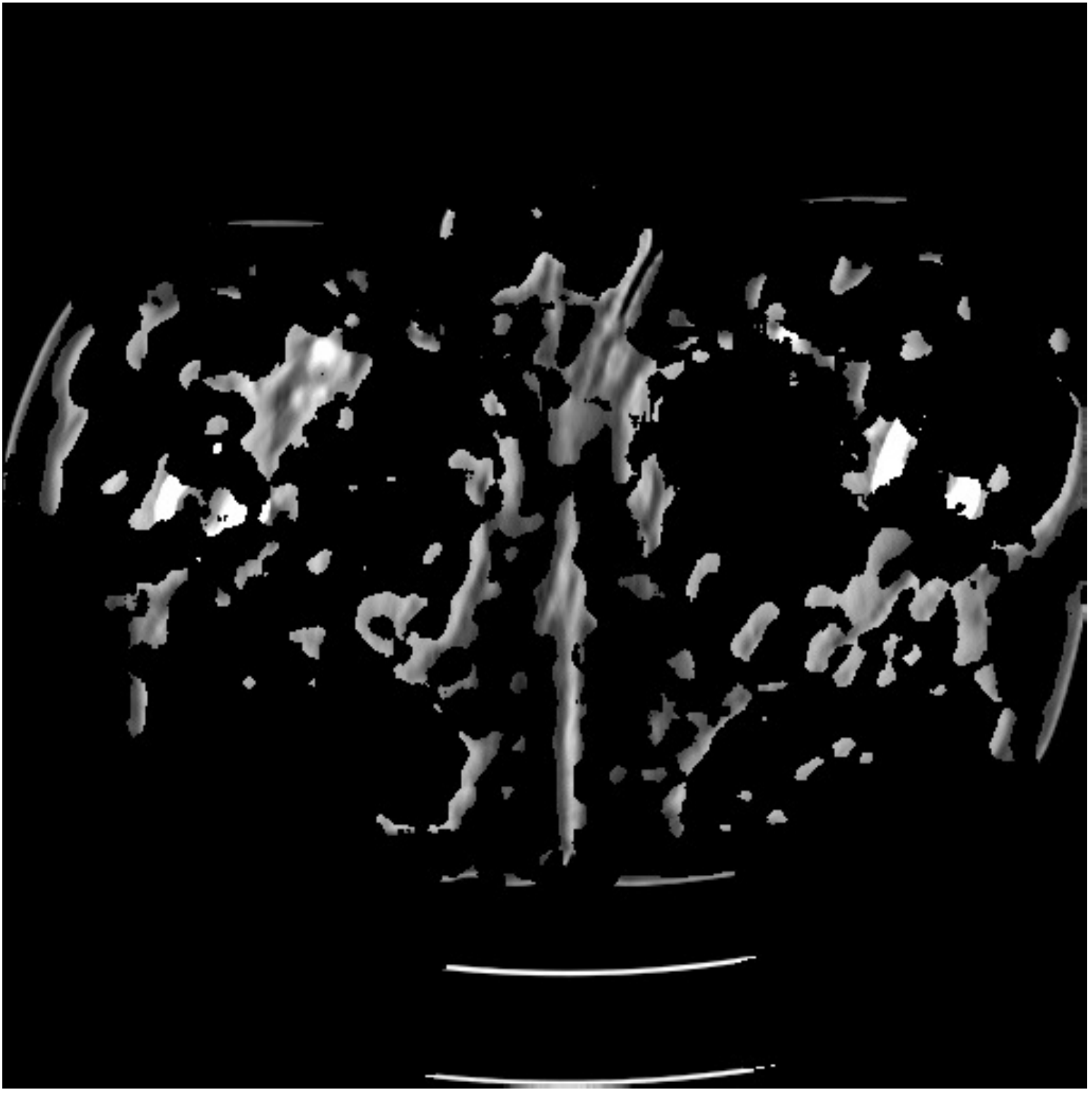}
		\put(30,85){ \color{white}{\bf \large{Class 4}}} 
	\end{overpic}
	\begin{overpic}[scale=0.195]{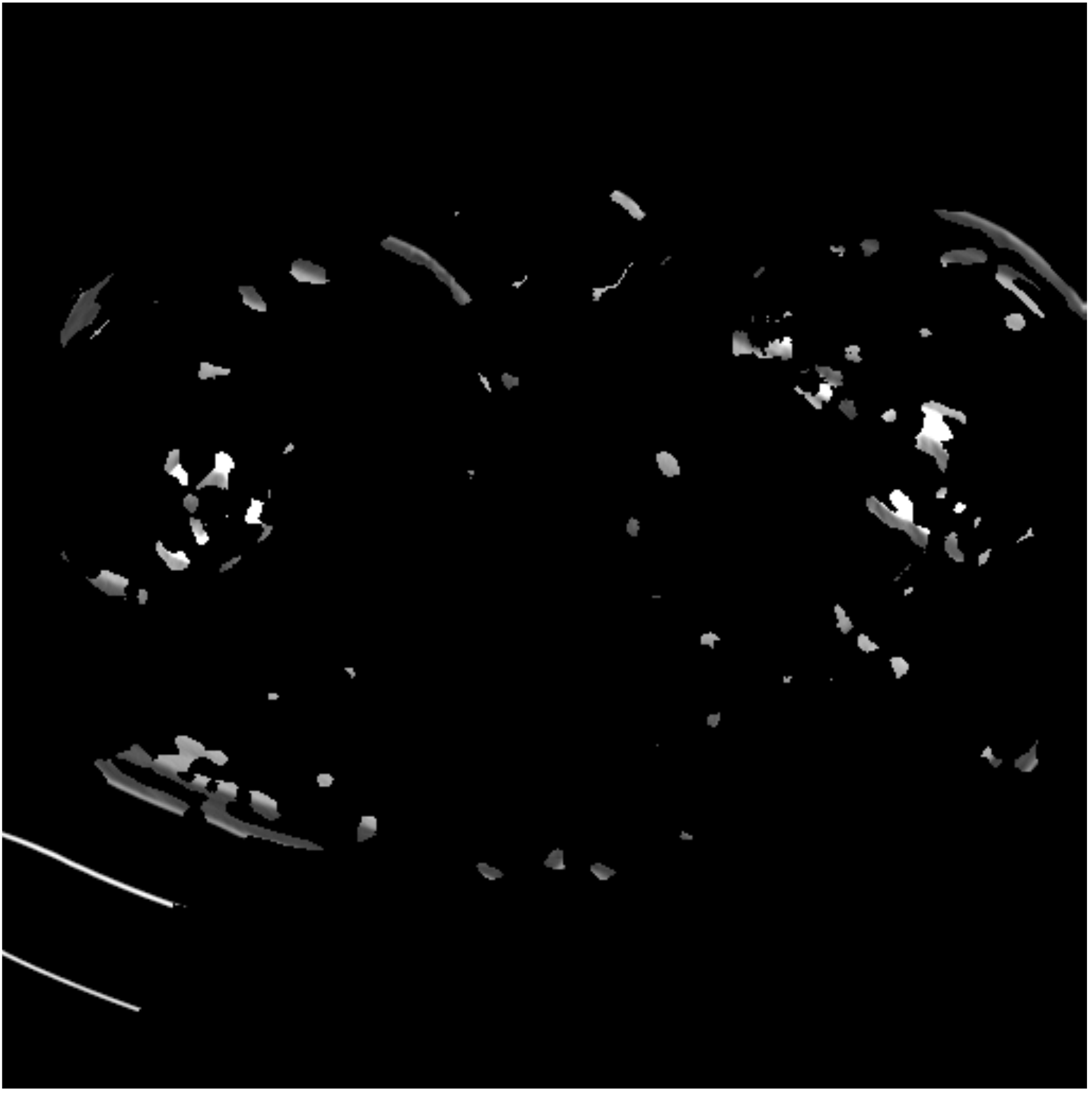}
		\put(30,85){ \color{white}{\bf \large{Class 5}}} 
	\end{overpic}
	\caption{Top and bottom rows show the Pixel-level clustering results of SUPER-ULTRA for Test \#1 and Test \#3, respectively. The display window is [800 1200]~HU.}
	\label{fig:clstering}
\end{figure*}


Next, to better illustrate the learned clustering in the SUPER-ULTRA model, Fig.~\ref{fig:clstering} shows examples of pixel-level clustering results from the last super layer for test slices \#1 and \#3.



\end{document}